%% file: ms.tex
\documentclass[final,12pt]{clear2024} % Include author names

\usepackage[numbers,compress]{natbib}
\usepackage{tocloft}
\usepackage{float}
\usepackage{color,xcolor}
\usepackage[utf8]{inputenc} % allow utf-8 input
\usepackage[T1]{fontenc}    % use 8-bit T1 fonts
\usepackage{booktabs}       % professional-quality tables
\usepackage{nicefrac}       % compact symbols for 1/2, etc.
\usepackage{microtype}      % microtypography
\usepackage{mathtools}
\usepackage[page]{appendix}
\usepackage{enumitem}
\usepackage{comment}

\DeclareMathOperator{\tr}{tr}
\newcommand{\G}{\mathcal{G}}

\DeclareSymbolFont{matha}{OML}{txmi}{m}{it}
\DeclareMathSymbol{\varv}{\mathord}{matha}{118}
\usepackage{wrapfig}
 
\usepackage{subfig}
\usepackage{tikz}
\usetikzlibrary{bayesnet}
\newdimen\arrowsize
\pgfarrowsdeclare{arcsq}{arcsq}
{
	\arrowsize=0.2pt
	\advance\arrowsize by .5\pgflinewidth
	\pgfarrowsleftextend{-4\arrowsize-.5\pgflinewidth}
	\pgfarrowsrightextend{.5\pgflinewidth}
}
{
	\arrowsize=0.8pt
	\advance\arrowsize by .5\pgflinewidth
	\pgfsetdash{}{0pt}
	\pgfsetroundjoin
	\pgfsetroundcap
	\pgfpathmoveto{\pgfpoint{0\arrowsize}{0\arrowsize}}
	\pgfpatharc{-90}{-140}{4\arrowsize}
	\pgfusepathqstroke
	\pgfpathmoveto{\pgfpointorigin}
	\pgfpatharc{90}{140}{4\arrowsize}
	\pgfusepathqstroke
}
\tikzset{>=arcsq}
\title[Structure Learning with Continuous Optimization: A Sober Look and Beyond]{Structure Learning with Continuous Optimization:\\A Sober Look and Beyond}
\usepackage{times}
\clearauthor{%
 \Name{Ignavier Ng} \Email{ignavierng@cmu.edu}\\
 \addr Carnegie Mellon University
 \AND
 \Name{Biwei Huang} \Email{bih007@ucsd.edu}\\
 \addr University of California San Diego
 \AND
 \Name{Kun Zhang} \Email{kunz1@cmu.edu}\\
 \addr Carnegie Mellon University\\
 Mohamed bin Zayed University of Artificial Intelligence
}

\begin{document}
\maketitle

\begin{abstract}%
  \input{sections/0abstract.tex}
\end{abstract}

\begin{keywords}%
  structure learning, continuous optimization, directed acyclic graphs
\end{keywords}

 \input{sections/1introduction.tex}
\input{sections/2background.tex}
\input{sections/3methodology.tex}
\input{sections/4discussion.tex}

% \clearpage
% Acknowledgments---Will not appear in anonymized version
\acks{The authors would like to thank Peter Spirtes, Clark Glymour, Pradeep Ravikumar, and the anonymous reviewers for helpful comments and suggestions. This project is partially supported by NSF Grant 2229881, the National Institutes of Health (NIH) under Contract R01HL159805, and grants from Apple Inc., KDDI Research Inc., Quris AI, and Infinite Brain Technology.}

{
\bibliography{ms}
\bibliographystyle{abbrvnat}
}

\clearpage
\appendix
\tableofcontents
\input{sections/5appendix.tex}

\end{document}

%% file: sections/0abstract.tex
This paper investigates in which cases continuous optimization for directed acyclic graph (DAG) structure learning can and cannot perform well and why this happens, and suggests possible directions to make the search procedure more reliable. \citet{Reisach2021beware} suggested that the remarkable performance of several continuous structure learning approaches is primarily driven by a high agreement between the order of increasing marginal variances and the topological order, and demonstrated that these approaches do not perform well after data standardization. We analyze this phenomenon for continuous approaches assuming equal and non-equal noise variances, and show that the statement may not hold in either case by providing counterexamples, justifications, and possible alternative explanations. We further demonstrate that nonconvexity may be a main concern especially for the non-equal noise variances formulation, while recent advances in continuous structure learning fail to achieve improvement in this case. Our findings suggest that future works should take into account the non-equal noise variances formulation to handle more general settings and for a more comprehensive empirical evaluation. Lastly, we provide insights into other aspects of the search procedure, including thresholding and sparsity, and show that they play an important role in the final solutions.

%% file: sections/1introduction.tex
\section{Introduction}
Bayesian networks are a class of probabilistic graphical models that encode probabilistic distributions in a compact way \citep{Pearl1988probabilistic,Koller09probabilistic}. Learning their graphical structures from data, represented by directed acyclic graphs (DAGs), has been applied in various fields, including genetics \citep{Peters2017elements} and education~\citep{Gong2022neurips}. Classical approaches for structure learning typically involve discrete procedures, such as constraint-based methods using conditional independence tests \citep{Spirtes1991pc,Spirtes2001causation} and score-based methods that search for a high-scoring structure \citep{Koivisto2004exact,Singh2005finding,Cussens2011bayesian,Yuan2013learning,Chickering2002optimal,Peters2013identifiability}. Greedy search is often employed in score-based methods because of the large space of possible structures \citep{Chickering1996learning,Chickering2004large}, such as GES \citep{Chickering2002optimal} and GDS \citep{Peters2013identifiability}.

Recently, \citet{Zheng2018notears} proposed a smooth characterization of acyclicity and transformed the structure learning problem of discrete nature into a continuous, nonconvex optimization problem, thus enabling the application of efficient gradient-based optimization. This formulation has been extended and applied to a wide range of settings, including nonlinear cases \citep{Yu19daggnn, Lachapelle2020grandag, Zheng2020learning,Ng2022masked,Kalainathan2022structural}, interventional data~\citep{brouillard2020differentiable,Faria2022differentiable}, unobserved confounding \citep{Bhattacharya2020differentiable,Bellot2021deconfounded}, incomplete data \citep{Wang2020causal,Gao2022missdag}, time series \citep{Pamfil2020dynotears,Sun2021ntsnotears}, multi-task learning~\citep{Chen2021multitask}, multi-domain data \citep{Zeng2020causal}, federated learning \citep{Ng2022towards,Gao2023feddag}, and representation learning \citep{Yang2021causalvae}.

Given the growing interest in continuous structure learning \citep{Vowels2022dags}, various theoretical and empirical aspects of these approaches have gained considerable attention. In particular, \citet{Wei2020nofears,Ng2022convergence} studied the optimality conditions and convergence property of continuous constrained approaches \citep{Zheng2018notears}, while \citet{Deng2023global} showed that a proper optimization scheme converges to the global minimum of the least squares objective in the bivariate case. \citet{Zhang2022truncated,Bello2022dagma} demonstrated that existing DAG constraints \citep{Zheng2018notears,Yu19daggnn} may encounter gradient vanishing issues in practice and proposed improved variants. \citet{Reisach2021beware} suggested that the remarkable performance of several continuous structure learning approaches \citep{Zheng2018notears,Ng2020role} is primarily driven by a high agreement between the order of increasing marginal variances and the topological order, and demonstrated that these approaches do not perform well after data standardization. Similar phenomenon was also observed by \citet{Kaiser2022unsuitability}. These empirical findings regarding data standardization provide insights into the performance of continuous structure learning, and may at first appear surprising. One of our goals is to provide further analysis of this phenomenon.

In this work, we investigate in which cases continuous structure learning approaches can and cannot perform well and why this happens, and suggest possible directions to make the search procedure more reliable. Our main contributions are:
\begin{itemize}
\setlength\itemsep{0em}
\item We analyze the statements and observations by \citet{Reisach2021beware} in the linear case with equal (Section \ref{sec:varsortability_ev_case}) and non-equal noise variances (Section \ref{sec:varsortability_nv_case}), and show that the statements may not hold in either case by providing counterexamples and justifications. We also provide possible alternative explanations for the observations by \citet{Reisach2021beware} that continuous structure learning approaches do not perform well after data standardization.
\item  We show that nonconvexity may be a main concern especially for the non-equal noise variances formulation, while recent advances in continuous structure learning, including search strategies, DAG contraints, and nonlinear approaches, fail to achieve improvement in this case (Section \ref{sec:nonconvexity}). Our findings suggest that future works should take into account the non-equal noise variances formulation to handle more general settings and for a more comprehensive empirical evaluation.\looseness=-1
\item  We provide insights into other aspects of the search procedure, including thresholding (Section~\ref{sec:thresholding}) and sparsity (Section \ref{sec:sparsity}), and show that they play a crucial role in the final solutions.\looseness=-1
\end{itemize}

%% file: sections/2background.tex
\section{Preliminaries}\label{sec:preliminaries}
In this section, we briefly describe several existing continuous structure learning approaches that this work focuses on, and explain the notion of varsortability proposed by \citet{Reisach2021beware}.

\vspace{-0.05em}
\subsection{Structure Learning with Continuous Optimization}\label{sec:continuous_structure_learning}
\paragraph{Setup.}
Let $\G$ be a DAG with node set $\{X_1,\dots,X_d\}$. In a Bayesian network, each node $X_i$ of $\G$ corresponds to a random variable, and the joint distribution of random variables $X=(X_1,\dots,X_d)$, denoted as $P(X)$, is Markov w.r.t. DAG $\G$. We consider the linear case in which variables $X$ follow a linear structural equation model (SEM) $X=B^T X + N$, where $B\in\mathbb{R}^{d\times d}$ is an acyclic weighted adjacency matrix whose nonzero entries represent the edges in DAG $\G$, and $N=(N_1,\dots,N_d)$ consists of the independent noise variables with covariance matrix $\Omega\coloneqq\operatorname{diag}(\sigma_1^2,\dots,\sigma_d^2)$. If $\sigma_1^2=\dots=\sigma_d^2$, we refer to it as the equal noise variances (EV) case, and otherwise as the non-equal noise variances (NV) case. Structure learning aims at estimating DAG $\G$ or its Markov equivalence class (MEC) given the data matrix $\mathbf{X}\in\mathbb{R}^{n\times d}$ consisting of $n$ i.i.d. samples from distribution $P(X)$.

\paragraph{NOTEARS.}
\citet{Zheng2018notears} proposed to solve the constrained optimization problem
\begin{equation*}\label{eq:notears}
\vspace{-0.1em}
	\min_{B\in\mathbb{R}^{d\times d}}\ \  \ell(B;\mathbf{X})\coloneqq\frac{1}{2n}\|\mathbf{X} -  \mathbf{X}B\|_F^2 \quad \mathrm{subject~to}\ \  h(B)= 0,
\end{equation*}
 where $\ell(B;\mathbf{X})$ is the least squares objective and $h(B)\coloneqq\tr(e^{B\odot B}) - d$ is the DAG constraint term. An $\ell_1$ penalty term $\lambda\|B\|_1$ is also incorporated into the objective function, where $\|\cdot\|_1$ is defined element-wise and $\lambda$ is a hyperparameter. Since the above formulation focuses on the linear case with equal noise variances \citep{Peters2013identifiability,Loh2014high}, we refer to it as NOTEARS-EV throughout this work. 

\paragraph{GOLEM.}
\citet{Ng2020role} solve an unconstrained optimization problem:
\begin{equation}\label{eq:golem}
\min_{B\in\mathbb{R}^{d\times d}} \ \ \mathcal{L}(B;\mathbf{X}) - \log|\det(I - B)| + \lambda_1\|B\|_1 + \lambda_2 h(B),
\end{equation}
where $\mathcal{L}(B;\mathbf{X})$ is defined as
\begin{flalign*}
\mathcal{L}_\text{EV}(B;\mathbf{X})=\frac{d}{2}\log\|\mathbf{X} -  \mathbf{X}B\|_F^2 \text{\quad and \quad}
\mathcal{L}_\text{NV}(B;\mathbf{X})=\frac{1}{2}\sum_{i=1}^d\log\|\mathbf{X}_{\cdot,i}-\mathbf{X}B_{\cdot,i} \|_2^2
\end{flalign*}
in the linear Gaussian case with equal and non-equal noise variances, respectively. Here, $\lambda_1$ and $\lambda_2$ are hyperparameters. The above two variants are denoted as GOLEM-EV and GOLEM-NV, respectively.\looseness=-1

\vspace{-0.1em}
\subsection{Varsortability}\label{sec:background_varsortability}
Recently, \citet{Reisach2021beware} introduced \emph{varsortability}, denoted as $\varv$, as a measure of agreement between increasing marginal variances and topological order. It is defined as the proportion of directed paths which start from a variable with lower marginal variance than the variable they end in~\citep{Reisach2021beware}. When all variables have higher marginal variances than their ancestors, we have $\varv=1$.

\citet{Reisach2021beware} provided insights into the performance of several continuous structure learning approaches \citep{Zheng2018notears,Ng2020role}, and suggested that their remarkable performance is primarily driven by high varsortability. Here we list some excerpts from \citet{Reisach2021beware}: ``Our experiments demonstrate that varsortability dominates the optimization and helps achieve state-of-the-art performance provided the ground-truth data scale'' and ``we focus on the first optimization steps to explain a) why continuous structure learning algorithms that assume equal noise variance work remarkably well in the presence of high varsortability''. For clarity and ease of further analysis, we give a partial formulation of the statements (on which this work focuses) in the following. We provide a further discussion of how these statements are formulated and possible alternative formulations in Appendix \ref{app:discussion_statements}.
\begin{statement}[Equal Noise Variances Formulation]\label{statement:varsortability_ev_case}
Continuous structure learning approaches that assume equal noise variances, specifically NOTEARS-EV and GOLEM-EV, perform well in the presence of high varsortability. \vspace{-0.4em}
\end{statement}
\begin{statement}[Non-Equal Noise Variances Formulation]\label{statement:varsortability_nv_case}
Continuous structure learning approaches that assume non-equal noise variances, specifically GOLEM-NV, perform well in the presence of high varsortability.
\end{statement}
\citet{Reisach2021beware} showed that the synthetic data used by \citet{Zheng2018notears,Ng2020role} to benchmark their methods exhibits high varsortability, e.g., higher than $0.94$ on average, and that these approaches do not perform well after data standardization which removes such patterns in the marginal variances.\looseness=-1

%% file: sections/3methodology.tex
\section{Varsortability and Data Standardization}\label{sec:varsortability}
In this section, we analyze the statements and observations by \citet{Reisach2021beware} in the linear case with equal and non-equal noise variances formulations, and show that the statements may not hold in either case. That is, continuous structure learning approaches often do not perform well even in the presence of high varsortability. We also provide possible alternative explanations for the observations by \citet{Reisach2021beware} that continuous approaches do not perform well after data standardization.

\vspace{-0.05em}
\subsection{With Equal Noise Variances Formulation}\label{sec:varsortability_ev_case}

\paragraph{Varsortability and continuous structure learning.}
We first take a closer look at Statement \ref{statement:varsortability_ev_case} and consider the least squares score \citep{Loh2014high,Zheng2018notears}. For bivariate case, say $X=(X_1,X_2)$ with ground truth $X_1\rightarrow X_2$, \citet{Reisach2021beware} showed that estimations based on varsortability and least squares are consistent in the sense that $\varv=1$, i.e., $\operatorname{Var}(X_1)<\operatorname{Var}(X_2)$ if and only if the least squares score computed with $X_1\rightarrow X_2$ is smaller than that with $X_2\rightarrow X_1$; see \citet[Appendix~A]{Reisach2021beware} for a proof. Thus, with high varsortability in the bivariate case, i.e., $\varv=1$, NOTEARS-EV, loosely speaking, is able to asymptotically output the correct structure assuming that the global minimizer can be found. Therefore, Statement \ref{statement:varsortability_ev_case} holds in the bivariate case, at least for NOTEARS-EV.

However, the property in the bivariate case above cannot be extended to general cases with more than two variables; see the following example with three variables.
\begin{example}
\label{example:single_counterexample}
Consider the linear SEM over $X=(X_1,X_2,X_3)$ with the true weighted adjacency matrix $\tilde{B}$ and noise covariance matrix $\tilde{\Omega}$, as well as an alternative weighted adjacency matrix $\hat{B}$:
\begin{equation*}
\tilde{B} = \begin{bmatrix}
0           & 0           & 0 \\ 
\frac{1}{2} & 0           & 0 \\
1           & \frac{1}{2} & 0
\end{bmatrix},\quad
\tilde{\Omega} = \begin{bmatrix}
\frac{1}{2} & 0 & 0 \\ 
0           & 1 & 0 \\
0           & 0 & \frac{1}{2}
\end{bmatrix},\quad
\hat{B} = \begin{bmatrix}
0           & \frac{2}{3}  & 0 \\ 
0           & 0            & 0 \\
\frac{5}{4} & -\frac{1}{3} & 0
\end{bmatrix}.
\end{equation*}
In the large sample limit, we have
\[
\operatorname{Var}(X_1)>\operatorname{Var}(X_2)>\operatorname{Var}(X_3) \text{\quad and \quad} 
\ell(\hat{B};\mathbf{X})<\ell(\tilde{B};\mathbf{X}).
\]
That is, even in the presence of high varsortability, i.e., $\varv=1$, the estimation with least squares will be the structure indicated by $\hat{B}$, instead of that indicated by $\tilde{B}$, and thus Statement \ref{statement:varsortability_ev_case} does not hold.
\end{example}
The weighted adjacency matrices $\tilde{B}$ and $\hat{B}$ represent the triangle structures in Figures \ref{fig:ground_truth_triangle} and~\ref{fig:alternative_triangle}, respectively. Here a discrete exhaustive DAG search with least squares score cannot return the true structure, and similarly for NOTEARS-EV assuming that the global minimizer can be found. This is also the case for GOLEM-EV because $\mathcal{L}_\text{EV}(\hat{B};\mathbf{X})<\mathcal{L}_\text{EV}(\tilde{B};\mathbf{X})$.
To verify it, we conduct $100$ simulations using $\tilde{B}$ and $\tilde{\Omega}$ defined in Example \ref{example:single_counterexample}, and generate $10^6$ random samples in each simulation. In all simulations, we have $\operatorname{Var}(X_1)>\operatorname{Var}(X_2)>\operatorname{Var}(X_3)$, and observe that NOTEARS-EV (with $\lambda=0$ and threshold of $0.1$), GOLEM-EV (with $\lambda_1=0.02$, $\lambda_2=5$, and threshold of~$0.1$), and exhaustive search with least squares score return the structure in Figure \ref{fig:alternative_triangle}. In the example above, even in the presence of high varsortability, i.e., $\varv=1$, NOTEARS-EV and GOLEM-EV return a completely incorrect structure, indicating that Statement \ref{statement:varsortability_ev_case} does not hold in general.

\input{sections/triangle_examples.tex}

One may wonder whether the parameters of $\tilde{B}$ and $\tilde{\Omega}$ in Example \ref{example:single_counterexample} have to be exactly ``tuned'' to obtain such an outcome, which is analogous to the violation of faithfulness assumption \citep{Spirtes2001causation} that occurs with Lebesgue measure zero. We show that this is generally not the case via the following proposition; specifically, such counterexample exists for a set of parameters with nonzero Lebesgue measure. The proof is provided in Appendix \ref{app:proof_nonzero_measure_counterexample}.
\begin{proposition}
\label{proposition:nonzero_measure_counterexample}
Consider the parameters $(B,\Omega)$ of the linear SEMs over variables $X=(X_1,\dots,X_d)$, $d\geq 3$, where the noise variables follow Gaussian distributions. In the large sample limit, the set of parameters such that varsortabiltiy equals one and that the true DAG does not have the lowest least squares score has a nonzero Lebesgue measure.
\end{proposition}
\vspace{-0.3em}
On the other hand, the proposition below shows that there are cases where the varsortability is low but the true DAG yields the lowest least squares score, with a proof given in Appendix \ref{app:proof_low_varsortability_but_correct_dag}. To illustrate, if $\G$ is a fully connected DAG or a chain, then we have $\varv_\G^{\textrm{source}}=\frac{2^{d-1}-1}{2^d-d-1}$ or $\varv_\G^{\textrm{source}}=\frac{2}{d}$, respectively. Note that, as $d\rightarrow\infty$, their limits equal $0.5$ or $0$, respectively, indicating a low varsortability.  However, loosely speaking, methods that adopt least squares score, such as NOTEARS-EV and discrete search, are able to asymptotically find the correct structure assuming that the global minimizer can be found.

\vspace{-0.35em}
\begin{proposition}
\label{proposition:low_varsortability_but_correct_dag}
Consider the parameters $(B,\Omega)$ of the linear SEMs over variables $X=(X_1,\dots,X_d)$ induced by DAG $\G$, where the noise variables follow Gaussian distributions. In the large sample limit, the set of parameters such that varsortabiltiy equals $\varv_\G^{\textrm{source}}$ and that the true weighted adjacency matrix $B$ yields the lowest least squares score has a nonzero Lebesgue measure, where 
\vspace{-0.5em}
\[
\varv_\G^{\textrm{source}}\coloneqq \frac{\textrm{Number of distinct paths from the source nodes in DAG $\G$}}{\textrm{Number of distinct paths in DAG $\G$}}.
\]
\end{proposition}
\vspace{-0.7em}
\paragraph{An alternative explanation by noise ratio.}
The above argument indicates that Statement \ref{statement:varsortability_ev_case} generally does not hold, and so may not explain why NOTEARS-EV does not perform well after data standardization. Here, we provide a possible alternative explanation of this phenomenon, i.e., the theoretical guarantee of least squares used by NOTEARS-EV does not accommodate standardization in general; thus, such phenomenon may not be surprising. Specifically, define the noise ratio
$r=\frac{\max(\sigma_1^2, \dots, \sigma_d^2)}{\min(\sigma_1^2, \dots, \sigma_d^2)}$,
which may intuitively be viewed as a measure of how far the SEM is from having equal noise variances. \citet[Theorem~9]{Loh2014high} showed that if $r< 1+\frac{\xi}{d}$, where $\xi$ is defined as the difference between the score of the true DAG and the next best DAG, then, in the large sample limit, minimizing the least squares in the space of acyclic weighted adjacency matrices $B$ returns the true structure for linear SEMs. This allows a certain degree of misspecification of the noise variances---if $\xi$ is larger, then the least squares score will be more robust to such misspecification. The result of equal noise variances \citep[Theorem~7]{Loh2014high} can be viewed as a special case with $r=1< 1+\frac{\xi}{d}$.

Thus, assuming that the global minimizer can be found, NOTEARS-EV, loosely speaking, is able to return the true DAG in the large sample limit under the assumption specified above, i.e., $r< 1+\frac{\xi}{d}$. However, after data standardization, this assumption may no longer hold, i.e., the noise ratio becomes
$r'=\frac{\max(\sigma_1^2/\operatorname{Var}(X_1), \dots, \sigma_d^2/\operatorname{Var}(X_d))}{\min(\sigma_1^2/\operatorname{Var}(X_1), \dots, \sigma_d^2/\operatorname{Var}(X_d))}$,
and there is no guarantee that $r'< 1+\frac{\xi'}{d}$ holds. In this case, Theorem~9 of \citet{Loh2014high}  does not apply and we are left with no guarantee that NOTEARS-EV or least-squares-based methods can find the true structure. Therefore, they may not perform well after data standardization, and the observation by \citet{Reisach2021beware} may not be surprising.\looseness=-1

In fact, after data standardization, the data may be far from having equal noise variances, i.e., the noise ratio $r'$ may be very large. An example is provided in Figure \ref{fig:standardized_ratio_variance}, in which we conduct $1000$ random simulations on linear Gaussian-EV model and Erd\"{o}s--R\'{e}nyi \citep{Erdos1959random} graphs with varying number of variables and degrees, and further compute the noise ratio in the large sample limit after standardizing the data. One observes that a larger number of variables and degree leads to a higher noise ratio $r'$; in this setting considered, the noise ratio appears to increase approximately exponentially in the degree. For example, for $50$-node graphs with degree of $4$ (one of the settings considered by \citet{Reisach2021beware}), the noise ratio  $r'$ is $2079.30\pm 189.6$, which is far from the equal noise variances case.

With such a large noise ratio $r'$ in the scenarios above, the guarantee \citep[Theorem~9]{Loh2014high} of least squares may not apply after standardization. In other words, least-squares-based search methods, whether it be continuous or discrete ones, are not expected to perform well after standardizing the data.\looseness=-1

\paragraph{Empirical studies.}
To validate the argument above, we compare the performance of various structure learning methods that assume equal noise variances on different noise ratios $r$. Following previous work \citep{Zheng2018notears,Ng2020role}, we simulate Erd\"{o}s--R\'{e}nyi \citep{Erdos1959random} graphs with $kd$ edges, denoted as ER$k$. Unless otherwise stated, each edge weight in the true weighted adjacency matrix $B$ is sampled uniformly at random from $[-2,-0.5]\cup [0.5,2]$. We consider ER1 graphs with $d\in\{15, 50\}$ variables and $n\in\{100, 10^6\}$ samples. We use Gaussian noises where the variances of two randomly chosen noise variables are set to $1$ and $r$, respectively, while the variances of the remaining noise variables are sampled uniformly at random from $[1, r]$. The noise ratios considered are $r\in\{1,4,16,64,256,1024\}$.
Apart from continuous approaches, i.e., NOTEARS-EV and GOLEM-EV, we also consider discrete ones, including a greedy approach, GDS~\citep{Peters2013identifiability} and an exact approach, A* \citep{Yuan2013learning}. For both discrete approaches, we adopt least squares score. We report the structural Hamming distance (SHD), F1 score, and recall over $30$ simulations, as well as their standard errors. Further details about the approaches and evaluation metrics are provided in Appendix~\ref{app:implementation_details}.

\begin{figure*}[!t]
	\minipage{0.32\textwidth}
	\includegraphics[width=\linewidth]{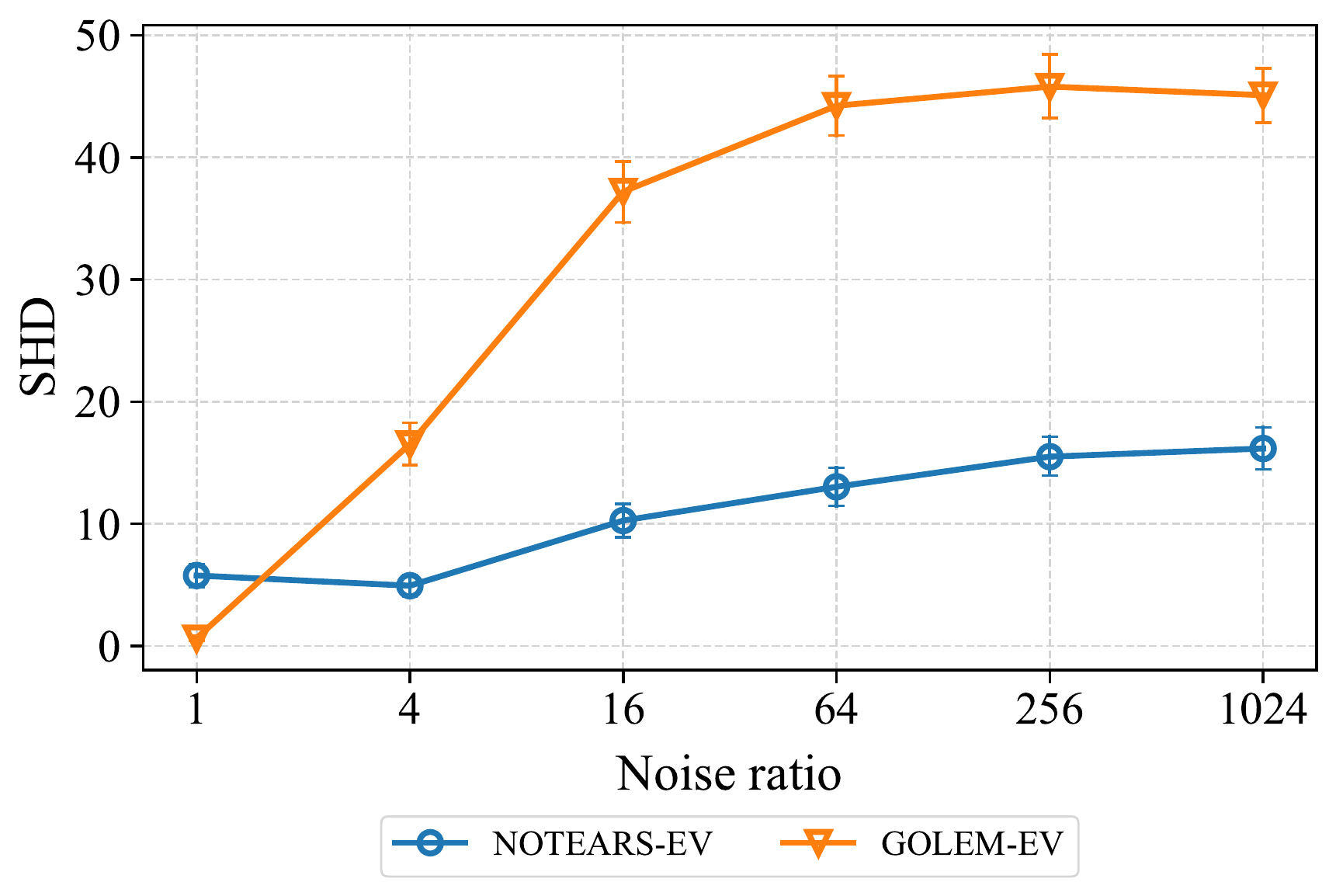}
 \caption{\small Linear Gaussian-EV formulation without standardization.}\label{fig:varsortability_ev_case_shd}
	\endminipage\hfill
	\minipage{0.32\textwidth}
	\includegraphics[width=\linewidth]{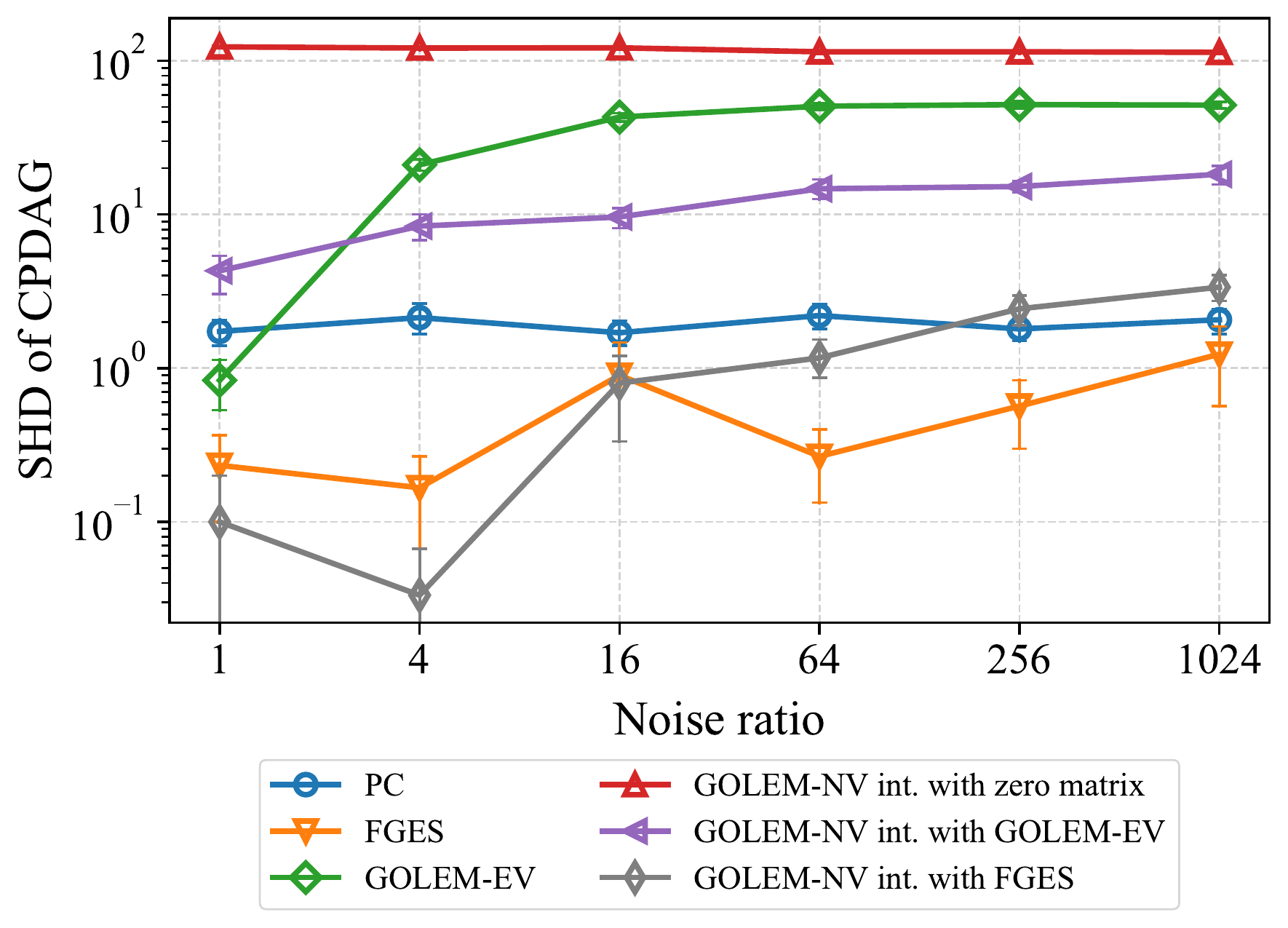}
 \caption{\small Linear Gaussian-NV formulation without standardization.}\label{fig:varsortability_nv_case_golem_shd_cpdag_unstandardized}
	\endminipage\hfill
	\minipage{0.32\textwidth}
	\includegraphics[width=\linewidth]{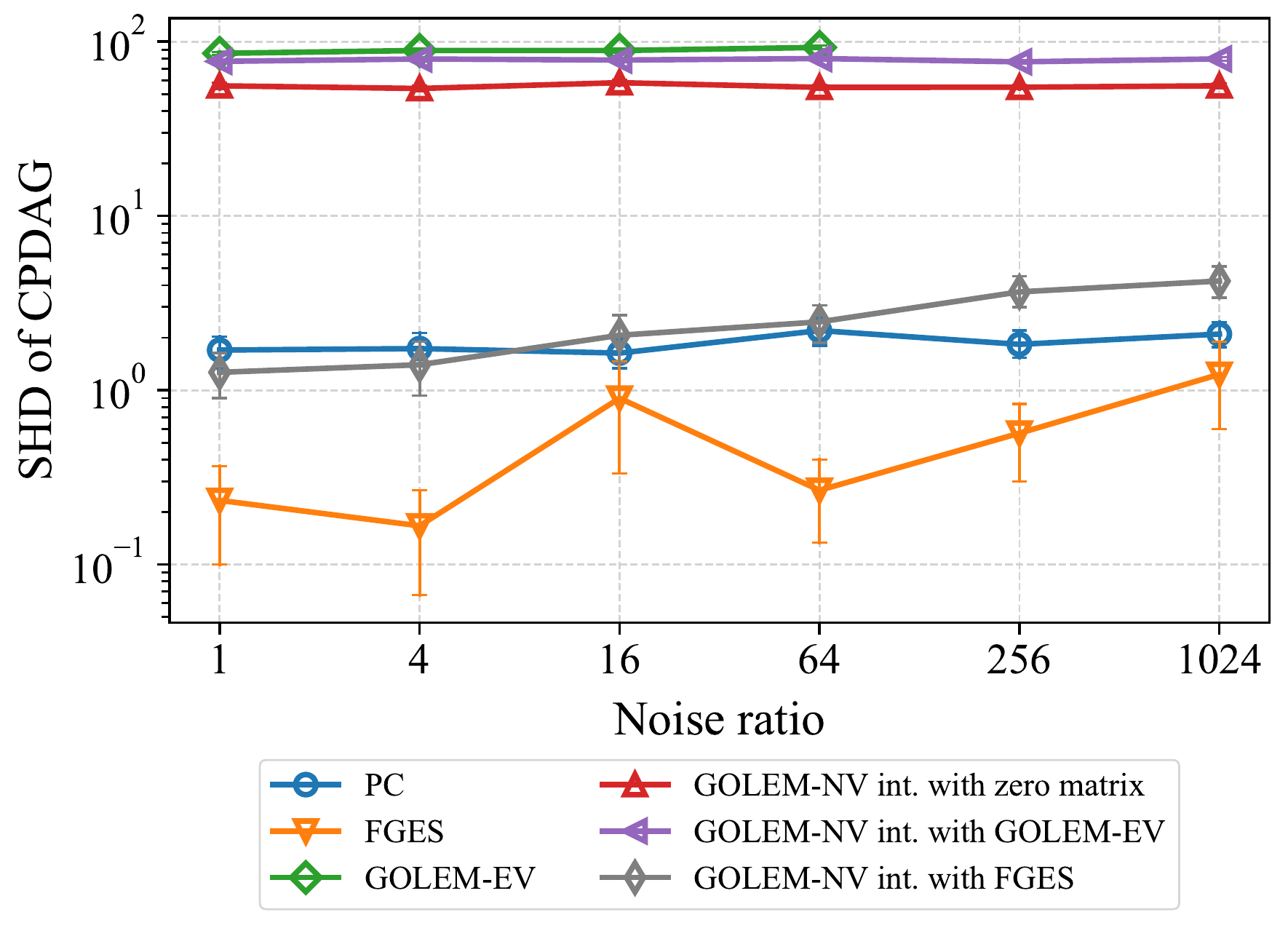}
 \caption{\small Linear Gaussian-NV formulation with standardization.}\label{fig:varsortability_nv_case_golem_shd_cpdag_standardized}
	\endminipage
 \vspace{-0.2em}
\end{figure*}

The results for $d=50$ and $n=10^6$ are shown in Figure \ref{fig:varsortability_ev_case_shd}, while complete results are available in Figure \ref{fig:varsortability_ev_case_result} in Appendix \ref{app:varsortability_ev_case}. We do not report the results of A* and GDS for $50$ variables because of the long running time. As the noise ratio increases, the performance of all four methods deteriorates in all cases, including NOTEARS-EV and GOLEM-EV. 
This demonstrates that, even with a large sample size, these methods that assume equal noise variances are not expected to perform well when the noise ratio $r$ is large. As shown in Figure~\ref{fig:standardized_ratio_variance}, the noise ratio after data standardization $r'$ is very large, e.g., could be as large as $10^6$, which is much larger than that considered in our experiment here. Thus, it may not be surprising that the structure learning methods considered do not perform well.

\subsection{With Non-Equal Noise Variances Formulation}\label{sec:varsortability_nv_case}
We next turn to the general linear Gaussian case, i.e., the non-equal noise variances formulation. We provide two arguments to explain why Statement \ref{statement:varsortability_nv_case} may not hold.

\paragraph{Argument 1.} \emph{GOLEM-NV often does not perform well in the presence of high varsortability.}
We demonstrate that GOLEM-NV with the standard initialization scheme (i.e., zero matrix) does not perform well in the presence of high varsortability, by comparing its estimated completed partially DAGs (CPDAGs) to standard methods such as PC \citep{Spirtes1991pc} and FGES \citep{Ramsey2017million}. Note that GOLEM-NV outputs a DAG, and thus an additional step is needed to convert it into a CPDAG that represents the MEC. We consider linear Gaussian models with $50$ variables and ER1 graphs, where the noise ratio $r$ is $2$. The sample size is set to $n=10^6$ to reduce finite-sample errors. Here, the varsortability is $0.97\pm 0.003$, which is consistent with the high varsortability reported by \citet[Section~3.3]
{Reisach2021beware}. \looseness=-1

\begin{wraptable}{r}{0.51\textwidth}
\vspace{-0.9em}
	\centering
	\caption{Empirical results.}
 \vspace{-0.4em}
 {\small
	\begin{tabular}{rccccccccccc} 
	\toprule
		~ & SHD of CPDAG  & F1 of skeleton\\
		\midrule
		GOLEM-NV & $120.27\pm 2.91$ & $0.50\pm 0.00$ \\
            PC & $1.57\pm 0.35$ & $0.99\pm 0.00$ \\ 
            FGES & $0.63\pm 0.47$ & $1.00\pm 0.00$ \\ 
		\bottomrule
    \end{tabular}
}
  \label{tab:varsortability_nv_case}
\end{wraptable}
The results in Table \ref{tab:varsortability_nv_case} show that PC and FGES can nearly recover the true CPDAG with a high F1 score, while GOLEM-NV has a low F1 score and high SHD. Note that the performance of GOLEM-NV is worse than an empty graph whose SHD is $50$. Therefore, GOLEM-NV does not perform well in the presence of high varsortability, and it seems that high varsortability does not help GOLEM-NV achieve a remarkable performance. The question is then why \citet{Ng2020role} observed a remarkable performance for GOLEM-NV on synthetic data as noticed by \citet{Reisach2021beware}, which we address next.

\paragraph{Argument 2.} \emph{The quality of the final solution estimated by GOLEM-NV depends largely on the initial solution possibly owing to nonconvexity.}
\citet[Section~4.1]{Ng2020role} adopted an initialization scheme for GOLEM-NV which uses the solution returned by GOLEM-EV as the initial solution, because the authors noticed that the optimization procedure of GOLEM-NV is prone to local solutions. Here, we experiment with different initialization schemes and find that they play an important role in the quality of the final solutions, possibly because of the nonconvex optimization formulation.

\vspace{-0.05em}
\paragraph{Empirical studies.}
We consider $50$-variable linear Gaussian model with ER1 graphs and noise ratios $r\in\{1,4,16,64,256,1024\}$. To focus on the aspect of nonconvex optimization, we use a large sample size, i.e., $n=10^6$, to reduce finite-sample errors. We compare various initialization schemes, including GOLEM-NV initialized with zero matrix, with the solution of GOLEM-EV, and with the solution of FGES\footnote{Since FGES outputs a CPDAG, we generate a DAG consistent with it following the procedure developed by \citet{Dor1992simple}, and compute the least squares coefficients for this specific DAG.}. We also provide empirical studies for GOLEM-NV with random initializations in Appendix \ref{sec:random_initialization}, and for GOLEM-EV with random perturbations in Appendix \ref{app:golem_ev_perturbation}.

The SHDs before and after standardization are available in Figures \ref{fig:varsortability_nv_case_golem_shd_cpdag_unstandardized} and \ref{fig:varsortability_nv_case_golem_shd_cpdag_standardized}, where the other metrics are available in Figure \ref{fig:varsortability_nv_case_result} in Appendix \ref{app:varsortability_nv_case}. Consistent with the results in Table \ref{tab:varsortability_nv_case}, GOLEM-NV initialized with zero matrix does not perform well across all settings. Moreover, before standardization, the performance of both GOLEM-EV and GOLEM-NV initialized with GOLEM-EV degrades as the noise ratio increases, and their gap with PC and FGES is also enlarged. A possible reason is that GOLEM-NV may be susceptible to suboptimal local solutions owing to nonconvexity, and thus its performance largely depends on the quality of solution of GOLEM-EV. As the noise ratio increases, the quality of solution returned by GOLEM-EV deteriorates, as discussed in Section \ref{sec:varsortability_ev_case}; therefore, GOLEM-NV does not have a decent initial solution. Similar explanation may also apply to its results after standardization, i.e., GOLEM-NV initialized with GOLEM-EV has a poor performance after standardization (also observed by \citet{Reisach2021beware}) because GOLEM-EV does not perform well owing to the large noise ratio $r'$, as discussed in Section~\ref{sec:varsortability_ev_case}. By contrast, GOLEM-NV initialized with the solution of FGES performs well in all cases (regardless of whether the data is standardized), where the results are close to FGES and PC. This is because FGES returns a solution with high quality across different noise ratio $r$, which serves as a good initial solution for optimization of GOLEM-NV.\looseness=-1

Note that the formulation of GOLEM-NV in Eq. \eqref{eq:golem} involves only matrix $B$ because it profiles out the parameter $\Omega$ (i.e., the noise variances) \citep[Appendix~C.1]{Ng2020role}. In Appendix \ref{sec:alternative_likelihood}, we further consider the likelihood function without profiling out such parameter and observe that it also leads to a poor performance when initialized with zero matrix, similar to the original version of GOLEM-NV. 

Instead of high varsortability, the observations above suggest that the performance of GOLEM-NV largely depends on the initial solution. A possible reason is that the optimization problem of GOLEM-NV may be highly nonconvex and contain many suboptimal local solutions, because (1) a large sample size $n=10^6$ is used which leads to small finite-sample errors, and (2) GOLEM-NV has been shown to be consistent under mild conditions, assuming that the global minimizer can be found and $\ell_0$ penalty is used (note that \citet{Ng2020role} did not provide guarantee for $\ell_1$ penalty).\footnote{With global minimizer and $\ell_0$ penalty, the theoretical guarantee of GOLEM-NV \citep[Theorem~2]{Ng2020role} holds in the large sample limit under the specified conditions, regardless of the initial solution, varsortability, and data standardization.} This is in contrast with the observation by \citet[Section~5.3]{Zheng2018notears} for the equal noise variances formulation, which finds that the solution of NOTEARS-EV is in practice close to the global minimizer. We provide further analysis on the nonconvexity of GOLEM-NV in Appendix \ref{app:analysis_score_and_shd}. Moreover, the above empirical studies may explain the remarkable performance of GOLEM-NV in the synthetic data experiments conducted by \citet{Ng2020role}, as noticed by \citet{Reisach2021beware}, which simulate linear Gaussian model with a relatively small noise ratio $r\leq 4$. In this case, GOLEM-EV achieves a decent performance, and using its solution to initialize GOLEM-NV leads to a remarkable performance.\looseness=-1

\section{Nonconvexity of Non-Equal Noise Variances Formulation}\label{sec:nonconvexity}
The empirical studies in Section \ref{sec:varsortability_nv_case} suggest that continuous structure learning approach that assumes non-equal noise variances, specifically GOLEM-NV, is prone to suboptimal local solutions, possibly owing to nonconvexity. On the other hand, many existing works \citep{Zheng2018notears,Yu2021nocurl,Charpentier2022differentiable,Bello2022dagma} considered the equal noise variances formulation (by adopting least squares objective) and observed remarkable performance. This indicates that the optimization procedure of non-equal noise variances formulation may in practice be more susceptible to suboptimal local solutions than the equal noise variances formulation, and thus empirically suggests that the nonconvexity issue might be more severe for the former formulation. In this section, we demonstrate that the recent advances for the equal noise variances formulation fail to achieve improvement for the non-equal noise variances formulation, thus suggesting that nonconvexity may be a key concern for the latter formulation. In particular, we consider recently developed search strategies and DAG constraints in Sections \ref{sec:search_strategies} and \ref{sec:dag_constraints}, respectively, and other cases in Section \ref{sec:other_cases}, namely linear non-Gaussian and nonlinear cases.

\subsection{Search Strategies}\label{sec:search_strategies}
We investigate whether different search strategies perform well for the non-equal noise variances formulation, namely NOTEARS \citep{Zheng2018notears},  DAGMA \citep{Bello2022dagma}, NOCURL~\citep{Yu2021nocurl}, DPDAG \citep{Charpentier2022differentiable}, and GOLEM \citep{Ng2020role}. These methods represent different strategies to traverse the search space for estimating a DAG with continuous optimization; a summary of them is provided in Appendix \ref{app:search_strategies_details}. For these methods except GOLEM, we replace their least squares objective $\ell(B;\mathbf{X})$ with $\mathcal{L}_\text{NV}(B;\mathbf{X})$, which corresponds to the likelihood of linear Gaussian DAGs assuming non-equal noise variances \citep[Appendix~C.1]{Ng2020role} (note that we omit the term $- \log|\det(I - B)|$ used by GOLEM-NV as it vanishes when $B$ represents a DAG).
We denote the resulting methods by NOTEARS-NV, DAGMA-NV, NOCURL-NV, and DPDAG-NV, respectively. We consider linear Gaussian model with $50$-node ER1 graphs and noise ratio $r$ of $16$, as well as sample sizes $n\in\{10^2,10^3,10^4,10^5,10^6\}$.

\begin{figure*}[!t]
	\minipage{0.33\textwidth}
 	\includegraphics[width=\linewidth]{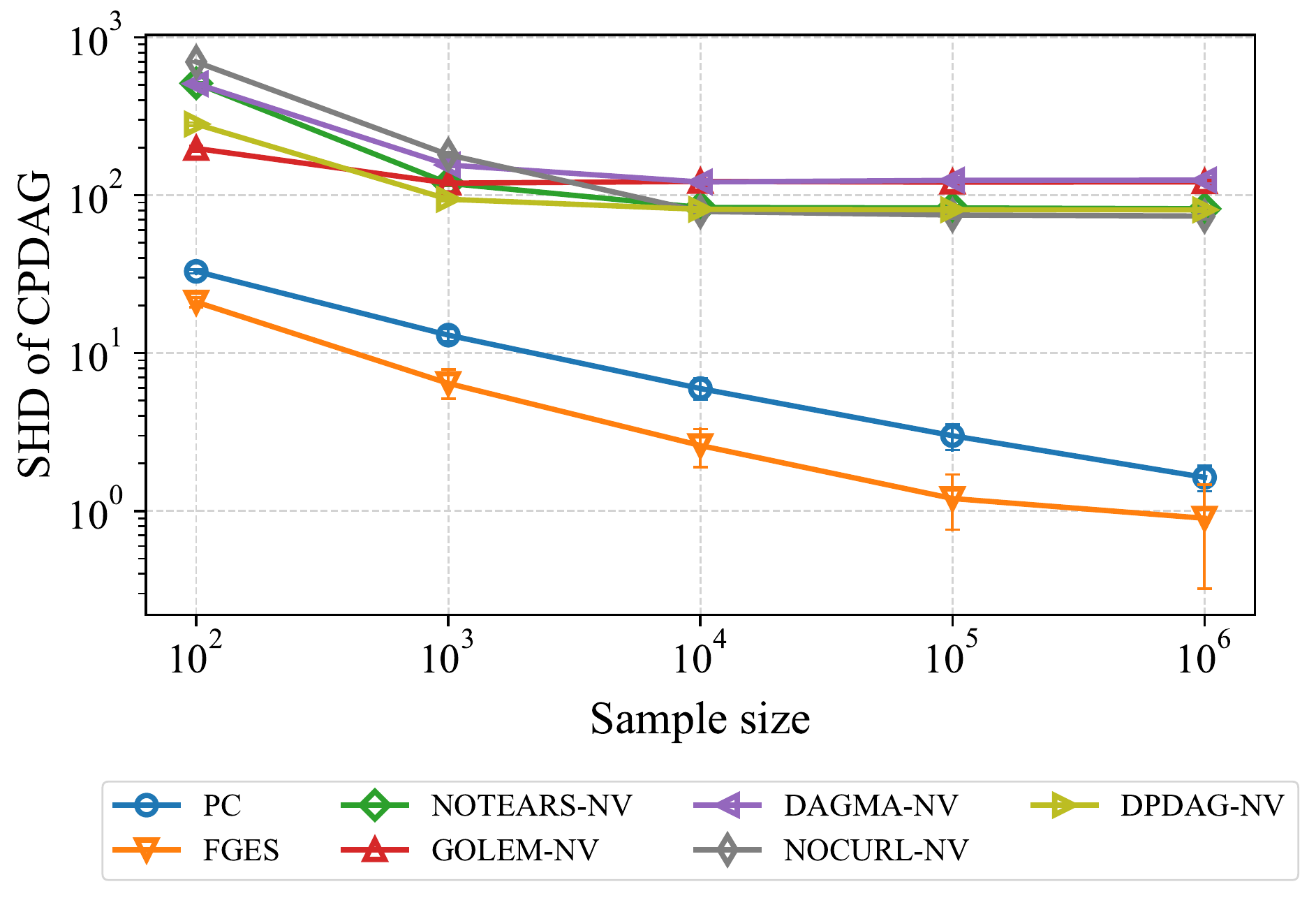}
 \caption{\small Search strategies.}\label{fig:search_strategies_shd_cpdag}
	\endminipage\hfill
	\minipage{0.33\textwidth}
	\includegraphics[width=\linewidth]{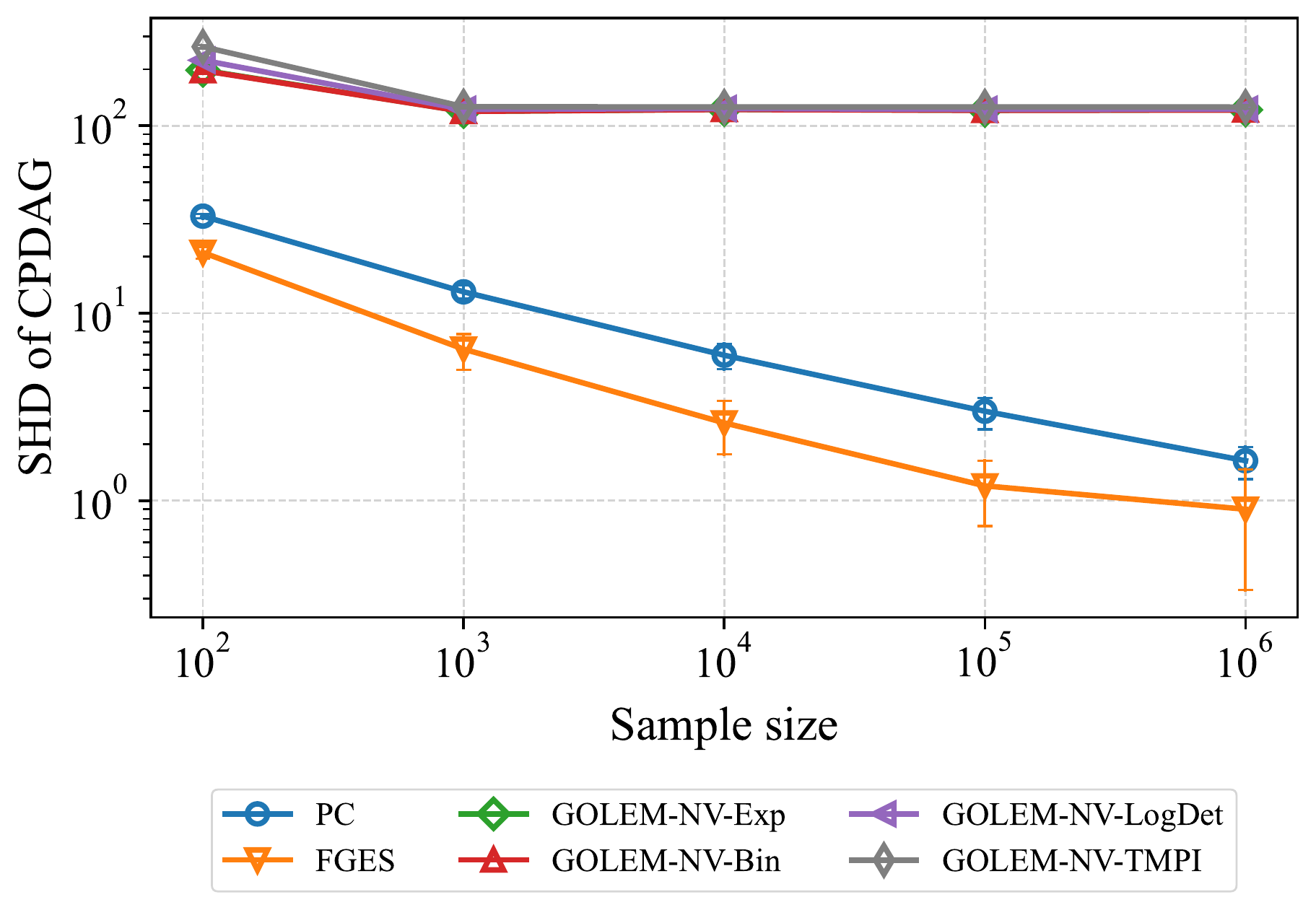}
 \caption{\small DAG constraints.}\label{fig:dag_constraints_golem_shd_cpdag}
	\endminipage\hfill
	\minipage{0.34\textwidth}
	\includegraphics[width=\linewidth]{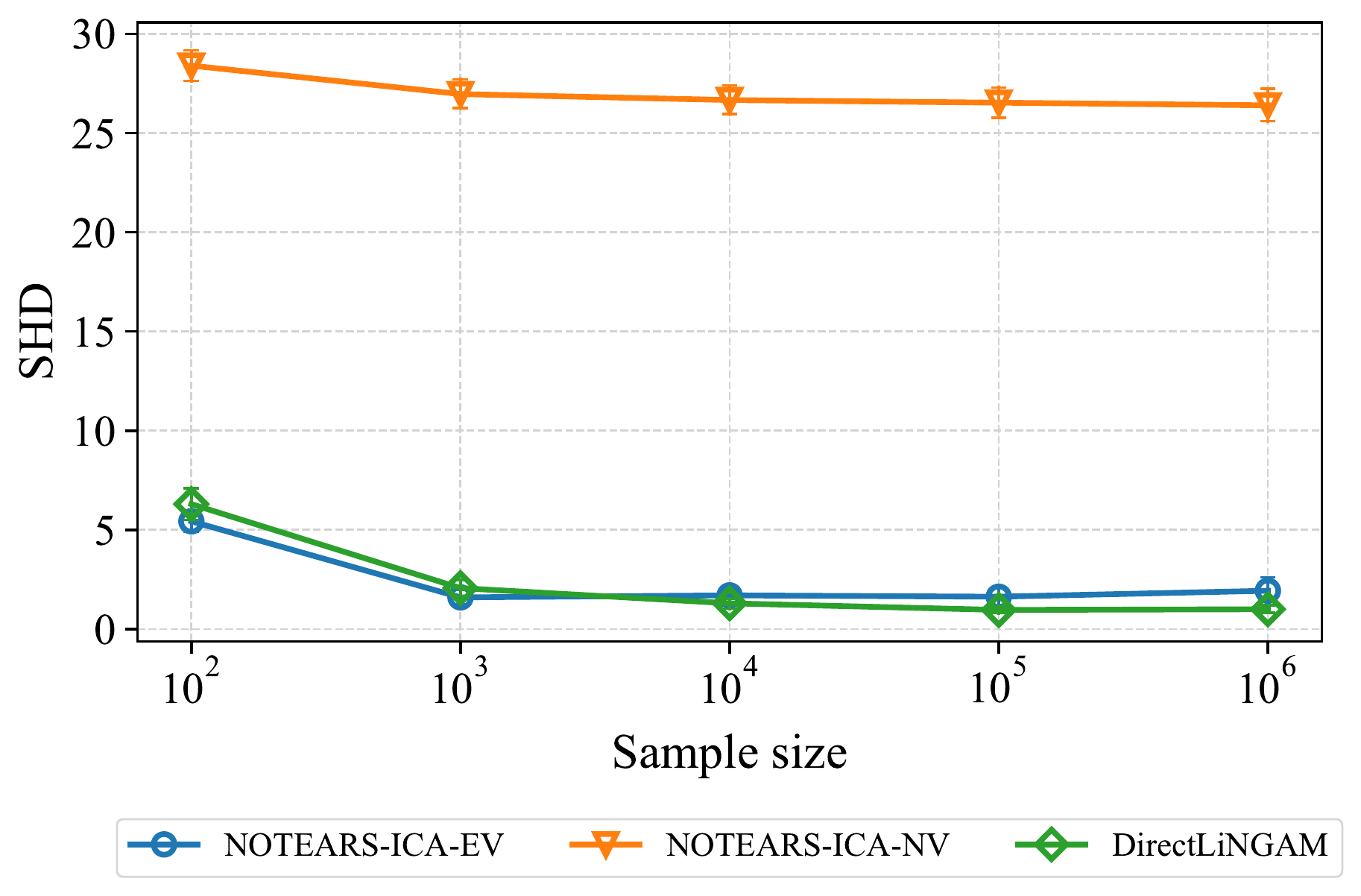}
 \caption{\small Linear non-Gaussian case.}\label{fig:notears_ica_shd}
	\endminipage
\end{figure*}

The SHDs computed over CPDAGs are available in Figure \ref{fig:search_strategies_shd_cpdag}, where the other metrics are available in Figure \ref{fig:search_strategies_result} in Appendix \ref{app:search_strategies}. It is observed that none of the continuous structure learning approaches perform well compared to the discrete approaches, i.e., PC and FGES. Note that PC and FGES improve with a larger sample size and eventually reach SHDs of $5.97\pm 0.43$ and $5.13\pm 0.51$, respectively. By contrast, the quality of solution estimated by continuous structure learning approaches does not improve much with a larger sample size and reaches a performance plateau after $n=10^4$. Specifically, the SHD of these continuous approaches is larger than $70$ even when the sample size is $n=10^6$. Note that an empty graph has an SHD of $50$. A possible reason is that the optimization problem of these continuous approaches may be highly
nonconvex and contain many suboptimal local solutions, thus leading to estimated CPDAGs that are far from the true ones.

\subsection{DAG Constraints}\label{sec:dag_constraints}
We consider the DAG constraints commonly adopted by continuous structure learning approaches based on matrix exponential \citep{Zheng2018notears} and binomial \citep{Yu19daggnn}, as well as more recent DAG constraints based on log-determinant \citep{Bello2022dagma} and truncated matrix power iteration \citep{Zhang2022truncated} that have been demonstrated to be less sensitive to the gradient vanishing issue encountered by the former two constraints. We apply these four constraints to GOLEM-NV, denoted as GOLEM-NV-Exp, GOLEM-NV-Bin, GOLEM-NV-LogDet, and GOLEM-NV-TMPI, respectively, and 
 similarly for NOTEARS-NV. 

Here we use the same setup of simulated data as that of Section \ref{sec:search_strategies}.
The SHDs of GOLEM-NV are depicted in Figure \ref{fig:dag_constraints_golem_shd_cpdag}, while complete results including those for NOTEARS-NV are available in Figure \ref{fig:dag_constraints_result} in Appendix \ref{app:dag_constraints}. Similar to the observation in Section \ref{sec:search_strategies}, one observes that GOLEM-NV and NOTEARS-NV equipped with these DAG constraints quickly reach a performance plateau and do not improve much with increasing sample sizes. Specifically, the SHDs of GOLEM-NV and NOTEARS-NV with these DAG constraints are larger than $115$ and $75$, respectively. Note that \citet{Bello2022dagma,Zhang2022truncated} observed that their proposed DAG constraints lead to a noticeable improvement over those existing ones \citep{Zheng2018notears,Yu19daggnn} for the equal noise variances formulation. Clearly, this experiment demonstrates that such an improvement cannot be straightforwardly translated into the non-equal noise variances formulation, at least in the settings considered here. Moreover, this suggests that the optimization problem of the non-equal noise variances formulation might be more challenging than the equal noise variances formulation as the former may contain more suboptimal local solutions.\looseness=-1

\subsection{Other Settings}\label{sec:other_cases}
We now investigate whether the observations in the previous linear Gaussian experiments also hold in the other settings, namely linear non-Gaussian \citep{Shimizu2006lingam} and nonlinear \citep{Hoyer2009nonlinear} cases.

\paragraph{Linear non-Gaussian case.}
\citet{Zheng2020thesis} developed an extension of NOTEARS to handle linear non-Gaussian case, called NOTEARS-ICA. It is worth noting that \citet[Section~4.3.2]{Zheng2020thesis} provided a general formulation in the non-equal noise variances case; as the same time, their experiments, as discussed by \citet[Section~4.5]{Zheng2020thesis}, seem to adopt the equal noise variances formulation. Here, we consider both equal and non-equal noise variances formulation for NOTEARS-ICA, denoted as NOTEARS-ICA-EV and NOTEARS-ICA-NV, respectively. We simulate $15$-node ER1 graphs and linear SEMs with standard Laplace noises. DirectLiNGAM \citep{Shimizu2011directlingam,Ikeuchi2023python} is also compared here as a baseline. The SHDs are reported in Figure \ref{fig:notears_ica_shd}, while complete results are shown in Figure \ref{fig:notears_ica_result} in Appendix \ref{app:other_cases}. We observe that NOTEARS-ICA-EV performs similarly to DirectLiNGAM, both of which improve with more samples; specifically, their SHDs are very close to zero when the sample size is large. However, the performance of NOTEARS-ICA-NV does not improve much even when the sample size is large. For example, when $n=10^6$, NOTEARS-ICA-EV and DirectLiNGAM have SHDs of $1.93\pm 0.59$ and $1.00\pm 0.19$, respectively, while NOTEARS-ICA-NV has SHD of $26.40\pm 0.80$. Note that an empty graph leads to an SHD of~$15$.

\paragraph{Nonlinear case.}
There are several nonlinear extensions of NOTEARS, some of which rely on pre/post-processing steps such as pruning \citep{Lachapelle2020grandag, Ng2022masked}, making it hard to study the performance of continuous optimization procedure. We consider NOTEARS-MLP \citep{Zheng2020learning} that does not rely on such pre/post-processing steps, and extend it to the non-equal noise variances formulation by replacing its least squares objective with $\mathcal{L}_\text{NV}(B,\mathbf{X})$, in which the residuals are computed with multilayer perceptrons (MLPs) instead of linear regressions. We adopt the data generating procedure used by \citet{Zheng2020learning} with MLPs and standard Gaussian noises. The results of $15$-node ER1 graphs are provided in Figure~\ref{fig:notears_mlp_result} in Appendix \ref{app:other_cases}. Similar to the observation in the linear non-Gaussian case, NOTEARS-MLP-EV achieves a noticeably lower SHD when the sample size is large, while NOTEARS-MLP-NV does not improve much and its SHD remains at $10.07\pm 0.57$ even when the sample size is as large as $10^5$. \looseness=-1

\section{Thresholding and Sparsity Penalty}\label{sec:other_issues}
Apart from the possible nonconvexity issue discussed in Section \ref{sec:nonconvexity}, we show that the performance of continuous approaches is sensitive to other aspects of the search procedure, i.e., thresholding and sparsity. These technical issues should be made transparent, as they play an important role in the final solutions. In a concurrent work, \citet{Xu2022sparse} also studied the aspects of thresholding and sparsity in NOTEARS-EV. Here, our analysis of thresholding includes GOLEM-EV and discrete approaches such as A* and GDS, as well as both small ($n=100$) and large ($n=10^6$) sample sizes; therefore, the analysis and observations are different. For sparsity, \citet{Xu2022sparse} adopted the adaptive $\ell_1$ penalty~\citep{Zou2006adaptive} for NOTEARS-EV, while we consider the smoothly clipped absolute deviation (SCAD) penalty~\citep{Fan2001variable} and minimax concave penalty (MCP) \citep{Zhang2010nearly} for NOTEARS-EV and GOLEM-EV.\looseness=-1

\subsection{Thresholding}\label{sec:thresholding}

\begin{wrapfigure}{r}{0.49\textwidth}
\vspace{-2.9em}
  \begin{center}
    \includegraphics[width=0.48\textwidth]{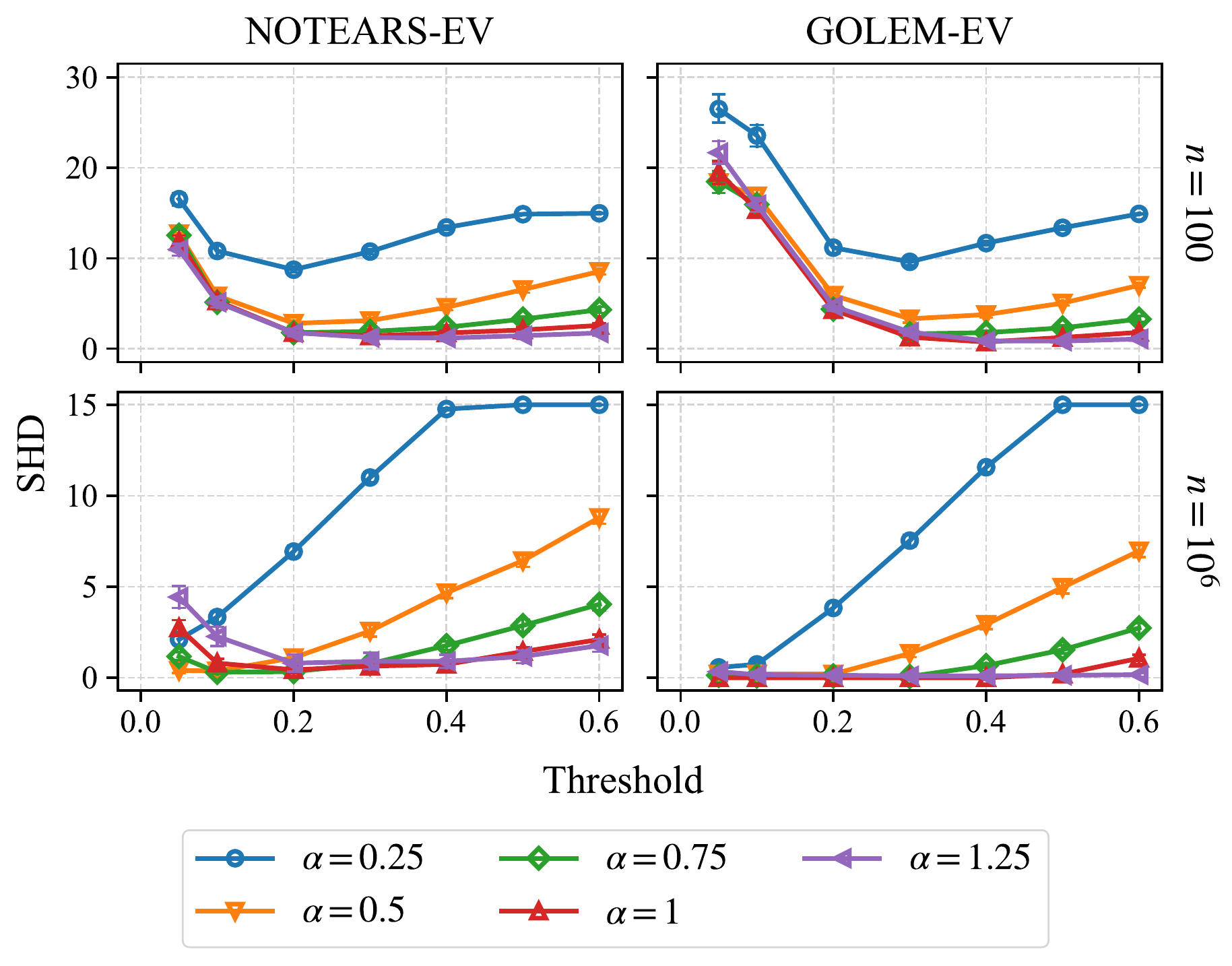}
  \end{center}
  \vspace{-1em}
  \caption{SHDs of different thresholds and $15$ variables under different weight scales.}
  \label{fig:thresholding_shd}
\vspace{-0.4em}
\end{wrapfigure}

We investigate the impact of threshold specification on continuous structure learning, and show that it plays a crucial role for the final solutions. Existing continuous approaches~\citep{Zheng2018notears,Ng2020role,Yu2021nocurl,Bello2022dagma} typically use edge weights sampled uniformly at random from $[-2, -0.5] \cup [0.5, 2]$ for the true weighted adjacency matrix and apply a threshold of $0.3$ to identify the final structure. Here, we consider $15$-node ER1 graphs with different ranges of edge weights, i.e., $[-2\alpha, -0.5\alpha] \cup [0.5\alpha, 2\alpha]$ where $\alpha\in\{0.25,0.5,0.75,1,1.25\}$, and sample sizes $n\in\{100,10^6\}$. Apart from continuous approaches, i.e., NOTEARS-EV and GOLEM-EV, we also report results for discrete approaches A* and GDS. The thresholds considered are $0.05, 0.1, 0.2, 0.3, 0.4, 0.5$, and $0.6$. Note that A* and GDS output a DAG, and thus we apply thresholding on its least squares coefficients. The SHDs are provided in Figure \ref{fig:thresholding_shd}, while complete results for F1 score and recall are given in Figure \ref{fig:thresholding_result} in Appendix \ref{app:thresholding}. The observations are described below.\looseness=-1

\paragraph{Observation 1.} \emph{When $n=100$, a relatively large threshold (e.g., $0.3$) is beneficial for all methods considered.} With a small sample size, the estimated DAGs by both continuous and discrete approaches may contain many false discoveries owing to finite-sample errors, and a relatively large threshold helps remove some of them. This seems to increase the precision at the cost of decreasing the recall.

\paragraph{Observation 2.} \emph{When $n=10^6$, using a relatively large threshold (e.g., 0.3) is harmful in many cases.}
In this case, using a threshold of $0.3$ increases the SHDs of all methods for $\alpha=0.25,0.5$, since the recall is decreased. A possible reason is that the threshold applied is larger than many edge weights in the true weighted adjacency matrix. Therefore, the fixed threshold of $0.3$ used in existing works may be undesirable in many cases, as it may wrongly remove many edges and lead to a low recall. To avoid such a case, we adopt a relatively small threshold (e.g., $0.1$) in the other experiments.

\paragraph{Observation 3.} \emph{When $\alpha=1$, the optimal thresholds for NOTEARS-EV are $0.3$ and $0.2$ for $n=100,10^6$, respectively.} Specifically, its SHDs are $1.40\pm 0.31$ and $0.43\pm 0.20$ for $n=100,10^6$, respectively. For $n=10^6$, the second best threshold for NOTEARS-EV is $0.3$, leading to an SHD of $0.63\pm 0.23$. Moreover, the optimal thresholds for NOTEARS-EV may be different across various settings. Since the empirical studies by \citet{Zheng2018notears} and several follow-up works adopt $\alpha=1$ and a threshold of $0.3$, this indicates that their comparison with existing baselines, particularly discrete approaches, may not be completely fair as it may implicitly assume prior knowledge about the range of edge weights in the data generating procedure.

\paragraph{Observation 4.} \emph{When $n=10^6$, NOTEARS-EV requires a relatively large threshold (e.g., $0.2$) to perform well for $\alpha=1,1.25$, while a small threshold (i.e., $0.05$) is sufficient for GOLEM-EV for all $\alpha$ considered.} Specifically, with a threshold of $0.05$, GOLEM-EV achieves SHDs of $0.57\pm 0.18$, $0.20\pm 0.10$, $0.13\pm0.10$, $0 \pm 0$, and $0.33 \pm 0.18$ for $\alpha=0.25, 0.5, 0.75, 1, 1.25$, respectively, indicating that the estimated DAGs by GOLEM-EV are very close to ground truths. Since the sample size $n=10^6$ used here is large and leads to considerably small finite-sample errors, this appears to suggest that the nonconvex landscape of GOLEM-EV may contain fewer suboptimal local solutions in this setting; thus, when initialized with zero matrix, continuous optimization procedure reaches a solution close to the ground truth. By contrast, this is not the case of NOTEARS-EV, which, with a threshold of $0.05$, achieves SHDs of $2.10\pm0.28$, $0.40\pm0.14$, $1.17 \pm 0.23$, $2.73 \pm 0.40$, and $4.43\pm0.61$ for $\alpha=0.25, 0.5, 0.75, 1, 1.25$, respectively. For $\alpha=1,1.25$, a threshold of $0.2$ (or $0.3$) results in a better performance (i.e., $0.43 \pm 0.20$ and $0.80 \pm 0.47$, respectively) for NOTEARS-EV. This indicates that it may be more likely for NOTEARS-EV to return suboptimal local solutions as compared to GOLEM-EV, at least in this setting considered with a large sample size. This observation also suggests that, in addition to reducing false discoveries resulting from finite-sample errors, \emph{thresholding may help reduce false discoveries caused by nonconvexity}, especially for NOTEARS-EV.\looseness=-1

It is clear from Observations 1, 2, 3, and 4 that different approaches have different optimal thresholds in different settings. Therefore, a possible direction is to develop a general procedure (e.g., adaptive thresholding) that can be reliably applied in different settings to identify edges from the solutions of continuous structure learning approaches. As discussed above, this may also help reduce false discoveries resulting from both finite-sample errors and nonconvexity (see Observation 4).

\subsection{Sparsity Penalty}\label{sec:sparsity}

\begin{wrapfigure}{r}{0.46\textwidth}
\vspace{-3em}
  \begin{center}
    \includegraphics[width=0.47\textwidth]{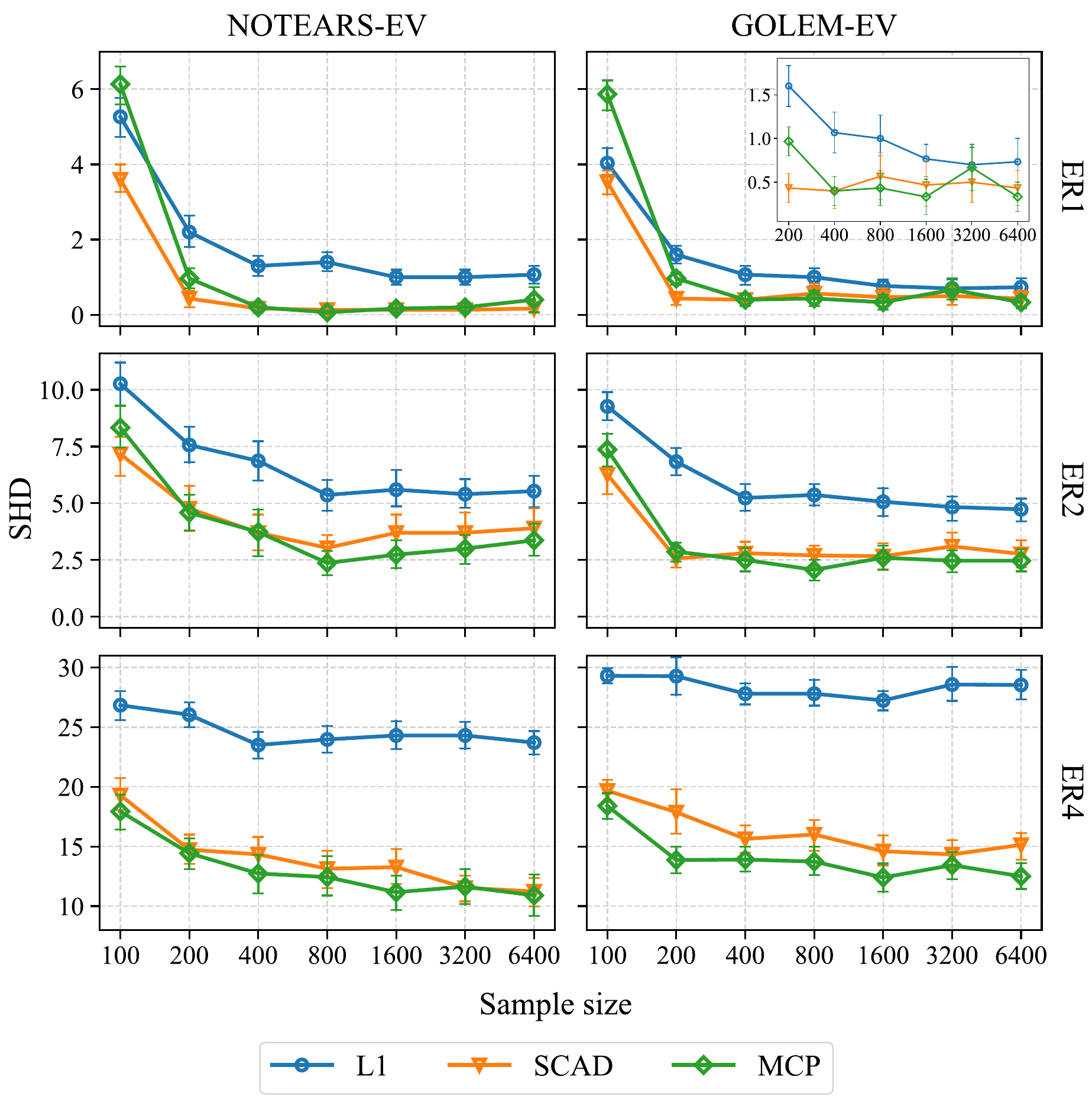}
  \end{center}
  \vspace{-0.8em}
  \caption{SHDs of different sparsity penalties and $15$ variables under different sample sizes.}
  \label{fig:sparsity_15nodes_shd}
\vspace{-1em}
\end{wrapfigure}
Apart from nonconvexity, another key factor that differentiates continuous structure learning approaches from discrete search approaches (e.g., A* \citep{Yuan2013learning}, dynamic programming \citep{Singh2005finding}, GES, \citep{Chickering2002optimal}, and GDS \citep{Peters2013identifiability}) is that the latter use $\ell_0$ penalty, while the former adopt $\ell_1$ penalty, which is known to be biased and negatively affect the performance \citep{Fan2001variable,Breheny2011coordinate}. The reason is that it penalizes all coefficients with the same rate, including those with large values. Thus, we consider alternative forms of sparsity penalty that help overcome such an issue, namely the SCAD penalty \citep{Fan2001variable} and MCP \citep{Zhang2010nearly}, where less shrinkage is imposed to larger coefficients. Moreover, SCAD penalty and MCP do not require the incoherence condition for support recovery \citep{Loh2017support} that is needed by $\ell_1$ penalty in several statistical problems \citep{Wainwright2009sharp,Ravikumar2011high}, which might be a rather restrictive assumption in practice. This may also be the case for structure learning, e.g., \citet{Aragam2019globally} established high dimensional structure consistency of least squares in the space of acyclic weighted adjacency matrices (corresponding to the formulation of NOTEARS), for which the incoherence condition is not needed when using MCP.

We compare the performance of NOTEARS-EV and GOLEM-EV under specification of the sparsity penalties discussed above. We consider ER1, ER2, and ER4 graphs with graph sizes $d\in\{15,50\}$ and sample sizes $n\in\{100,200,400,800,1600,3200,6400\}$. The SHDs of $d=15$ are reported in Figure \ref{fig:sparsity_15nodes_shd}, and complete results are provided in Figure \ref{fig:sparsity_result} in Appendix \ref{app:sparsity}. In most settings, we observe that SCAD penalty and MCP achieve noticeable improvement over $\ell_1$ penalty for NOTEARS-EV and GOLEM-EV, especially when the degree is large. Moreover, SCAD penalty and MCP lead to similar performance in many settings.

%% file: sections/triangle_examples.tex
\begin{figure}
\hspace{1em}\begin{minipage}{0.45\textwidth}
\centering
\subfloat[Ground truth.]{
    \tikz{
        \node[latent] (x3) {$X_3$};
        \node[latent,above=of x3,xshift=-1cm] (x1) {$X_1$};
        \node[latent,above=of x3,xshift=1cm] (x2) {$X_2$};
        \edge {x2,x3} {x1}
        \edge {x3} {x2}
    }\label{fig:ground_truth_triangle}
}\quad
\subfloat[Incorrect structure.]{
    \tikz{
        \node[latent] (x3) {$X_3$};
        \node[latent,above=of x3,xshift=-1cm] (x1) {$X_1$};
        \node[latent,above=of x3,xshift=1cm] (x2) {$X_2$};
        \edge {x3} {x1}
        \edge {x1,x3} {x2}
    }\label{fig:alternative_triangle}
}
\caption{Examples of triangle structures.}
\label{fig:triangle_examples}
\end{minipage}\hfill
\begin{minipage}{0.47\textwidth}
\centering
\includegraphics[width=0.72\linewidth]{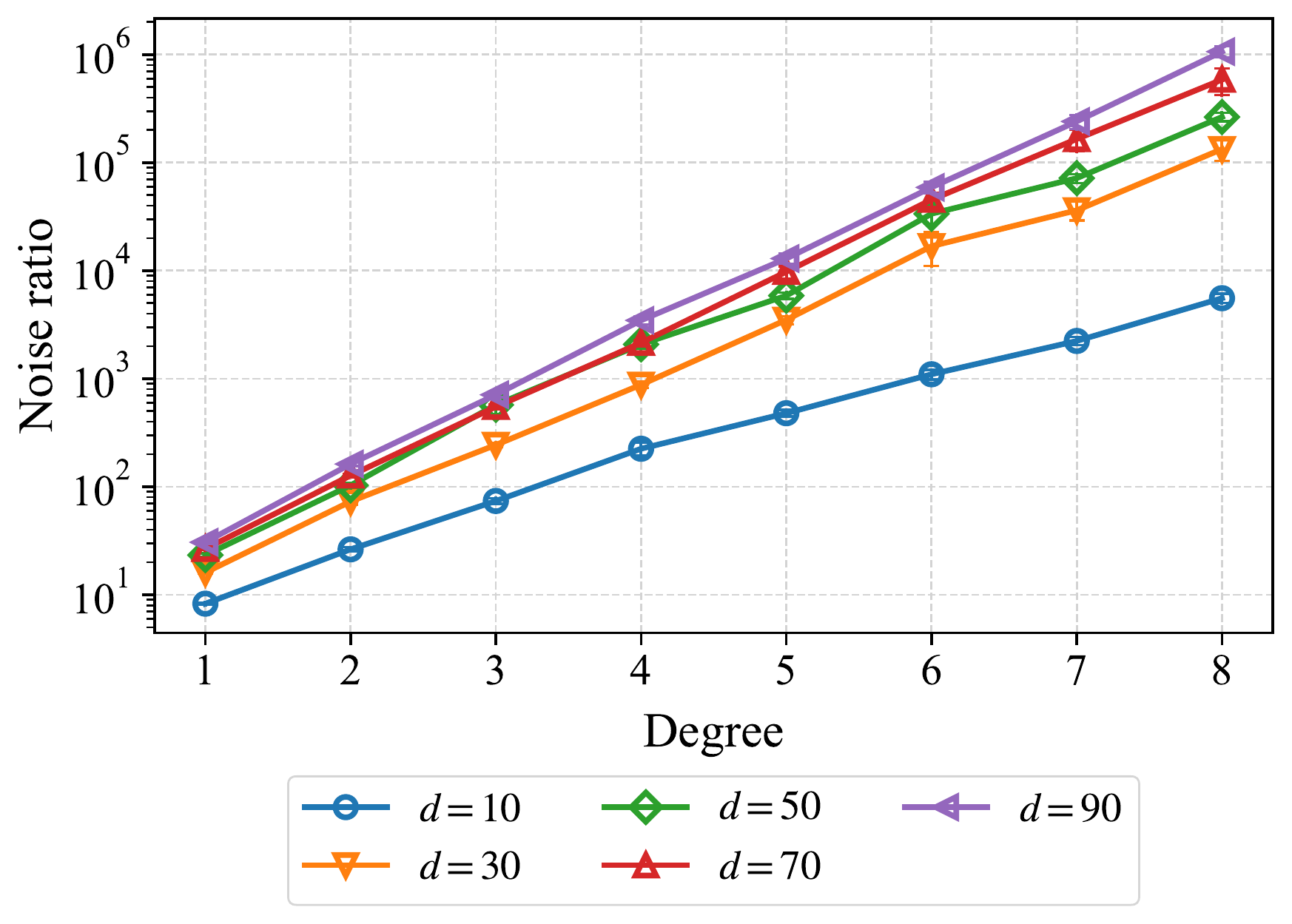}
\vspace{-0.6em}
\caption{Noise ratio after data standardization.}
\label{fig:standardized_ratio_variance}
\end{minipage}\hspace{1em}
\vspace{-0.45em}
\end{figure}

%% file: sections/4discussion.tex
\section{Conclusion and Discussion}\label{sec:discussion}
We investigate in which cases continuous structure learning approaches can and cannot perform well and why this happens. We focus on several aspects of the data and search procedure, including varsortability, data standardization, nonconvexity, thresholding, and sparsity. Despite the simplicity of continuous structure learning approaches, we demonstrate that they may suffer from various issues discussed below. Our goal here is not to resolve all these issues, but rather to analyze them, make them transparent, and suggest possible directions. We hope that our studies could stimulate future works on developing more reliable continuous structure learning approaches, e.g., to overcome the nonconvexity and thresholding issues. Detailed discussions are provided below.

\vspace{-0.1em}
\paragraph{Varsortability and data standardization.}
We show that the statements by \citet{Reisach2021beware} may not hold for formulations with equal and non-equal noise variances, and provide possible alternative explanations for the observations that continuous approaches do not perform well after data standardization \citep{Reisach2021beware}. For the equal noise variances formulation, we provide counterexamples to the statement, and demonstrate that the performance of continuous approaches assuming equal noise variances degrade as the noise ratio increases, which may explain why they do not perform well after standardization. For the non-equal noise variances formulation, we show that nonconvexity may be the reason why continuous approaches do not perform well both before and after data standardization.

\vspace{-0.1em}
\paragraph{Nonconvexity.}
Many existing works in continuous structure learning considered only the equal noise variances formulation by adopting least squares and observed remarkable performance. We demonstrate that these recent advances for the equal noise variances formulation fail to achieve improvement for the non-equal noise variances one. This implies that nonconvexity may be a main concern especially for the latter formulation. Here, the scores of the structures far from the ground truth could still be close to the best score. Our findings suggest that future works should take into account the non-equal noise variances formulation to handle more general settings and for a more comprehensive empirical evaluation, and develop reliable approaches that mitigate the nonconvexity issue.\looseness=-1

\vspace{-0.1em}
\paragraph{Thresholding.}
Our experiments indicate that the choice of threshold plays an important role in the final solutions, and that the optimal thresholds may differ across settings. Specifically, (1) using a relatively large threshold of $0.3$ as in existing works may be harmful and remove many true positives, especially when the true edge weights are small. (2) Thresholding may be beneficial in certain cases as it not only helps reduce false discoveries resulting from finite-sample errors, but may also help reduce those caused by nonconvexity, especially for NOTEARS. Thus, the choice of threshold should be treated with care. A possible future direction is to develop a general procedure
(e.g., adaptive thresholding) that can be applied in different settings to identify edges from the solutions of continuous structure learning approaches, while also reducing false discoveries.

\vspace{-0.1em}
\paragraph{Sparsity penalty.}
We demonstrate that sparsity penalty plays a crucial role for continuous structure learning approaches, e.g., NOTEARS and GOLEM. Specifically, $\ell_1$ penalty may not perform well possibly due to its bias, while other penalties such as SCAD penalty and MCP could help remedy it.\looseness=-1

%% file: sections/5appendix.tex
\section{Discussion of Statements by \citet{Reisach2021beware}}\label{app:discussion_statements}
We first list a few excerpts from \citet{Reisach2021beware} that involve the relationship between varsortability and the performance of continuous structure learning approaches:
\begin{itemize}
\item Abstract: ``the remarkable performance of some continuous structure learning algorithms can be explained by high varsortability''.
\item Section 1: ``Our experiments demonstrate that varsortability dominates the optimization and helps achieve state-of-the-art performance provided the ground-truth data scale''.
\item Section 3.4: ``We explain how varsortability may dominate the performance of continuous structure learning algorithms''.
\item Section 3.4: ``For this reason we focus on the first optimization steps to explain a) why continuous structure learning algorithms that assume equal noise variance work remarkably well in the presence of high varsortability''.
\item Section 4.3: ``the evidence corroborates our claim that the remarkable performance on raw data and the overall behavior upon standardization of the continuous structure learning algorithms may be driven primarily by high varsortability''.
\end{itemize}

For clarity and ease of further analysis, in Section \ref{sec:background_varsortability} we attempt to provide  a partial formulation of the statements (on which this work focuses), separated into two different cases. Specifically, we formulate Statements \ref{statement:varsortability_ev_case} and \ref{statement:varsortability_nv_case} by following the wordings of the fourth excerpt above because it has a relatively clear technical interpretation. Note that our analysis in Section \ref{sec:varsortability} also applies if the statements are formulated based on some different excerpts. For instance, we provide alternative formulations of the statements (corresponding to Statements \ref{statement:varsortability_ev_case} and \ref{statement:varsortability_nv_case}, respectively) that follow the wordings of the second excerpt above. 

\begin{statement}[Equal Noise Variances Formulation]\label{statement:varsortability_ev_case_alternative}
Varsortability dominates the optimization of continuous structure learning approaches that assume equal noise variances, specifically NOTEARS-EV and GOLEM-EV, and helps them achieve state-of-the-art performance provided the ground-truth data scale.
\end{statement}

\begin{statement}[Non-Equal Noise Variances Formulation]\label{statement:varsortability_nv_case_alternative}
Varsortability dominates the optimization of continuous structure learning approaches that assume non-equal noise variances, specifically GOLEM-NV, and helps them achieve state-of-the-art performance provided the ground-truth data scale.
\end{statement}

In particular, the examples in Section \ref{sec:varsortability_ev_case}, including Example \ref{example:single_counterexample} and Proposition \ref{proposition:nonzero_measure_counterexample}, as well as the corresponding empirical studies, demonstrate that a high varsortability does not help NOTEARS-EV and GOLEM-EV achieve a good performance, even when the ground-truth data scale is used. This is also the case for GOLEM-NV, as demonstrated in Section \ref{sec:varsortability_nv_case}. Furthermore, these studies suggest that varsortability may not dominate the optimization of these continuous structure learning approaches, at least in the settings considered in this work.

\section{Proofs of Main Results}
In this section, we provide the proofs of Propositions \ref{proposition:nonzero_measure_counterexample} and \ref{proposition:low_varsortability_but_correct_dag} given in Section \ref{sec:varsortability_ev_case}.
\subsection{Proof of Proposition \ref{proposition:nonzero_measure_counterexample}}
\label{app:proof_nonzero_measure_counterexample}
Let $(\tilde{B},\tilde{\Omega})$ be the parameters associated with the linear SEMs over variables $X=(X_1,\dots,X_d)$, $d\geq 3$, where the noise variables follow Gaussian distributions.\footnote{For clarity in this proof, we use $(\tilde{B},\tilde{\Omega})$ to denote the parameters instead of $(B,\Omega)$ as in the proposition statement.}
By definition, $\tilde{\Omega}$ is a positive diagonal matrix, while the weighted adjacency matrix $\tilde{B}$ corresponds to a DAG.

The maximum number of free parameters in matrix $\tilde{B}$ is $\frac{d(d-1)}{2}$, which corresponds to fully connected DAGs with various topological orders. Since there are only a finite number of topological orders with $d$ nodes, it suffices to consider fully connected DAGs with a specific topological order, e.g., where $X_{i+1}$ precedes $X_i$ for $i=1,\dots,d-1$. We denote by $\tilde{\G}$ the fully connected DAG with this specific topological order. In this case, the corresponding weighted adjacency matrix $\tilde{B}$ is a strictly lower triangular matrix with $\frac{d(d-1)}{2}$ free parameters.

Note that, in the large sample limit, the covariance matrix of $X$ is
\[\tilde{\Sigma} = (I-\tilde{B})^{-T} \tilde{\Omega} (I-\tilde{B})^{-1},\]
and the least squares loss of matrix $B$ is
\[
\ell(B; \tilde{\Sigma})=\frac{1}{2}\tr\big( (I-B)^T \tilde{\Sigma} (I-B) \big).
\]
Clearly, the varsortability equals one if
\begin{equation}\label{eq:proof_1_varsortability_one}
\operatorname{Var}(X_i)>\operatorname{Var}(X_{i+1})\quad \iff\quad \tilde{\Sigma}_{i,i}-\tilde{\Sigma}_{i+1,i+1}>0,\quad \text{for \ } i=1,\dots,d-1.
\end{equation}
It is straightforward to show that the least squares score of $\tilde{\G}$ is $\ell(\tilde{B}; \tilde{\Sigma})$ \citep[Lemma~6]{Loh2014high}, given by
\begin{equation}\label{eq:proof_1_least_squares_true_parameters}
\ell(\tilde{B}; \tilde{\Sigma}) = \frac{1}{2}\tr\big( (I-\tilde{B})^T \tilde{\Sigma} (I-\tilde{B}) \big)=\frac{1}{2}\tr(\tilde{\Omega}).
\end{equation}

Since matrices $I-\tilde{B}$ and $\tilde{\Omega}$ are invertible, $\tilde{\Sigma}$ is symmetric positive definite. Let $\tilde{L}$ be the (unique) Cholesky factor of $\tilde{\Sigma}$, i.e., 
\begin{equation}\label{eq:proof_1_cholesky_factor}
\tilde{L}\tilde{L}^T=\tilde{\Sigma},
\end{equation}
where $\tilde{L}$ is lower triangular with positive diagonal entries. We now define $(\hat{B},\hat{\Omega})$ as
\begin{equation}\label{eq:proof_1_alternative_parameters}
\hat{\Omega}\coloneqq\operatorname{diag}\left((\tilde{L}_{1,1})^2,\dots,(\tilde{L}_{d,d})^2\right) \quad\text{and}\quad \hat{B}\coloneqq I - \tilde{L}^{-T}\hat{\Omega}^{\frac{1}{2}},
\end{equation}
which implies
\begin{equation}\label{eq:proof_1_alternative_parameters_and_cholesky_factor}
(I-\hat{B})^{-T}\hat{\Omega}^{\frac{1}{2}}=\tilde{L}
\end{equation}
and that $\hat{B}$ is strictly upper triangular. By Eqs. \eqref{eq:proof_1_cholesky_factor} and \eqref{eq:proof_1_alternative_parameters_and_cholesky_factor}, we have
\begin{equation}\label{eq:proof_1_Omega_hat_and_population_cov}
(I-\hat{B})^T \tilde{\Sigma} (I-\hat{B})=\hat{\Omega}.
\end{equation}

Because $\hat{B}$ is strictly upper triangular, the DAG defined by $\hat{B}$ is different from $\tilde{\G}$ (i.e., the DAG defined by $\tilde{B}$). By Eqs. \eqref{eq:proof_1_alternative_parameters} and \eqref{eq:proof_1_Omega_hat_and_population_cov}, the least squares loss of $\hat{B}$ in the large sample limit is
\begin{equation}\label{eq:proof_1_least_squares_alternative_parameters}
\ell(\hat{B}; \tilde{\Sigma}) = \frac{1}{2}\tr\big( (I-\hat{B})^T \tilde{\Sigma} (I-\hat{B}) \big)=\frac{1}{2}\tr(\hat{\Omega})=\frac{1}{2}\tr(\tilde{L}\odot \tilde{L}).
\end{equation}

For the alternative weighted adjacency matrix $\hat{B}$ to yield a lower least squares loss than the true one $\tilde{B}$, i.e., $\ell(\hat{B}; \tilde{\Sigma})<\ell(\tilde{B}; \tilde{\Sigma})$, we have the following inequality by substituting Eqs. \eqref{eq:proof_1_least_squares_true_parameters} and 
\eqref{eq:proof_1_least_squares_alternative_parameters}:
\begin{equation}\label{eq:proof_1_lower_least_squares}
\tr(\tilde{\Omega})-\tr(\tilde{L}\odot \tilde{L})>0.
\end{equation}

The following lemma concludes the proof of the proposition.
\begin{lemma}\label{lemma:nonzero_measure_counterexample_inequality}
Consider strictly lower triangular matrix $\tilde{B}$ and positive diagonal matrix $\tilde{\Omega}$. The set of parameters $(\tilde{B},\tilde{\Omega})$ that satisfy Inequalities \eqref{eq:proof_1_varsortability_one} and \eqref{eq:proof_1_lower_least_squares} has a nonzero Lebesgue measure.
\end{lemma}
\par\noindent{\bfseries\upshape Proof of Lemma \ref{lemma:nonzero_measure_counterexample_inequality}\ }
Here we give a proof for the cases of $d\geq 4$; such a proof holds similarly for $d=3$. First note that the LHS of Inequalities \eqref{eq:proof_1_varsortability_one} and \eqref{eq:proof_1_lower_least_squares} are continuous functions of (the entries of) $\tilde{B}$ and $\tilde{\Omega}$ on the given domain, because, given that $I-\tilde{B}$ is invertible, the mappings of matrix inversion \citep{Wilkinson1965rounding,Stewart1969continuity} and Cholesky factor \citep[Chapter~12]{Schatzman2002numerical} are continuous. In this case, it suffices to find a specific point $(\tilde{B},\tilde{\Omega})$ on the given domain that satisfies Inequalities~\eqref{eq:proof_1_varsortability_one} and~\eqref{eq:proof_1_lower_least_squares}, because these inequalities are defined by continuous functions and thus they hold in an open neighborhood of this point.\looseness=-1

According to the given domain, consider the following parameters:
\[
\tilde{B} = \begin{bmatrix}
0 & \cdots & 0 & 0 & 0 & 0 \\
\vdots & \ddots & \vdots & \vdots & \vdots & \vdots \\
0 & \cdots & 0 & 0 & 0 & 0 \\
0 & \cdots & 0 & 0 & 0 & 0 \\
0 & \cdots & 0 & \frac{1}{2} & 0 & 0 \\
0 & \cdots & 0 & \frac{3}{4} & \frac{1}{4} & 0 \\
\end{bmatrix}
\quad\text{and}\quad
\tilde{\Omega}=\operatorname{diag}\left(d-2, d-3,d-4,\dots, 4, 3, 2, \frac{1}{10},1,1\right),
\]
both of which are $d\times d$ matrices. Note that $\tilde{B}$ is a strictly lower triangular matrix. We then have
\begin{equation*}
\tilde{\Sigma} = (I-\tilde{B})^{-T} \tilde{\Omega} (I-\tilde{B})^{-1}=\begin{bmatrix}
d-2 & 0 & \cdots & 0 & 0 & 0 & 0 & 0 \\
0 & d-3 & \cdots & 0 & 0 & 0 & 0 & 0 \\
\vdots & \vdots & \ddots & \vdots & \vdots & \vdots & \vdots & \vdots \\
0 & 0 & \cdots & 3 & 0 & 0 & 0 & 0 \\
0 & 0 & \cdots & 0 & 2 & 0 & 0 & 0 \\
0 & 0 & \cdots & 0 & 0 & \frac{357}{320} & \frac{23}{32} & \frac{7}{8} \\
0 & 0 & \cdots & 0 & 0 & \frac{23}{32} & \frac{17}{16} & \frac{1}{4} \\
0 & 0 & \cdots & 0 & 0 & \frac{7}{8} & \frac{1}{4} & 1 \\
\end{bmatrix}.
\end{equation*}
Clearly, Inequality \eqref{eq:proof_1_varsortability_one} is satisfied. By simple calculation, the Cholesky factor $\tilde{L}$ of $\tilde{\Sigma}$ is
\begin{equation*}
\tilde{L}=\begin{bmatrix}
\sqrt{d-2} & 0 & \cdots & 0 & 0 & 0 & 0 & 0 \\
0 & \sqrt{d-1} & \cdots & 0 & 0 & 0 & 0 & 0 \\
\vdots & \vdots & \ddots & \vdots & \vdots & \vdots & \vdots & \vdots \\
0 & 0 & \cdots & \sqrt{3} & 0 & 0 & 0 & 0 \\
0 & 0 & \cdots & 0 & \sqrt{2} & 0 & 0 & 0 \\
0 & 0 & \cdots & 0 & 0 & \frac{\sqrt{1785}}{40} & 0 & 0 \\
0 & 0 & \cdots & 0 & 0 & \frac{23\sqrt{1785}}{1428} & \frac{\sqrt{76398}}{357} & 0 \\
0 & 0 & \cdots & 0 & 0 & \frac{\sqrt{1785}}{51} & \frac{-8\sqrt{76398}}{5457} & \frac{4\sqrt{107}}{107} \\
\end{bmatrix},
\end{equation*}
which indicates that Inequality \eqref{eq:proof_1_lower_least_squares} also holds. \hfill $\blacksquare$

\subsection{Proof of Proposition \ref{proposition:low_varsortability_but_correct_dag}}\label{app:proof_low_varsortability_but_correct_dag}
Let $(\tilde{B},\tilde{\Omega})$ be the parameters associated with the linear SEMs over variables $X=(X_1,\dots,X_d)$ induced by DAG $\tilde{\G}$, where the noise variables follow Gaussian distributions.\footnote{For clarity in this proof, we use $(\tilde{B},\tilde{\Omega})$ to denote the parameters instead of $(B,\Omega)$ as in the proposition statement. Similarly, we use $\tilde{\G}$ to denote the DAG instead of $\G$.}
By definition, $\tilde{\Omega}$ is a positive diagonal matrix, while the weighted adjacency matrix $\tilde{B}$ corresponds to a DAG.

We use $\text{SN}_{\tilde{\G}}$ and $\text{NSN}_{\tilde{\G}}$ to denote the set of source nodes and non-source nodes in DAG $\tilde{\G}$, respectively. We also denote by $\text{PA}_{\tilde{\G}}(X_i)$, $\text{CH}_{\tilde{\G}}(X_i)$, and $\text{DE}_{\tilde{\G}}(X_i)$ the set of parents, children, and descendants of $X_i$ in $\tilde{\G}$, respectively. 
For any two DAGs $\G_1$ and $\G_2$, we write $\G_1\subseteq \G_2$ if all edges in $\G_1$ are present in $\G_2$. Furthermore, let $\mathbb{D}^d$ be the set of $d$-node DAGs, and, for DAG $\G$, define $\mathbb{R}_\G$ as
\[
\mathbb{R}_\G\coloneqq\{B\in\mathbb{R}^{d\times d}: B_{i,j}=0 \text{ when } X_i\rightarrow X_j \text{ is not an edge in }\G\}.
\]
Note that, in the large sample limit, the covariance matrix of $X$ is
\[
\tilde{\Sigma} = (I-\tilde{B})^{-T} \tilde{\Omega} (I-\tilde{B})^{-1}.
\]

The varsortability is equal to $\varv_{\tilde{\G}}^{\textrm{source}}$, where $\varv_{\tilde{\G}}^{\textrm{source}}$ is defined as in the proposition statement, if the following inequalities hold:
\begin{equation}\label{eq:proof_2_inequality_source_nodes}
\operatorname{Var}(X_i)<\operatorname{Var}(X_j)\quad \iff\quad \tilde{\Sigma}_{i,i}-\tilde{\Sigma}_{j,j}<0,\quad 
\text{for \ } X_i\in\text{SN}_{\tilde{\G}}, \, X_j\in\text{DE}_{\tilde{\G}}(X_i),
\end{equation}
and
\begin{equation}\label{eq:proof_2_inequality_non_source_nodes}
\operatorname{Var}(X_i)>\operatorname{Var}(X_j)\quad \iff\quad \tilde{\Sigma}_{i,i}-\tilde{\Sigma}_{j,j}>0,\quad 
\text{for \ } X_i\in\text{NSN}_{\tilde{\G}}, \, X_j\in\text{CH}_{\tilde{\G}}(X_i).
\end{equation}

With a slight abuse of notation, let $\ell(\G;\tilde{\Sigma})$ denote the least squares score of DAG $\G$, and $\ell(B;\tilde{\Sigma})$ denote the least squares loss of matrix $B$. Specifically, we have
\begin{equation}\label{eq:proof_2_least_squares_score}
\ell(\G;\tilde{\Sigma}) \coloneqq \ \min_{B\in\mathbb{R}_\G} \  \ell(B;\tilde{\Sigma}).
\end{equation}
For any DAG $\hat{\G}\supseteq\tilde{\G}$, we have $\ell(\hat{\G};\tilde{\Sigma})=\ell(\tilde{B};\tilde{\Sigma})$ by \citet[Lemma~6]{Loh2014high}. Therefore, it suffices to consider the DAG $\hat{\G}$ where $\hat{\G}\not\supseteq\tilde{\G}$. We then consider the following inequality:
\begin{equation}\label{eq:proof_2_inequality_lowest_least_squares}
\ell(\tilde{B};\tilde{\Sigma})-\ell(\hat{\G};\tilde{\Sigma})<0,\quad \text{for \ } \hat{\G}\in\mathbb{D}^d,\, \hat{\G}\not\supseteq\tilde{\G},
\end{equation}
or equivalently, the true weighted adjacency matrix $\tilde{B}$ yields the lowest least squares score.

The following lemma concludes the proof of the proposition.

\begin{lemma}\label{lemma:low_varsortability_but_correct_dag_inequality}
Given DAG $\tilde{\G}$, consider matrix $\tilde{B}\in\mathbb{R}_{\tilde{\G}}$ and positive diagonal matrix $\tilde{\Omega}$. The set of parameters $(\tilde{B},\tilde{\Omega})$ that satisfy Inequalities \eqref{eq:proof_2_inequality_source_nodes}, \eqref{eq:proof_2_inequality_non_source_nodes}, and \eqref{eq:proof_2_inequality_lowest_least_squares} has a nonzero Lebesgue measure.
\end{lemma}
\par\noindent{\bfseries\upshape Proof of Lemma \ref{lemma:low_varsortability_but_correct_dag_inequality}\ }
First it is clear that the LHS of Inequalities \eqref{eq:proof_2_inequality_source_nodes} and \eqref{eq:proof_2_inequality_non_source_nodes} are continuous functions of (the entries of) $\tilde{B}$ and $\tilde{\Omega}$ on the given domain, because $I-\tilde{B}$ is invertible and the mapping of matrix inversion is continuous \citep{Wilkinson1965rounding,Stewart1969continuity}.
For Inequality \eqref{eq:proof_2_inequality_lowest_least_squares}, the least squares score $\ell(\hat{\G};\tilde{\Sigma})$ is computed by solving the optimization problem in Eq. \eqref{eq:proof_2_least_squares_score}; each column, say the $j$-th one, of the estimated solution consists of the coefficients obtained from linearly regressing $X_j$ upon its parents in DAG $\hat{\G}$ \citep[Remark~5]{Loh2014high}. Therefore, it is straightforward to show that the LHS of Inequality~\eqref{eq:proof_2_inequality_lowest_least_squares} is also a continuous function of $\tilde{B}$ and $\tilde{\Omega}$ on the given domain, because $\tilde{\Sigma}$ is  symmetric positive definite and so any of its principal submatrix is invertible. In this case, it suffices to find a specific point $(\tilde{B},\tilde{\Omega})$ on the given domain that satisfies Inequalities \eqref{eq:proof_2_inequality_source_nodes},
\eqref{eq:proof_2_inequality_non_source_nodes},
and \eqref{eq:proof_2_inequality_lowest_least_squares}, because these inequalities are defined by continuous functions and thus they hold in an open neighborhood of this point.

We start with Inequality \eqref{eq:proof_2_inequality_lowest_least_squares}.
For the free parameters in diagonal matrix $\tilde{\Omega}$, we set them to be equal, i.e., $\tilde{\sigma}_1^2=\dots=\tilde{\sigma}_d^2=\tilde{\sigma}^2$. This restriction leads to the equal noise variances case, which, by \citet[Theorem~7]{Loh2014high}, indicates that the true weighted adjacency matrix $\tilde{B}$ yields the lowest least squares score. Thus, Inequality \eqref{eq:proof_2_inequality_lowest_least_squares} holds with this restriction.

We now construct parameter $\tilde{B}$ that such that Inequalities \eqref{eq:proof_2_inequality_source_nodes}
and \eqref{eq:proof_2_inequality_non_source_nodes} hold. Consider the variable $X_j$. If all parents of $X_{j}$ are source nodes in DAG $\tilde{\G}$, we set
$\tilde{B}_{i,j}$ to arbitrary (strictly) positive value for each edge $X_i\rightarrow X_j$ in $\tilde{\G}$. If some parents of $X_{j}$ are not source nodes in $\tilde{\G}$, then we set
\begin{equation}\label{eq:proof_2_definition_B}
\tilde{B}_{i,j}\coloneqq \frac{1}{\left|\text{PA}_{\tilde{\G}}(X_j)\right|}\left(\frac{c_j}{\operatorname{Var}(X_i)}\right)^{\frac{1}{2}}
\end{equation}
for each edge $X_i\rightarrow X_j$ in $\tilde{\G}$, where $c_j$ is an arbitrary value from the following interval:
\begin{equation}\label{eq:proof_2_definition_c}
c_j\in\left(0,\min_{X_k\in \text{PA}_{\tilde{\G}}(X_j)\cap \text{NSN}_{\tilde{\G}}}\{\operatorname{Var}(X_k)\}-\tilde{\sigma}^2\right).
\end{equation}
We construct the above matrix by following the topological order of DAG $\tilde{\G}$. Specifically, when we are constructing the $j$-th column of $\tilde{B}$, every incoming edge into the ancestors of variable $X_{j}$ has been assigned a weight. Thus, the variances of the parents of $X_{j}$ can be computed (see Eq.~\eqref{eq:proof_2_definition_c}).

Note that all entries of the matrix $\tilde{B}$ constructed above that correspond to the edges in $\tilde{\G}$ are nonzero. Therefore, for $X_i\in\text{SN}_{\tilde{\G}}$ and $X_j\in\text{DE}_{\tilde{\G}}(X_i)$, we have
\[
\operatorname{Var}(X_j)=\operatorname{Var}(\tilde{B}^T_{\cdot,j}X+N_j)=\operatorname{Var}(\tilde{B}^T_{\cdot,j}X)+\operatorname{Var}(N_j)>\operatorname{Var}(N_j)=\tilde{\sigma}^2=\operatorname{Var}(X_i),
\]
which implies that Inequality \eqref{eq:proof_2_inequality_source_nodes} holds.

Moreover, for $X_i\in\text{NSN}_{\tilde{\G}}$ and $X_j\in\text{CH}_{\tilde{\G}}(X_i)$, we have
\begin{align*}
\operatorname{Var}(X_j)&=\operatorname{Var}\left(\sum_{X_l\in \text{PA}_{\tilde{\G}}(X_j)}\tilde{B}_{l,j}X_l+N_j\right)\\
&=\sum_{X_l,X_m\in \text{PA}_{\tilde{\G}}(X_j)}\tilde{B}_{l,j}\tilde{B}_{m,j}\operatorname{Cov}(X_l,X_m) + \operatorname{Var}(N_j)\\
&\leq \sum_{X_l,X_m\in \text{PA}_{\tilde{\G}}(X_j)}\tilde{B}_{l,j}\tilde{B}_{m,j}\operatorname{Var}(X_l)^{\frac{1}{2}}\operatorname{Var}(X_m)^{\frac{1}{2}} + \tilde{\sigma}^2 &&\text{($\because \tilde{B}_{l,j}\tilde{B}_{m,j}>0$)}\\
&=\left|\text{PA}_{\tilde{\G}}(X_j)\right|^2\cdot \frac{c_j}{\left|\text{PA}_{\tilde{\G}}(X_j)\right|^2}+\tilde{\sigma}^2 &&\text{(Substituting Eq. \eqref{eq:proof_2_definition_B})}\\
&<\min_{X_k\in \text{PA}_{\tilde{\G}}(X_j)\cap \text{NSN}_{\tilde{\G}}}\{\operatorname{Var}(X_k)\}&&\text{(Substituting Eq. \eqref{eq:proof_2_definition_c})}\\
&\leq\operatorname{Var}(X_i).
\end{align*}
Thus, Inequality \eqref{eq:proof_2_inequality_non_source_nodes} also holds. Therefore, we have constructed a specific point $(\tilde{B},\tilde{\Omega})$ on the given domain that satisfies Inequalities \eqref{eq:proof_2_inequality_source_nodes},
\eqref{eq:proof_2_inequality_non_source_nodes},
and \eqref{eq:proof_2_inequality_lowest_least_squares}. \hfill $\blacksquare$

\section{Analysis of Non-Equal Noise Variances Formulation}
We provide analysis of the non-equal noise variances formulation. We experiment with random initializations in Appendix \ref{sec:random_initialization}, and GOLEM-EV with random perturbations in Appendix \ref{app:golem_ev_perturbation}. We consider alternative form of likelihood function (without profiling out the paramter $\Omega$) in Appendix~\ref{sec:alternative_likelihood}, and investigate the connection between estimated score and SHD in Appendix \ref{app:analysis_score_and_shd}.

\subsection{Random Initializations}\label{sec:random_initialization}
In Sections \ref{sec:varsortability} and \ref{sec:nonconvexity}, we adopt the initialization scheme with zero matrix for various structure learning methods such as GOLEM and NOTEARS. Note that this initialization scheme has been used by several existing works \citep{Zheng2018notears,Ng2020role,Bello2022dagma}, especially those based on linear SEMs. Here, we conduct experiments (following the setup in Section \ref{sec:varsortability_nv_case}) for GOLEM-NV with random initializations to examine whether it produces a better solution and helps remedy the possible nonconvexity issue discussed in Sections \ref{sec:varsortability_nv_case} and~\ref{sec:nonconvexity}. Specifically, each entry of the initial solution $B$ is sampled uniformly at random from $[-\epsilon,\epsilon]$, where~$\epsilon\in\{0.01,0.05,0.1\}$. Furthermore, we run such random initialization for $10$ times, and select the final solution that leads to the best score (see Eq. \eqref{eq:golem}). Apart from random initializations, we also report results for GOLEM-NV initialized with the solution of GOLEM-EV and that of FGES.\looseness=-1

The empirical results are reported in Table \ref{tab:random_initialization}. We observe that random initializations perform similarly to the initialization scheme with zero matrix, both of which lead to a much worse performance as compared to GOLEM-NV initialized with GOLEM-EV or with FGES. This suggests that random initializations do not seem to resolve the possible nonconvexity issue, at least in the setting considered here. Moreover, this further validates the findings in Section \ref{sec:varsortability_nv_case}, i.e., a proper initialization scheme is crucial to the quality of the final solution for non-equal noise variances formulation.\looseness=-1

\begin{table}[!h]
	\centering
	\caption{Empirical results of different initialization schemes. The number of variables, sample size, and noise ratio are $50$, $10^6$, and $2$, respectively. The standard errors computed over $30$ random repetitions are also reported. Here, ``int.'' stands for ``initialized''.}
	\begin{tabular}{rccccccccccc} 
	\toprule
		~ & SHD of CPDAG  & F1 of skeleton &  F1 of arrows \\
		\midrule
		GOLEM-NV int. with zero matrix & $120.27\pm 2.91$ & $0.50\pm 0.00$  & $0.04\pm 0.00$ \\
            GOLEM-NV int. with GOLEM-EV & $4.50\pm 1.11$ & $0.98\pm 0.01$  & $0.97\pm 0.01$ \\ 
            GOLEM-NV int. with FGES & $0.47\pm 0.47$ & $1.00\pm 0.00$  & $0.99\pm 0.00$ \\ 
            GOLEM-NV int. with $\operatorname{Unif}[-0.01, 0.01]$ & $123.13\pm 2.90$ & $0.50\pm 0.01$  & $0.04\pm 0.00$ \\ 
            GOLEM-NV int. with $\operatorname{Unif}[-0.05, 0.05]$ & $119.20\pm 2.75$ & $0.51\pm 0.01$  & $0.04\pm 0.01$ \\ 
            GOLEM-NV int. with $\operatorname{Unif}[-0.1, 0.1]$ & $112.77\pm 3.11$ & $0.51\pm 0.01$  & $0.04\pm 0.01$ \\
		\bottomrule
    \end{tabular}
\label{tab:random_initialization}
\end{table}

\vspace{-0.2em}
\subsection{Equal Noise Variances Formulation with Random Perturbations}\label{app:golem_ev_perturbation}
We investigate if random perturbations on the solution of GOLEM-EV improve the performance in the non-equal noise variances case. Specifically, we randomly perturb the solution of GOLEM-EV by adding a value sampled uniformly at random from $[-\epsilon, \epsilon]$ to each entry of the estimated matrix. We then use the perturbed solution to initialize GOLEM-EV and run the method again, and repeat this procedure for $10$ times. We consider $\epsilon\in\{0.01,0.05,0.1\}$, and denote the resulting methods as GOLEM-EV-Perturbed-$0.01$, GOLEM-EV-Perturbed-$0.05$, and GOLEM-EV-Perturbed-$0.1$, respectively. Here, we follow the setup (i.e., $50$ variables, degree of $2$, and $10^6$ samples) in Section 3.2.\looseness=-1

\begin{wrapfigure}{r}{0.49\textwidth}
  \begin{center}
    \includegraphics[width=0.48\textwidth]{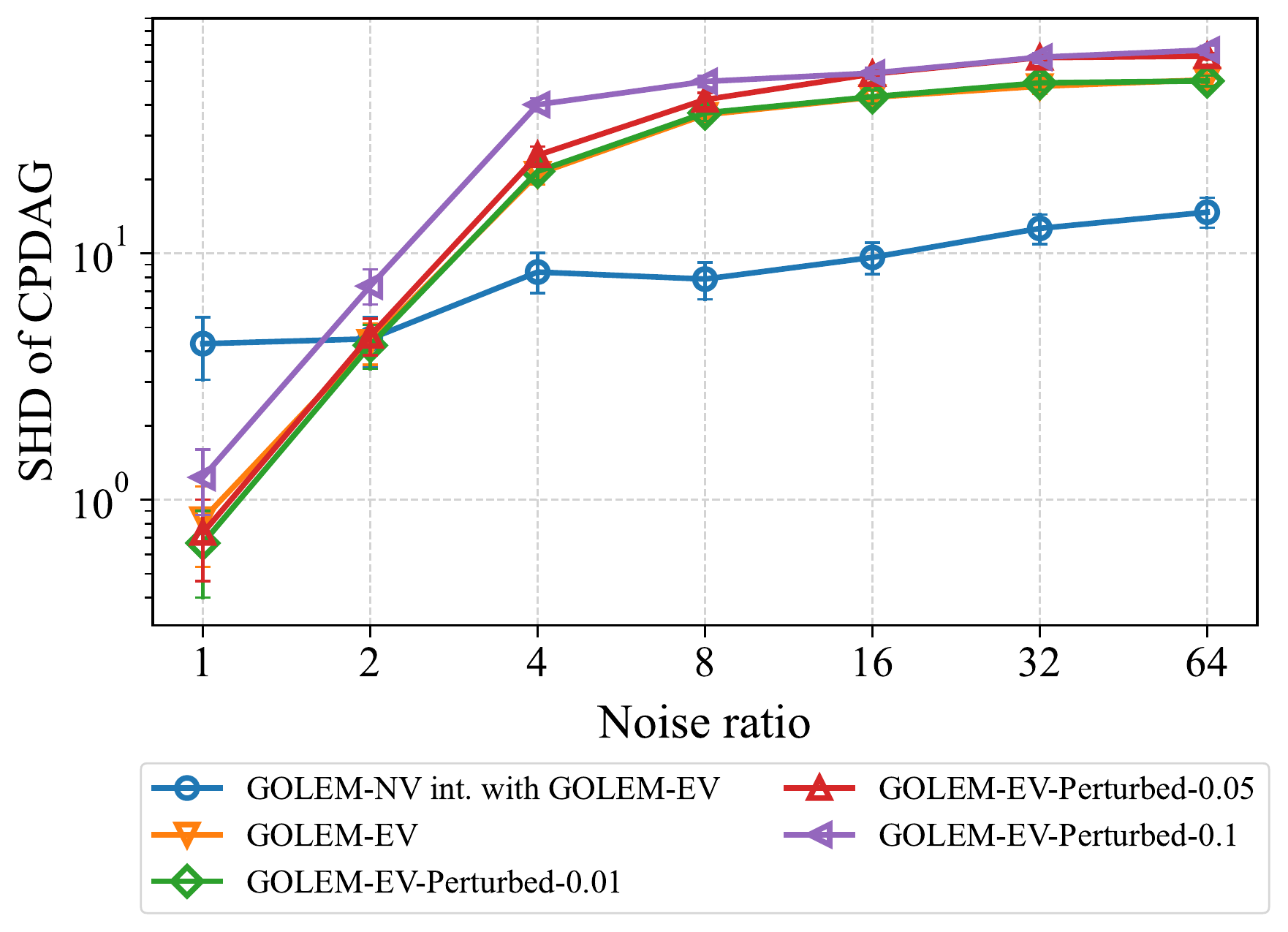}
  \end{center}
  \vspace{-1.1em}
  \caption{GOLEM-EV with perturbations.}
  \label{fig:golem_ev_perturbation}
\vspace{-0.3em}
\end{wrapfigure}
The results of GOLEM-EV-Perturbed, GOLEM-EV, and GOLEM-NV initialized with GOLEM-EV are shown in Figure~\ref{fig:golem_ev_perturbation}. It is observed that random perturbations on GOLEM-EV does not improve the performance much. Specifically, when the noise ratio $r\geq 4$, GOLEM-EV with perturbations, similar to that without perturbations, performs much worse than GOLEM-NV initialized with GOLEM-EV.  The reason is that when the noise ratio is high, there is model misspecification for GOLEM-EV because it assumes that the noise variances are equal; therefore, random perturbations on GOLEM-EV do not help. This further validates the importance of non-equal noise variances formulation (e.g., GOLEM-NV) to handle more general settings.

\vspace{-0.1em}
\subsection{Alternative Form of Likelihood Function}\label{sec:alternative_likelihood}
\begin{figure}[H]
\centering
\includegraphics[width=0.99\textwidth]{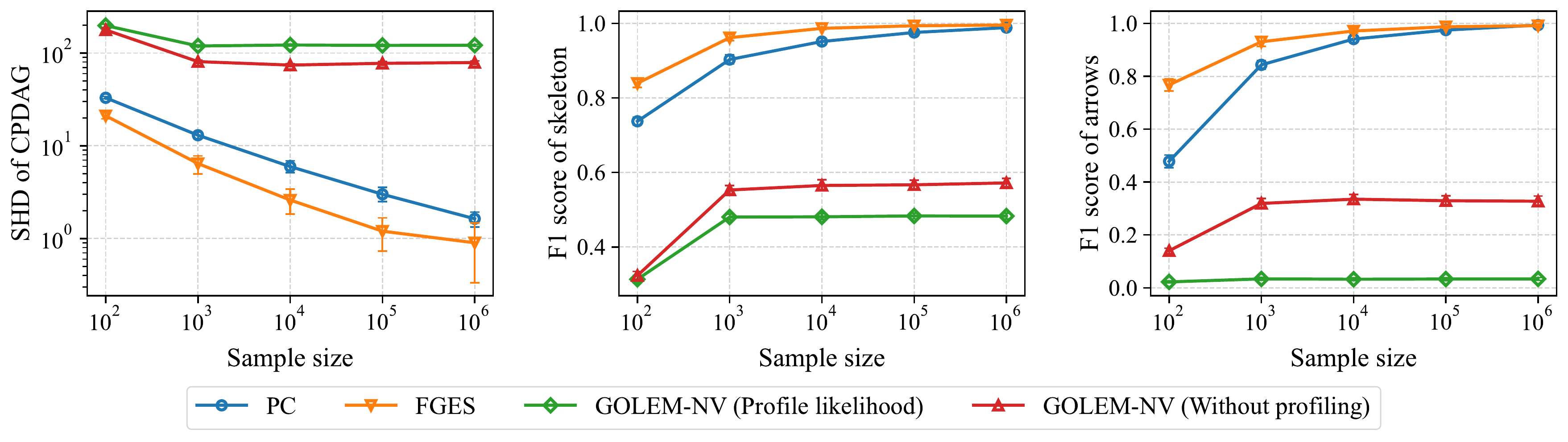}
\caption{Empirical results of different forms of likelihood function under different sample sizes. The number of variables is $50$. Error bars represent the standard errors computed over $30$ random repetitions.
}
\label{fig:profile_likelihood_result}
\end{figure}

In the derivation of GOLEM-NV \citep[Appendix~C.1]{Ng2020role}, the resulting optimization problem in Eq. \eqref{eq:golem} involves only the matrix $B$ because it profiles out the parameter $\Omega$ that corresponds to the noise variances. Here, we consider the form of likelihood function without profiling out such parameter to investigate whether it can perform well. Specifically, we consider the optimization problem
\vspace{-0.1em}
\begin{equation}\label{eq:golem_alternative}
\min_{\substack{B\in\mathbb{R}^{d\times d},\\ \sigma_1,\dots,\sigma_d>0}} \ \ \mathcal{L}_\text{NV}(B,\sigma_1,\dots,\sigma_d;\mathbf{X}) - \log|\det(I - B)| + \lambda_1\|B\|_1 + \lambda_2 h(B),
\end{equation}
\vspace{-0.1em}
where
\vspace{-0.1em}
\[
\mathcal{L}_\text{NV}(B,\sigma_1,\dots,\sigma_d;\mathbf{X})=\frac{1}{2}\sum_{i=1}^d\left(\log\sigma_i^2 + \frac{\|\mathbf{X}_{\cdot,i}-\mathbf{X}B_{\cdot,i} \|_2^2}{n\sigma_i^2}\right).
\]
\vspace{-0.1em}
The derivation of the likelihood function above is available at \citet[Appendix~C.1]{Ng2020role}. We follow the setting described in Section \ref{sec:search_strategies}, and report the empirical results in Figure \ref{fig:profile_likelihood_result}. One observes that both forms of likelihood functions with and without profiling out the noise variances lead to a poor performance (when initialized with zero matrix) even when the sample size is large. For instance, with $10^6$ samples, the SHDs of GOLEM with and without profiling are $121.43\pm 2.98$ and $78.97\pm 3.34$, respectively, while PC and FGES have much lower SHDs at $1.63\pm 0.32$ and $0.90\pm 0.58$, respectively. This suggests that both forms of likelihood functions may be susceptible to suboptimal local solutions possibly owing to nonconvexity. It is worth noting that we obtain the same observations with NOTEARS by replacing its least squares objective with $\mathcal{L}_\text{NV}(B,\sigma_1,\dots,\sigma_d;\mathbf{X})$ to handle the non-equal noise variances case, whose results
are omitted here for brevity.

Similar to the initialization sceheme described in Section \ref{sec:varsortability_nv_case}, we also considered using the solution of FGES to initialize the optimization problem \eqref{eq:golem_alternative}, by computing the initial solution of weighted matrix $B$ and noise variances $\sigma_1,\dots,\sigma_d$. A similar observation in Figure \ref{fig:varsortability_nv_case_golem_shd_cpdag_unstandardized} is obtained (the empirical results are omitted here for brevity), i.e., such an initialization strategy improves the performance of structure learning, which indicates that the alternative formulation of GOLEM-NV in Eq. \eqref{eq:golem_alternative} also depends largely on the initial solution. 

\subsection{Analysis of Estimated Score and SHD }\label{app:analysis_score_and_shd}
We investigate the connection between the score function of GOLEM-NV and the quality of the estimated structure. Specifically, we consider GOLEM-NV initialized with zero matrix as well as by solutions of GOLEM-EV and FGES. We also include the comparison with zero matrix and random DAG. For these methods, we compute the SHD of the estimated CPDAGs and the score of GOLEM-NV (before thresholding) given in Eq. \eqref{eq:golem}. We consider $50$-node ER1 graphs with a noise ratio of $16$. 

\begin{figure}[H]
\centering  
\subfloat[SHD of CPDAG.]{
    \includegraphics[width=0.31\textwidth]{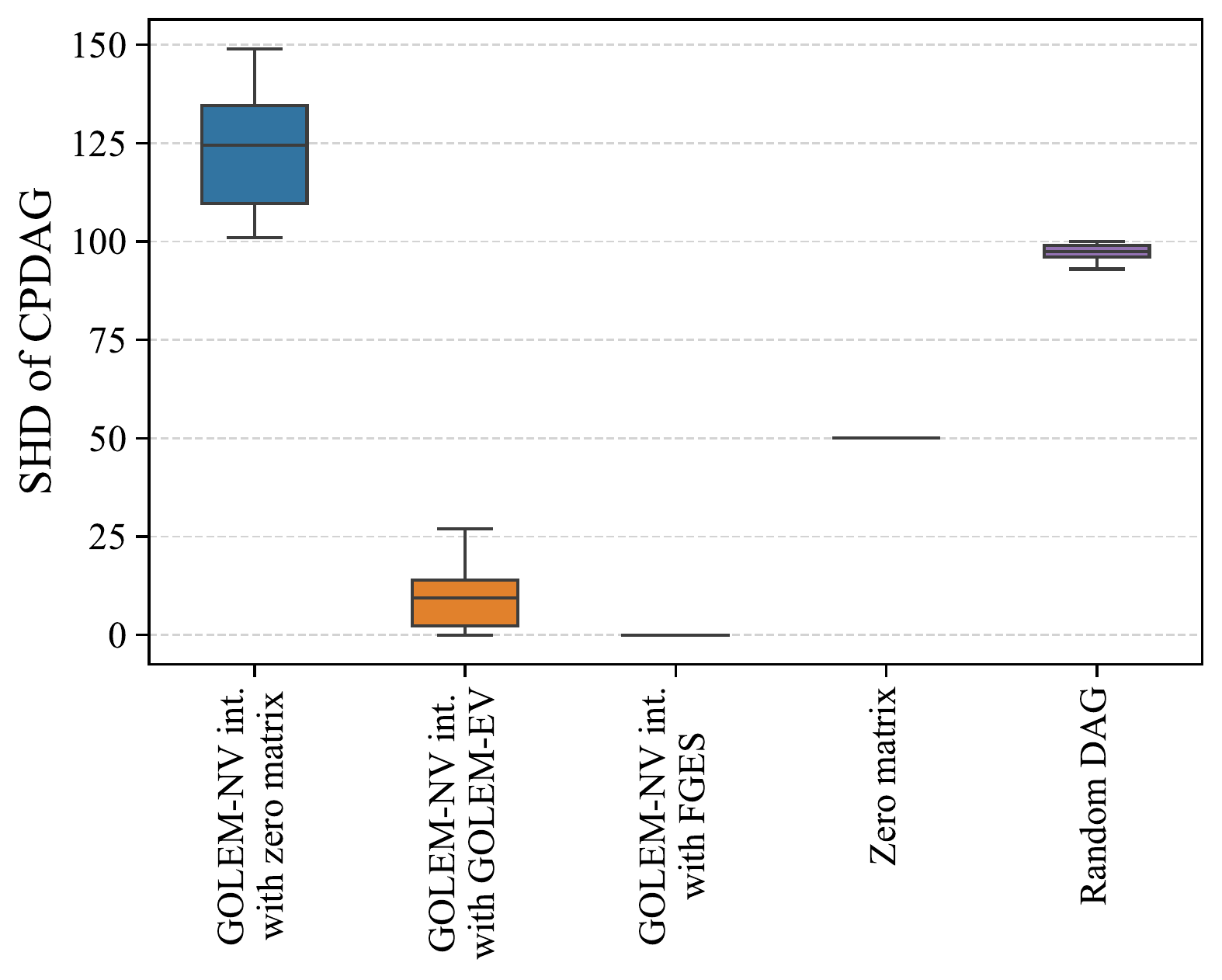}
    \label{fig:unstandardized_golem_shd_cpdag_box_plot}
}
\subfloat[Score.]{
    \includegraphics[width=0.31\textwidth]{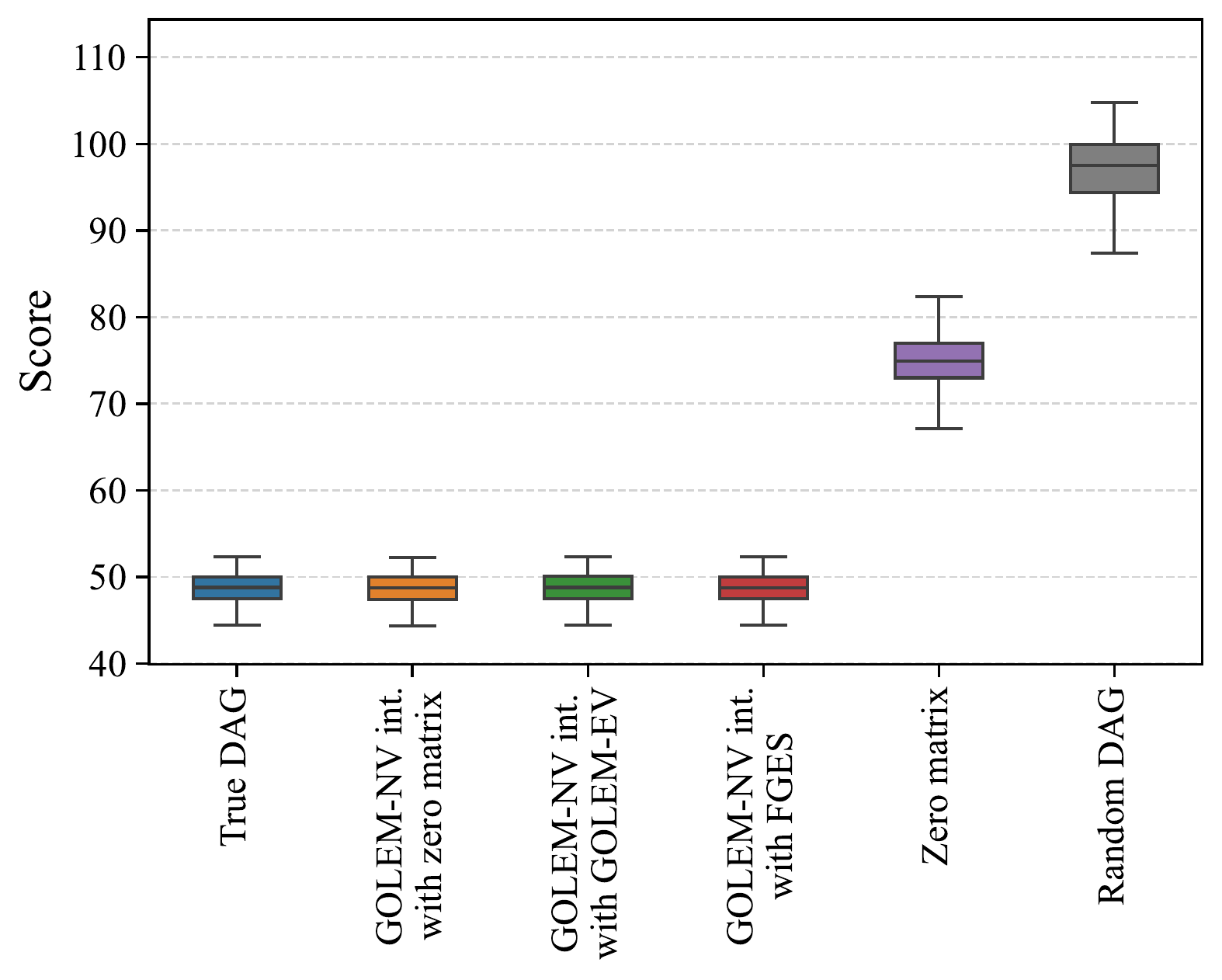}
    \label{fig:unstandardized_golem_score_box_plot}
}
\subfloat[SHD against score.]{
    \includegraphics[width=0.34\textwidth]{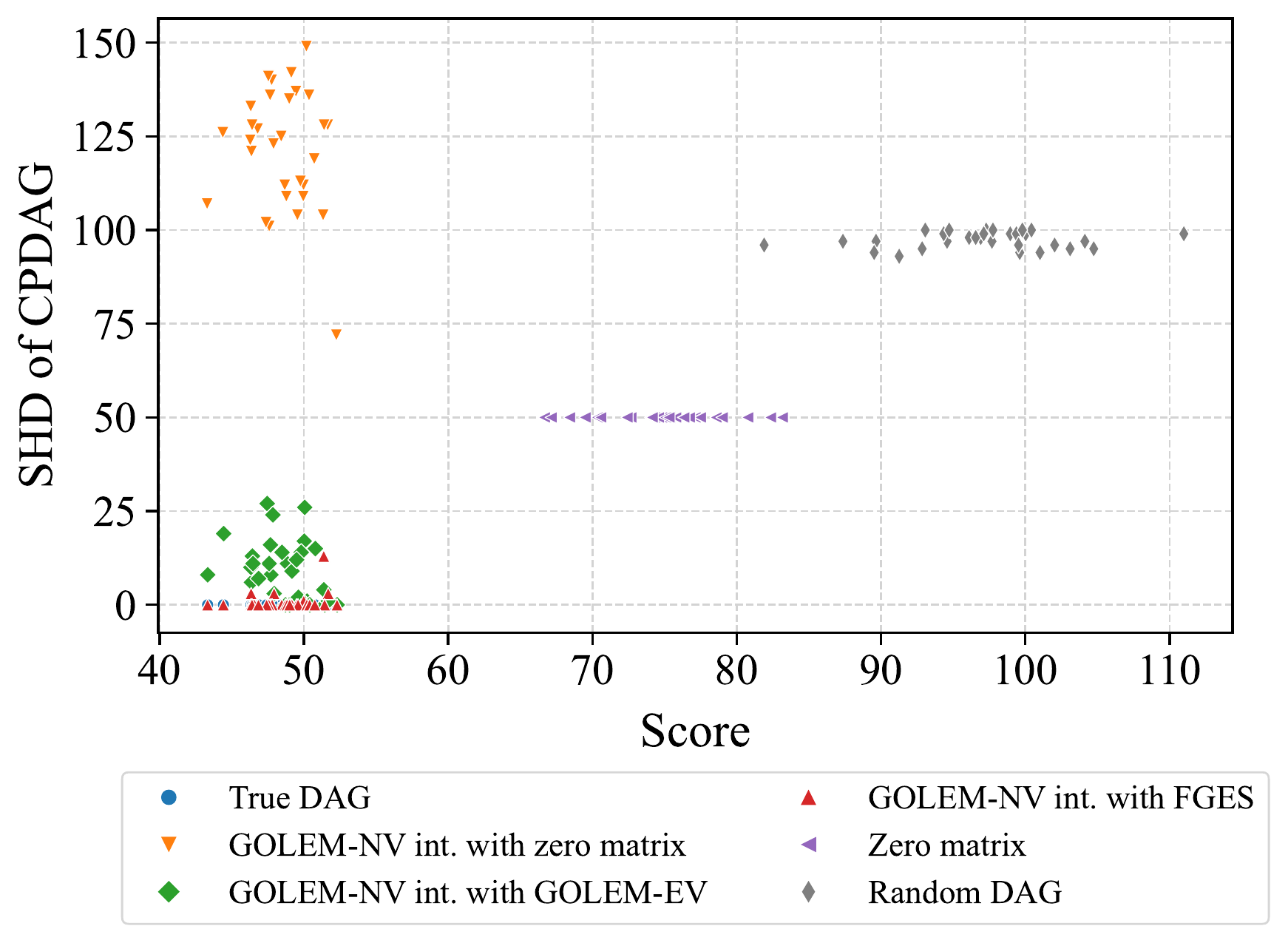}
    \label{fig:unstandardized_golem_shd_score_scatter_plot}
}
\caption{Visualization of estimated score and SHD for different methods over $30$ random repetitions. Lower is better for both SHD and score.}
\label{fig:score_and_shd}
\end{figure}

The results are reported in Figure \ref{fig:score_and_shd}. One observes that GOLEM-NV initialized with GOLEM-EV and by FGES achieves a low score that is close to the score of the true DAG. At the same time, they also have a low SHD, indicating that their estimated structures are close to the true CPDAGs. However, this is not the case for GOLEM-NV initialized with zero matrix. Specifically, it also achieves a score that is very close to that of the true DAG, but its SHD is $121.43\pm 2.98$. This appears to suggest that (1) a lower score does not necessarily lead to a low SHD-CPDAG, and (2) the nonconvex landscape of GOLEM-NV may contain suboptimal solutions whose scores are very close to the score of the global minimizer, but their corresponding structures may be very far from the true CPDAG. Thus, this demonstrates that nonconvexity may be a severe concern, at least in the setting considered here.
\looseness=-1

\section{Comparison between SCAD and $\ell_0$ Penalties}\label{app:comparison_scad_and_l0_penalties}
\begin{wrapfigure}{r}{0.48\textwidth}
\vspace{-1.3em}
  \begin{center}
    \includegraphics[width=0.47\textwidth]{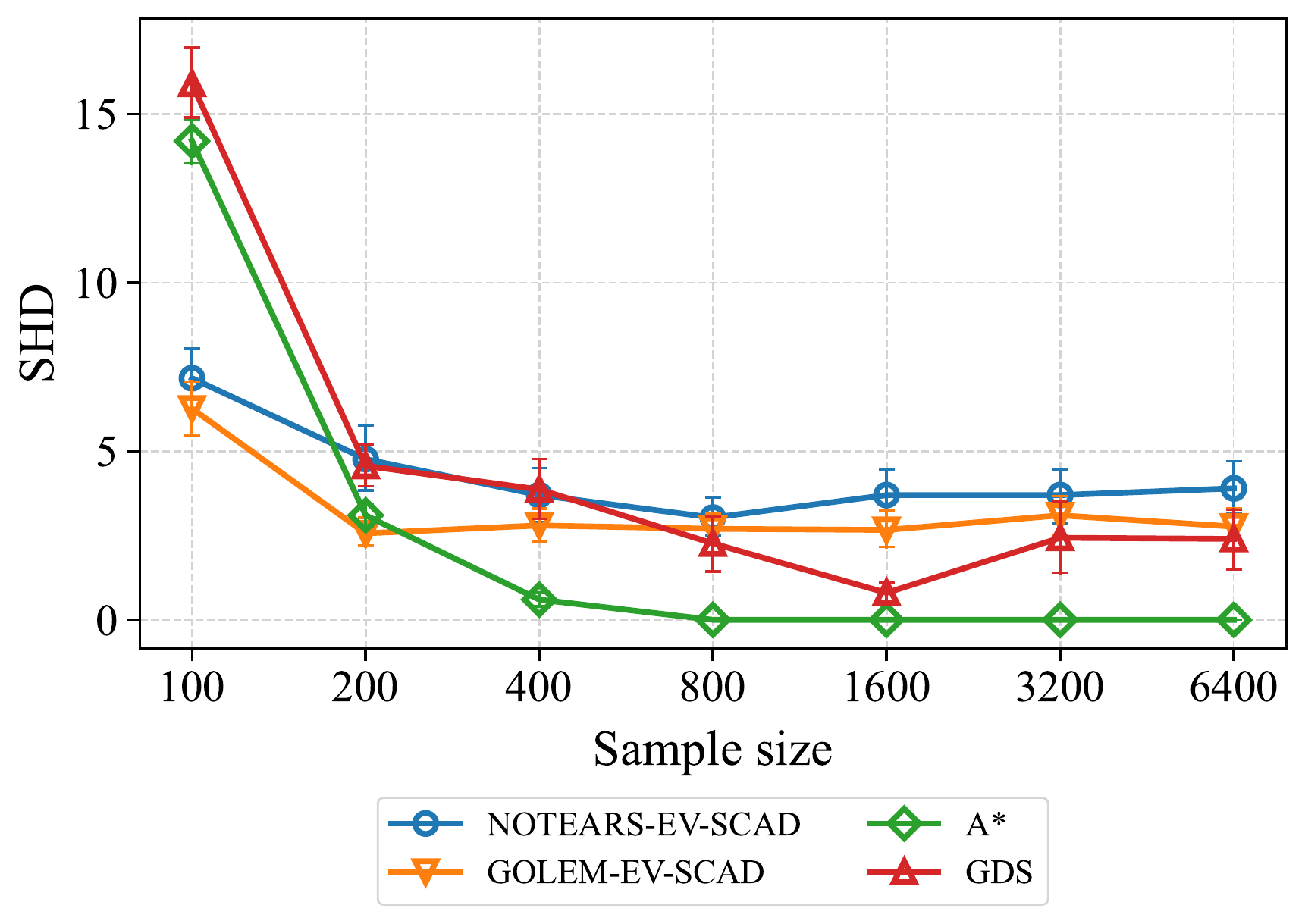}
  \end{center}
  \vspace{-1.1em}
  \caption{SCAD and $\ell_0$ penalties.}
  \label{fig:l0_15nodes_degree_4_shd}
\vspace{-0.3em}
\end{wrapfigure}
Based on the analysis in Section \ref{sec:sparsity}, one may wonder how different specifications of the sparsity penalty considered above compare to $\ell_0$ penalty. Since SCAD penalty and MCP perform similarly, we compare GOLEM-EV and NOTEARS-EV equipped with the SCAD penalty to discrete approaches, i.e., A* and GDS, that employ $\ell_0$ penalty. We consider 15-node ER2 graphs with varying sample sizes, and report the SHDs in Figure \ref{fig:l0_15nodes_degree_4_shd}. Comparing NOTEARS-EV, A*, and GDS that all use the least squares objective, one observes that A*, not surprisingly, has the lowest SHDs, while GDS is on par with it. On the other hand, NOTEARS-EV has a slightly higher SHD than that of GDS when the sample size increases. That is, NOTEARS-EV, even with the SCAD penalty, may perform worse than a greedy search method such as GDS, not to mention exact search method such as A*. Such a performance gap conveys a cautionary message that continuous structure learning approaches may be inevitably susceptible to the nonconvexity issue that might be severe in practice. This is in contrast with the observation by \citet[Section~5.3]{Zheng2018notears} that the estimated solution by NOTEARS-EV is very close to the global minimizer despite its nonconvexity. A possible reason is that we apply a relatively small threshold $0.1$, and as discussed in Section \ref{sec:thresholding}, NOTEARS-EV may require a larger threshold, e.g., $0.3$, to achieve a decent performance.

\section{Supplementary Experiment Details}\label{app:experiment_details}
We provide supplementary experiment details for Sections \ref{sec:varsortability_ev_case}, \ref{sec:varsortability_nv_case}, \ref{sec:search_strategies}, \ref{sec:dag_constraints}, \ref{sec:other_cases}, \ref{sec:thresholding}, and \ref{sec:sparsity}.

\subsection{Implementation Details and Evaluation Metrics}\label{app:implementation_details}
\paragraph{Implementation details.}
For the structure learning methods, we use the default hyperparameters and official implementations (including initialization scheme) from the authors, unless otherwise stated. As suggested in Section \ref{sec:thresholding}, using a relatively large threshold of $0.3$ as in existing works may be harmful and remove many true edges, especially when the edge weights in the true weighted adjacency matrix are small. In this work, we consider a smaller threshold of $0.1$ for continuous approaches. To ensure a fair comparison, we also apply the same thresholding on (the least squares coefficients of) the estimated solutions of GDS and A*. Furthermore, for NOTEARS-NV, DPDAG-NV, DAGMA-NV, and NOCURL-NV, we use the same hyperparameter as GOLEM-NV for sparsity penalty, i.e., $\lambda=0.002$, and replace their least squares objective $\ell(B;\mathbf{X})$ with $\mathcal{L}_\text{NV}(B;\mathbf{X})$, which correspond to the likelihood of linear Gaussian DAGs assuming non-equal noise variances. For sparsity, SCAD penalty and MCP require an additional hyperparameter $a$, which we set to $3.7$ following~\citet{Fan2001variable}.\looseness=-1

We provide further details for specific methods as follows:
\begin{itemize}
\item GOLEM \citep{Ng2020role}: We use L-BFGS to solve the unconstrained optimization problem, since it runs faster than Adam \citep{Kingma2014adam} and leads to a similar performance. For the experiments in Section~\ref{sec:sparsity}, we adopt $\lambda_1=0.1$ for GOLEM-EV as it performs better when the sample size is small.
\item NOTEARS \citep{Zheng2018notears}, NOTEARS-MLP \citep{Zheng2020learning}, and NOTEARS-ICA \citep{Zheng2020thesis}: We use the quadratic penalty method instead of augmented Lagrangian method to solve the constrained optimization problem~\citep{Bertsekas1999nonlinear,Nocedal2006numerical}, since \citet{Ng2022convergence} demonstrated that their performance is nearly identical, and that the former converges in a fewer number of iterations.
\vspace{-0.1em}
\item DPDAG \citep{Charpentier2022differentiable}: In the original work \citep{Charpentier2022differentiable}, DPDAG was combined with variational inference and MLPs, which leads to the VI-DP-DAG method for the nonlinear case. Here, we adapt it to learn linear DAGs (without the parts of variational inference and MLPs).
\vspace{-0.1em}
\item A* \citep{Yuan2013learning}: We adapt an implementation from the \texttt{causal-learn} package \citep{Zheng2024causallearn} and modify the BIC score \citep{Schwarz1978estimating} to least squares score \citep{Loh2014high}. We set the coefficient of $\ell_0$ penalty to $0.01$. 
\vspace{-0.1em}
\item GDS \citep{Peters2013identifiability}: We use our own implementation in Python. A key difference with the original method \citep{Peters2013identifiability} is that we use the least squares score \citep{Loh2014high} instead of likelihood score, both of which assume equal noise variances. We set the coefficient of $\ell_0$ penalty to $0.01$.
\vspace{-0.1em}
\item PC \citep{Spirtes1991pc} and FGES \citep{Ramsey2017million}: We use the implementation from the \texttt{py-causal} package available at the GitHub repository \url{https://github.com/bd2kccd/py-causal}, which is a Python wrapper of the TETRAD project \citep{Scheines1998tetrad}. We adopt the Fisher Z test and BIC score \citep{Schwarz1978estimating} for PC and FGES, respectively.
\end{itemize}
\vspace{-0.1em}
\paragraph{Evaluation metrics.}
For the linear Gaussian setting with equal noise variances formulations, as well as linear non-Gaussian and nonlinear settings, we report the SHD, F1 score, and recall of the estimated DAG. For the linear Gaussian setting with non-equal noise variances formulations, we report the SHD of estimated CPDAG, as well as the F1 score of estimated skeleton and directed edges. In the latter case, for those methods that output a DAG, an additional step is needed to convert it into a CPDAG. Furthermore, we provide the standard errors over $30$ random repetitions for all metrics.\looseness=-1

\vspace{-0.1em}
\subsection{Discussion of Search Strategies}\label{app:search_strategies_details}
As discussed in Section \ref{sec:continuous_structure_learning}, NOTEARS solves a constrained optimization problem with a hard DAG constraint by using augmented Lagrangian \citep{Zheng2018notears} or quadratic penalty method \citep{Ng2022convergence}, while GOLEM \citep{Ng2020role} solves an unconstrained optimization with a soft DAG constraint. These essentially represent different strategies to search for a DAG. Similar to NOTEARS, DAGMA \citep{Bello2022dagma} solve a constrained optimization problem, but adopts a procedure similar to the barrier method \citep{Nocedal2006numerical} with a log-determinant DAG constraint. Instead of enforcing acyclicity via a constrained optimization problem such as NOTEARS and DAGMA, another line of approaches directly search in the space of DAGs. Specifically, NOCURL~\citep{Yu2021nocurl} developed an algebraic representation of DAGs based on graph Hodge theory~\citep{Jiang2011statistical,Jorgen2009digraphs} that directly outputs weighted adjacency matrix of a DAG, while DPDAG~\citep{Charpentier2022differentiable} adopts a differentiable DAG sampling procedure that (1) samples a linear ordering of the variables using Gumbel-Sinkhorn~\citep{Mena2018learning} or Gumbel-Top-k \citep{Kool2019stochastic} reparametrizations, and (2) samples a weighted adjacency matrix consistent with the ordering. Note that DPDAG was combined with variational inference and MLPs, which leads to the VI-DP-DAG method for the nonlinear case; here, we adapt it to learn linear DAGs (without the parts of variational inference and MLPs). These several approaches represent different strategies to traverse the search space for estimating a DAG with continuous optimization.\looseness=-1

\vspace{-0.2em}
\section{Supplementary Experiment Results}\label{app:experiment_results}
We provide supplementary experiment results for Sections \ref{sec:varsortability_ev_case}, \ref{sec:varsortability_nv_case}, \ref{sec:search_strategies}, \ref{sec:dag_constraints}, \ref{sec:other_cases}, \ref{sec:thresholding}, and \ref{sec:sparsity}.

\clearpage
\subsection{With Equal Noise Variances Formulation}\label{app:varsortability_ev_case}

\begin{figure}[H]
\centering  
\subfloat[$15$ variables with $100$ samples.]{
    \includegraphics[width=0.92\textwidth]{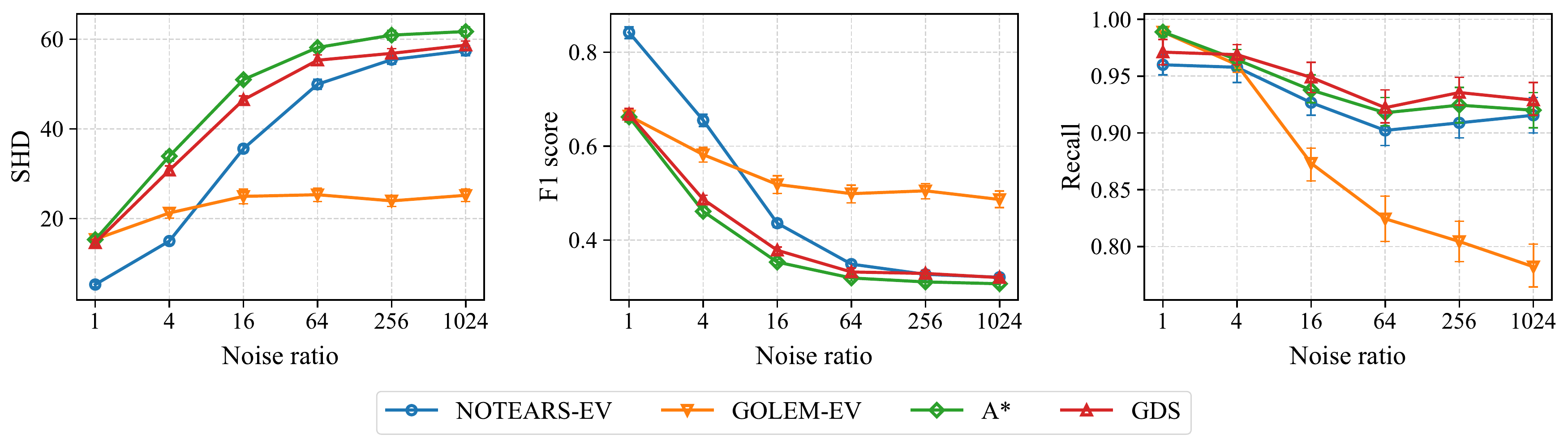}
    \label{fig:varsortability_ev_case_result_15nodes_100samples}
}\\
\subfloat[$15$ variables with $10^6$ samples.]{
    \includegraphics[width=0.92\textwidth]{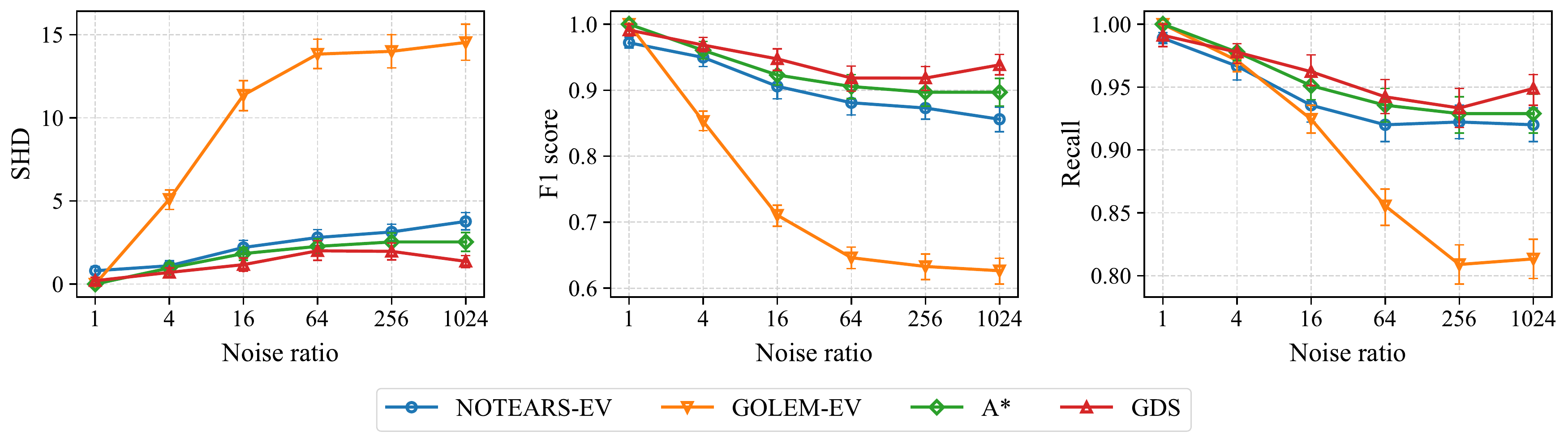}
    \label{fig:varsortability_ev_case_result_15nodes_1000000samples}
}\\
\subfloat[$50$ variables with $100$ samples.]{
    \includegraphics[width=0.92\textwidth]{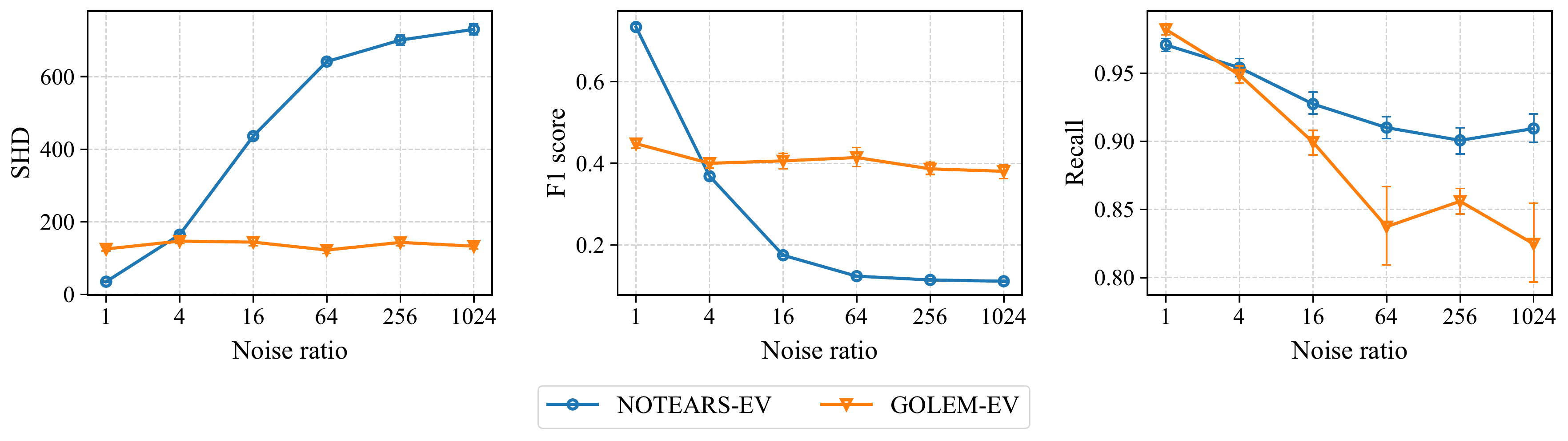}
    \label{fig:varsortability_ev_case_result_50nodes_100samples}
}\\
\subfloat[$50$ variables with $10^6$ samples.]{
    \includegraphics[width=0.92\textwidth]{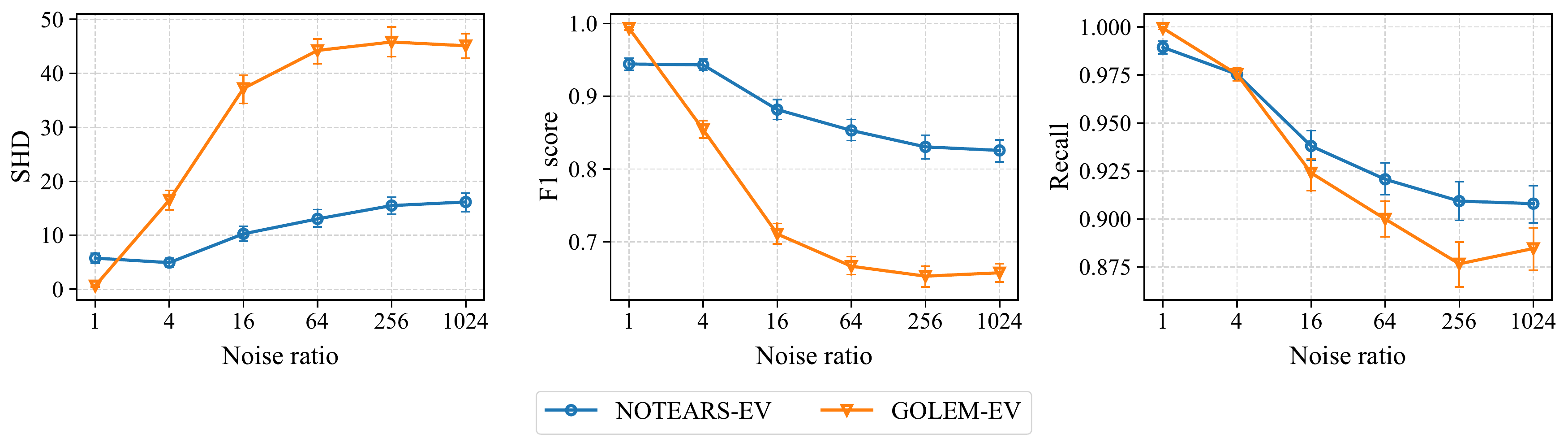}
    \label{fig:varsortability_ev_case_result_50nodes_1000000samples}
}
\caption{Empirical results of structure learning methods assuming equal noise variances under different noise ratios. Error bars represent the standard errors computed over $30$ random repetitions.}
\label{fig:varsortability_ev_case_result}
\end{figure}

\subsection{With Non-Equal Noise Variances Formulation}\label{app:varsortability_nv_case}
\begin{figure}[H]
\centering
\subfloat[Without data standardization.]{
    \includegraphics[width=0.99\textwidth]{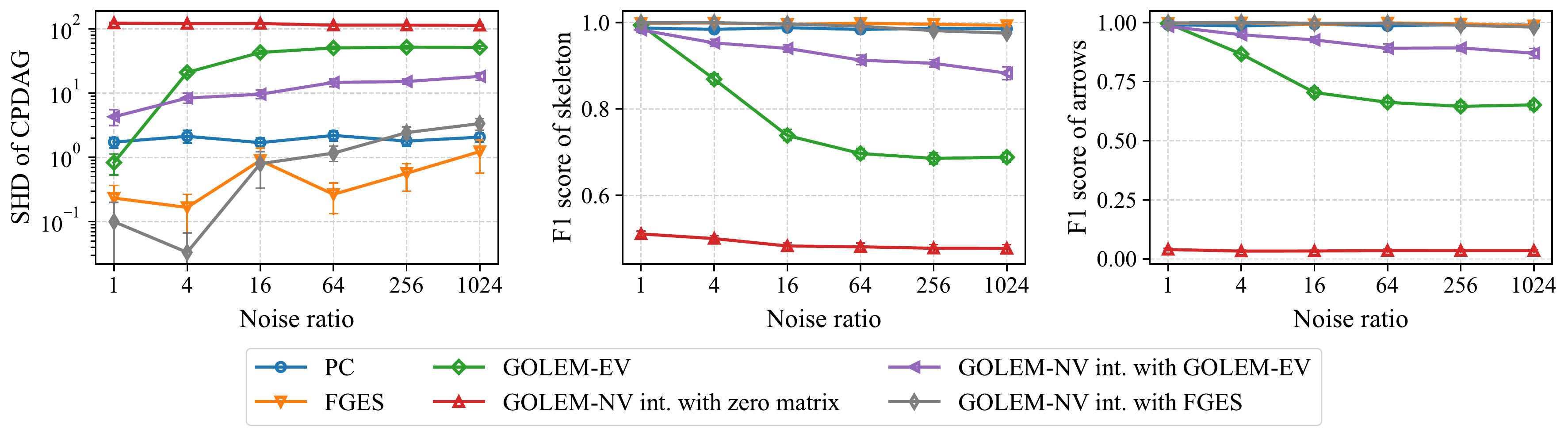}
    \label{fig:varsortability_nv_case_golem_unstandardized}
}\\
\subfloat[With data standardization.]{
    \includegraphics[width=0.99\textwidth]{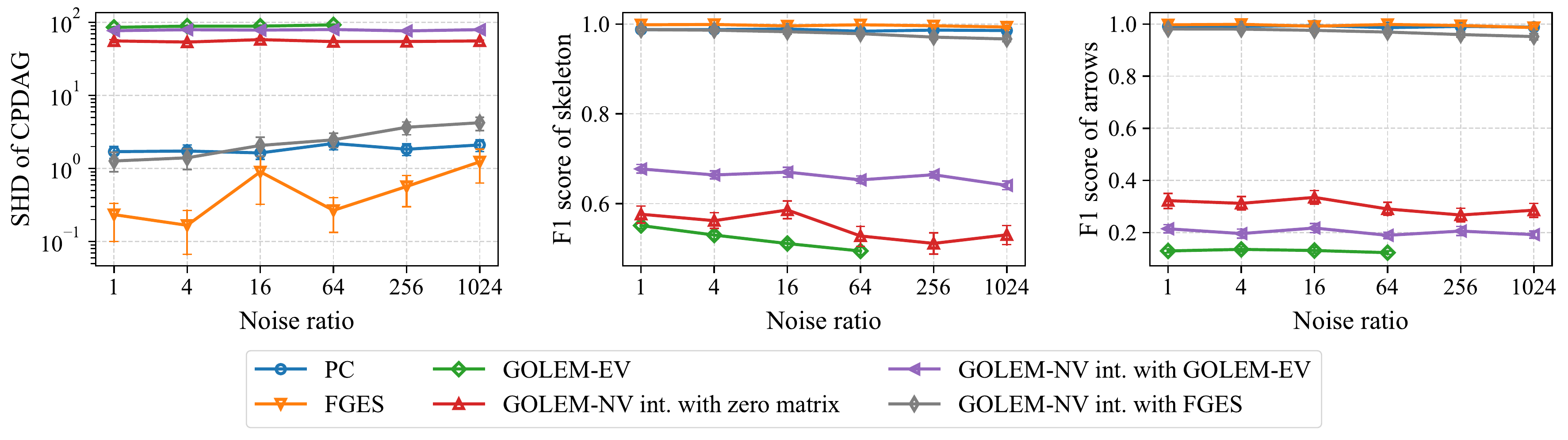}
    \label{fig:varsortability_nv_case_golem_standardized}
}
\caption{Empirical results of structure learning methods under different noise ratios. The number of variables is $50$ and the sample size is $10^6$. Error bars represent the standard errors computed over $30$ random repetitions. Here, ``int.'' stands for ``initialized''.}
\label{fig:varsortability_nv_case_result}
\end{figure}

\subsection{Search Strategies}\label{app:search_strategies}
\begin{figure}[H]
\centering
\includegraphics[width=0.99\textwidth]{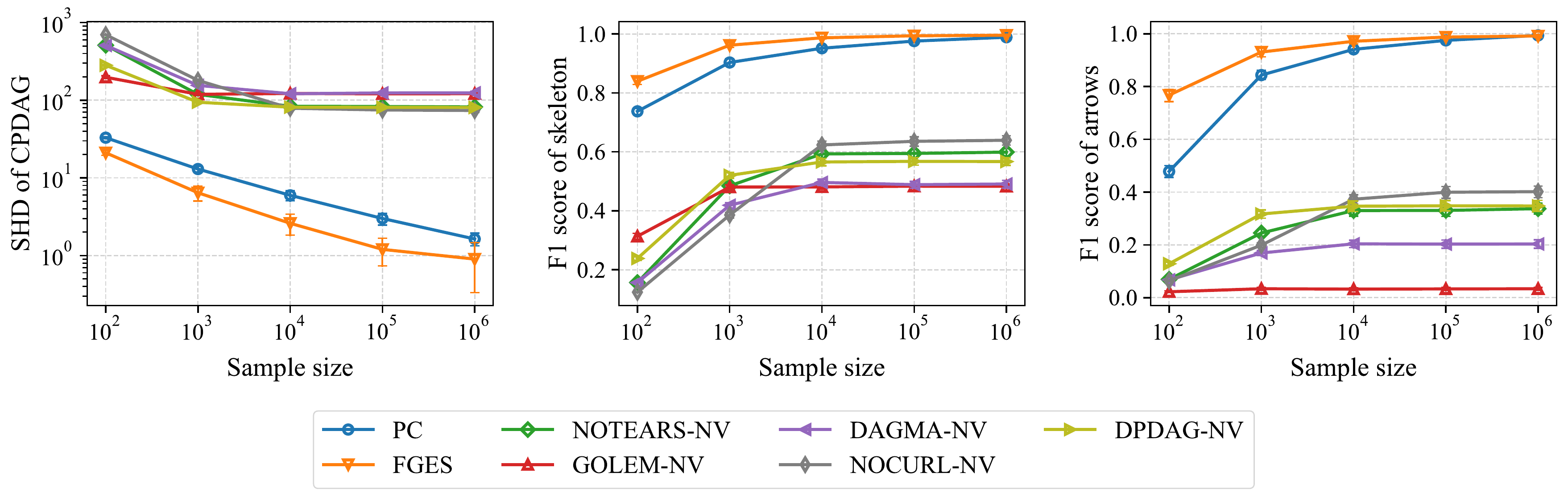}
\caption{Empirical results of different search strategies under different sample sizes. The number of variables is $50$. Error bars represent the standard errors computed over $30$ random repetitions.
}
\label{fig:search_strategies_result}
\end{figure}

\subsection{DAG Constraints}\label{app:dag_constraints}
\begin{figure}[H]
\centering  
\subfloat[GOLEM-NV with different DAG constraints.]{
    \includegraphics[width=0.99\textwidth]{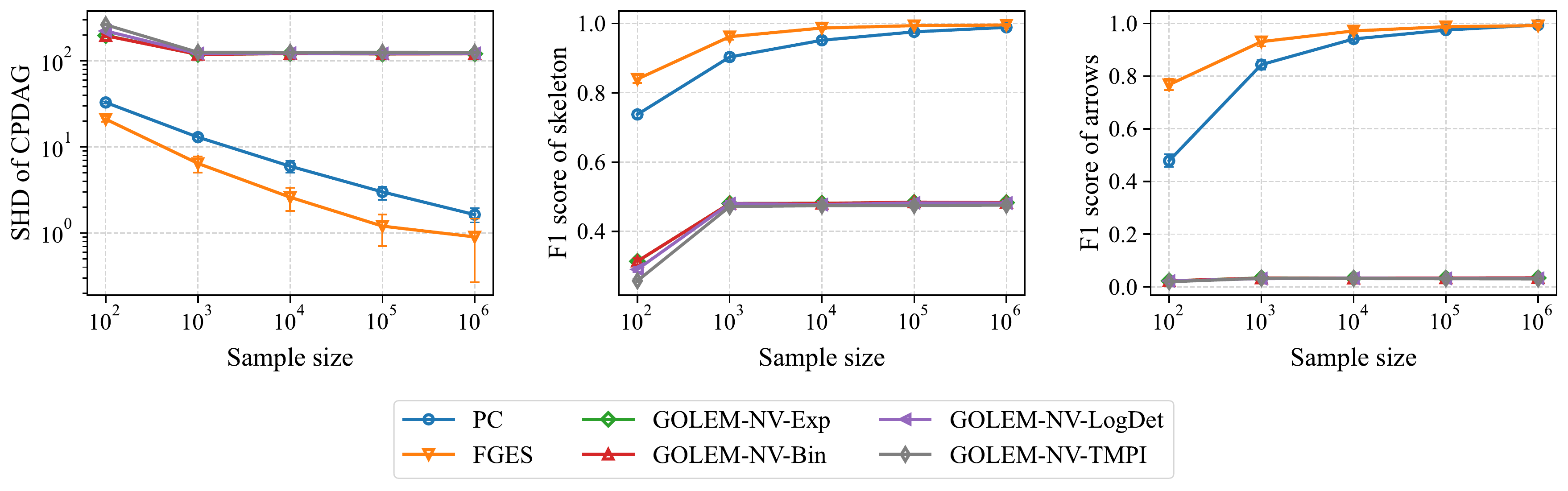}
    \label{fig:dag_constraints_result_golem}
}\\
\subfloat[NOTEARS-NV with different DAG constraints.]{
    \includegraphics[width=0.99\textwidth]{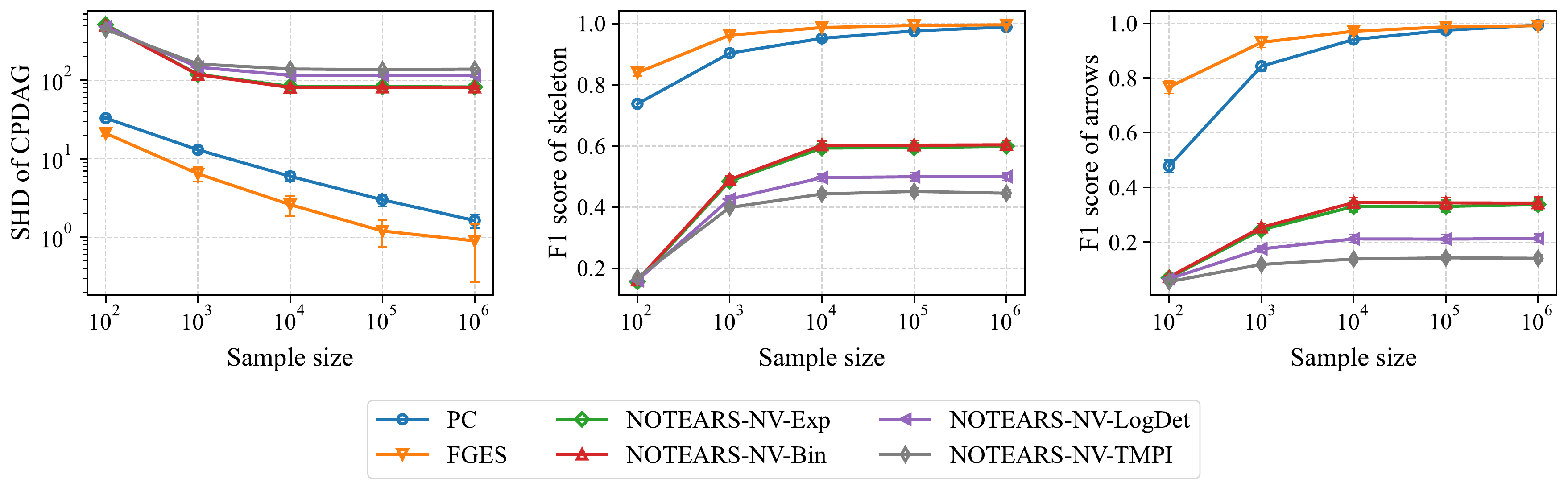}
    \label{fig:dag_constraints_result_notears}
}
\caption{Empirical results of different DAG constraints under different sample sizes. The number of variables is $50$. Error bars represent the standard errors computed over $30$ random repetitions.
}
\label{fig:dag_constraints_result}
\end{figure}

\clearpage
\subsection{Other Settings}\label{app:other_cases}
\begin{figure}[H]
\centering
\includegraphics[width=0.99\textwidth]{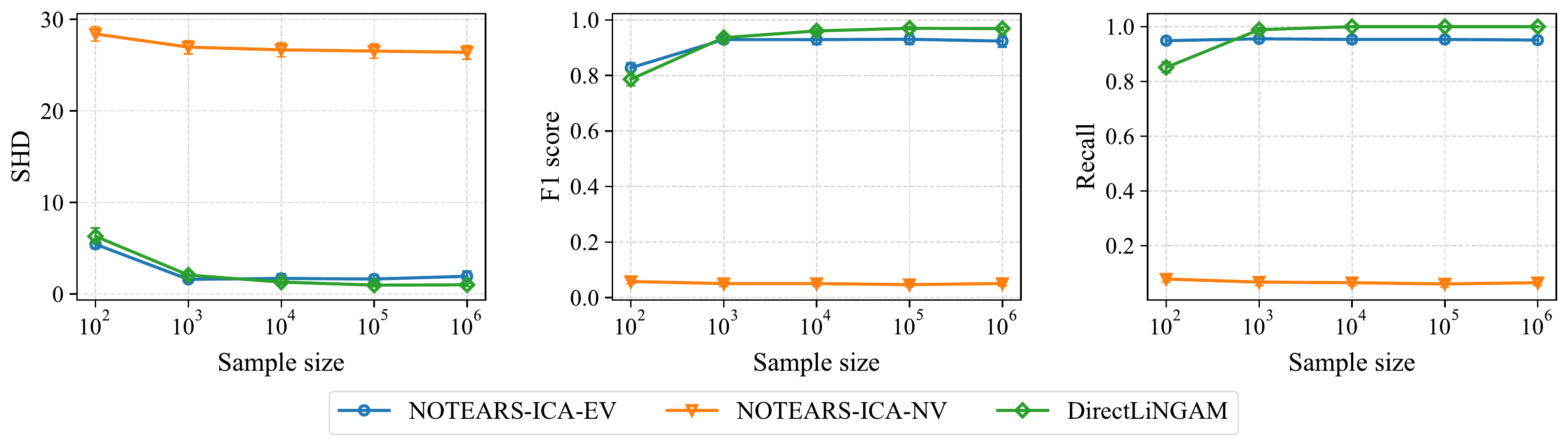}
\caption{Empirical results of NOTEARS-ICA under different sample sizes. The number of variables is $15$. Error bars represent the standard errors computed over $30$ random repetitions.
}
\label{fig:notears_ica_result}
\end{figure}
\begin{figure}[H]
\centering
\includegraphics[width=0.99\textwidth]{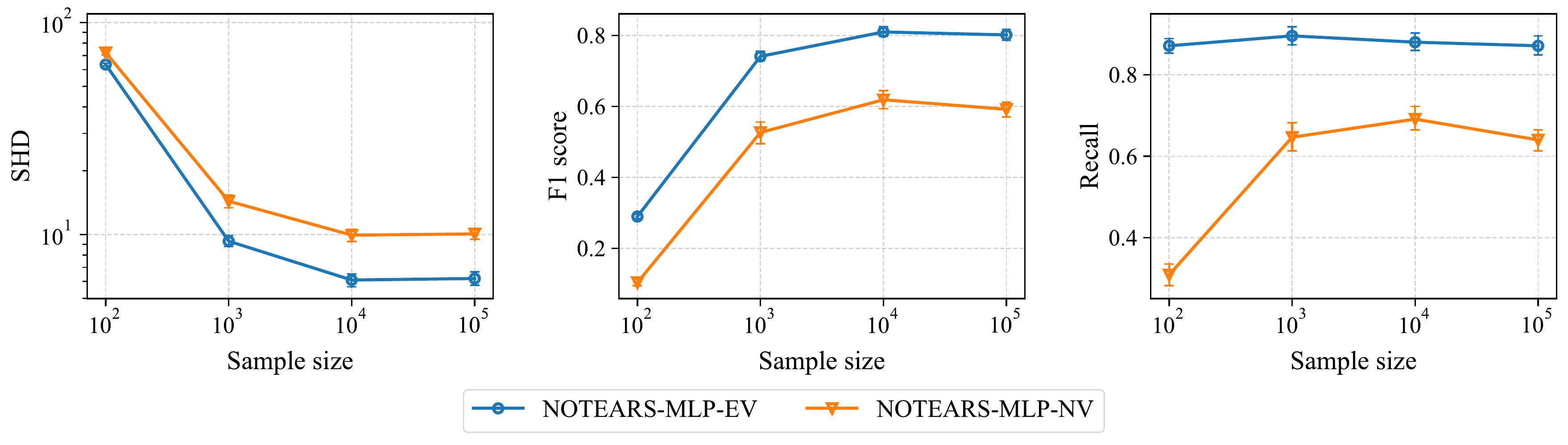}
\caption{Empirical results of NOTEARS-MLP under different sample sizes. The number of variables is $15$. Error bars represent the standard errors computed over $30$ random repetitions.
}
\label{fig:notears_mlp_result}
\end{figure}

\clearpage
\subsection{Thresholding}\label{app:thresholding}
\begin{figure}[H]
\centering  
\subfloat[SHD.]{
    \includegraphics[width=0.88\textwidth]{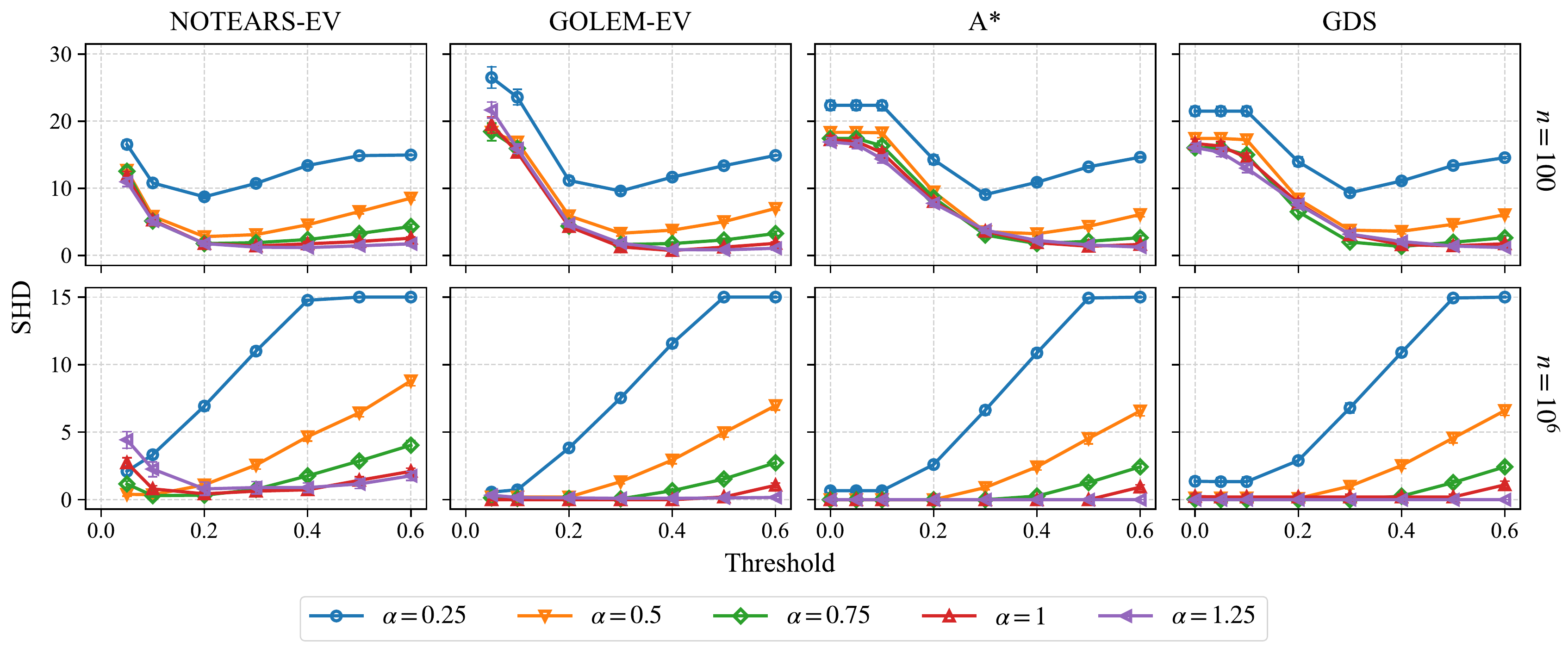}
    % \label{fig:thresholding_shd}
}\\
\subfloat[F1 score.]{
    \includegraphics[width=0.88\textwidth]{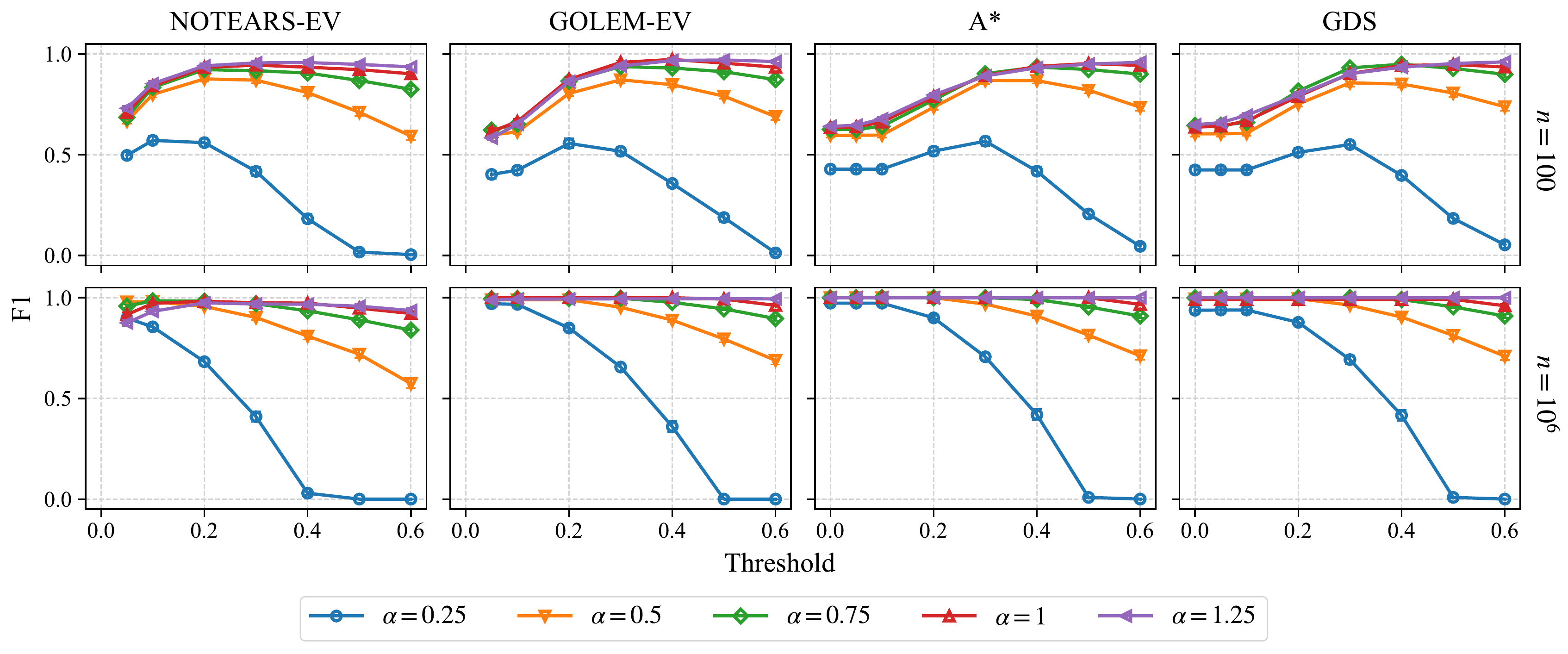}
    \label{fig:thresholding_f1}
}\\
\subfloat[Recall.]{
    \includegraphics[width=0.88\textwidth]{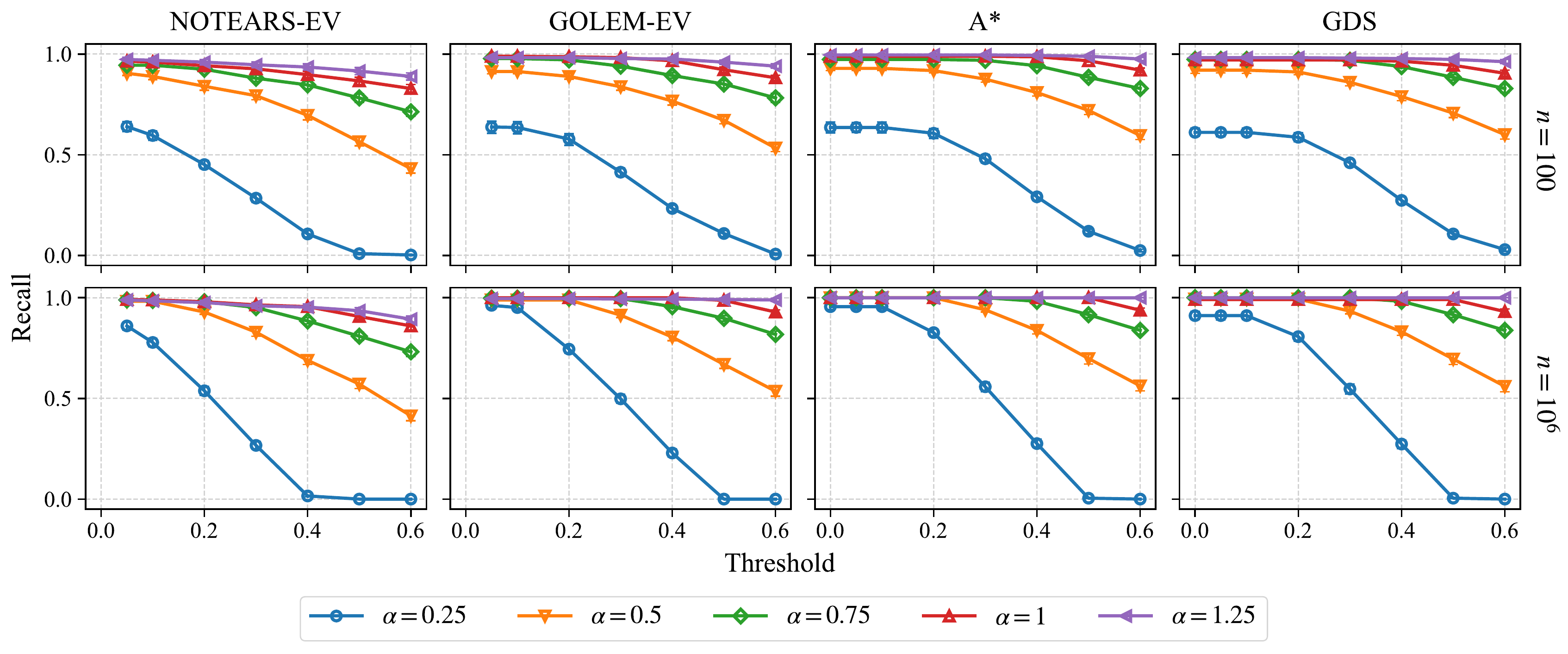}
    \label{fig:thresholding_recall}
}
\caption{Empirical results of different thresholds under different weight scales. The number of variables is $15$. Error bars represent the standard errors computed over $30$ random repetitions.
}
\label{fig:thresholding_result}
\end{figure}

\subsection{Sparsity Penalty}\label{app:sparsity}
\begin{figure}[H]
\centering
\subfloat[SHD for $15$ variables.]{
    \includegraphics[width=0.49\textwidth]{figures/other_issues/sparsity/15nodes_shd_zoomed.pdf}
    % \label{fig:sparsity_15nodes_shd}
}
\subfloat[F1 score for $15$ variables.]{
    \includegraphics[width=0.49\textwidth]{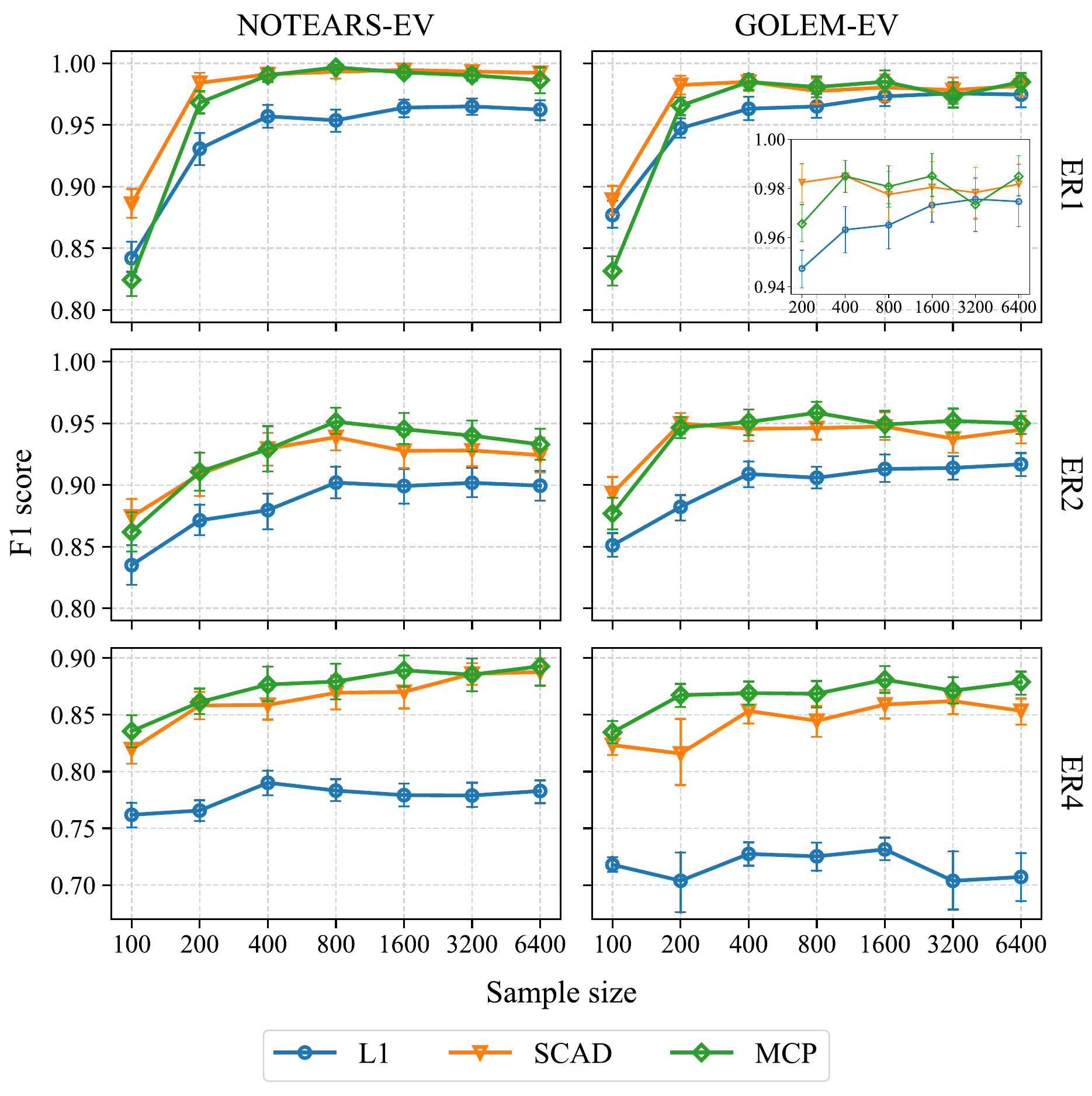}
    \label{fig:sparsity_15nodes_f1}
}\\
\subfloat[SHD for $50$ variables.]{
    \includegraphics[width=0.49\textwidth]{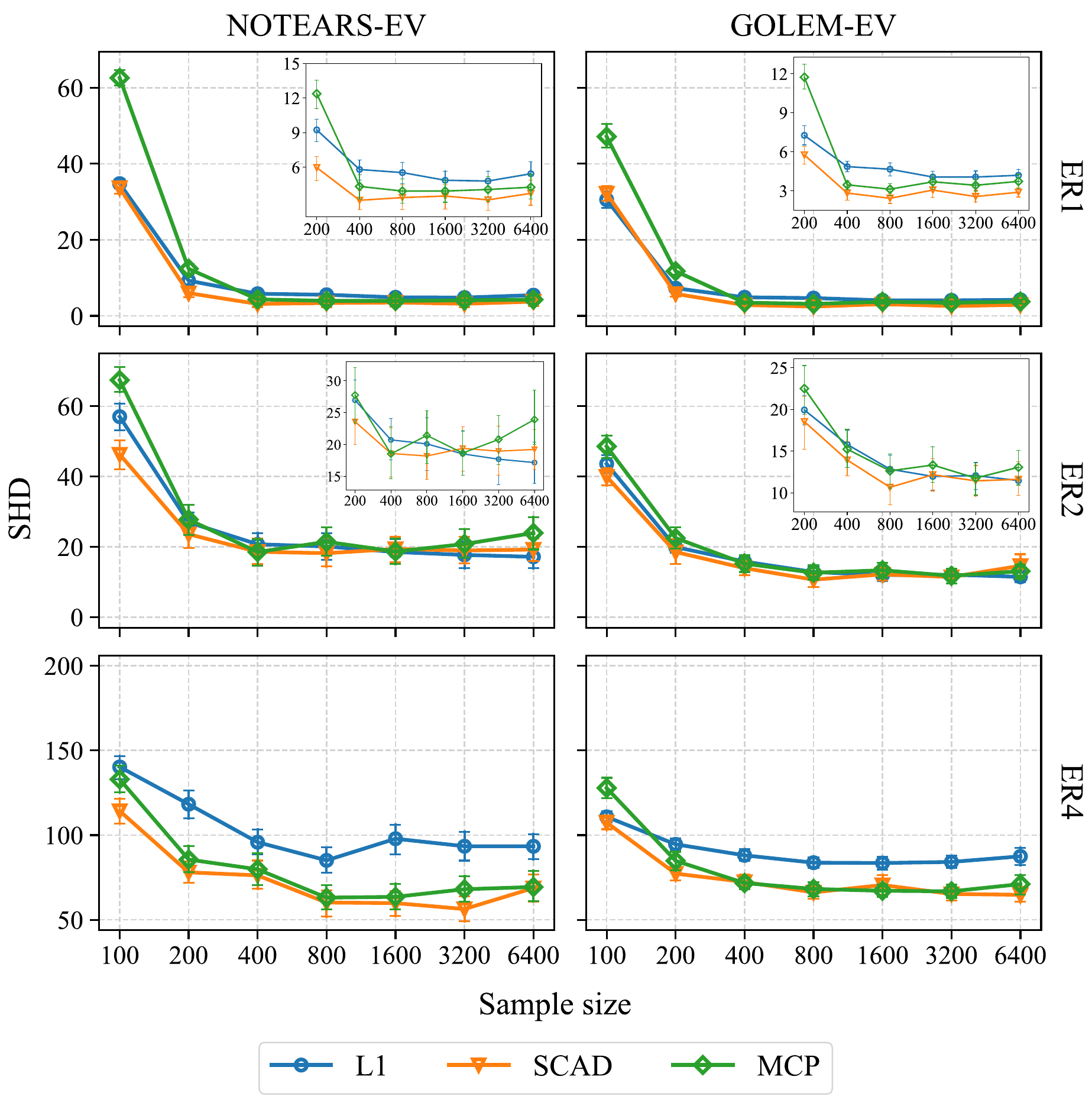}
    \label{fig:sparsity_50nodes_shd}
}
\subfloat[F1 score for $50$ variables.]{
    \includegraphics[width=0.49\textwidth]{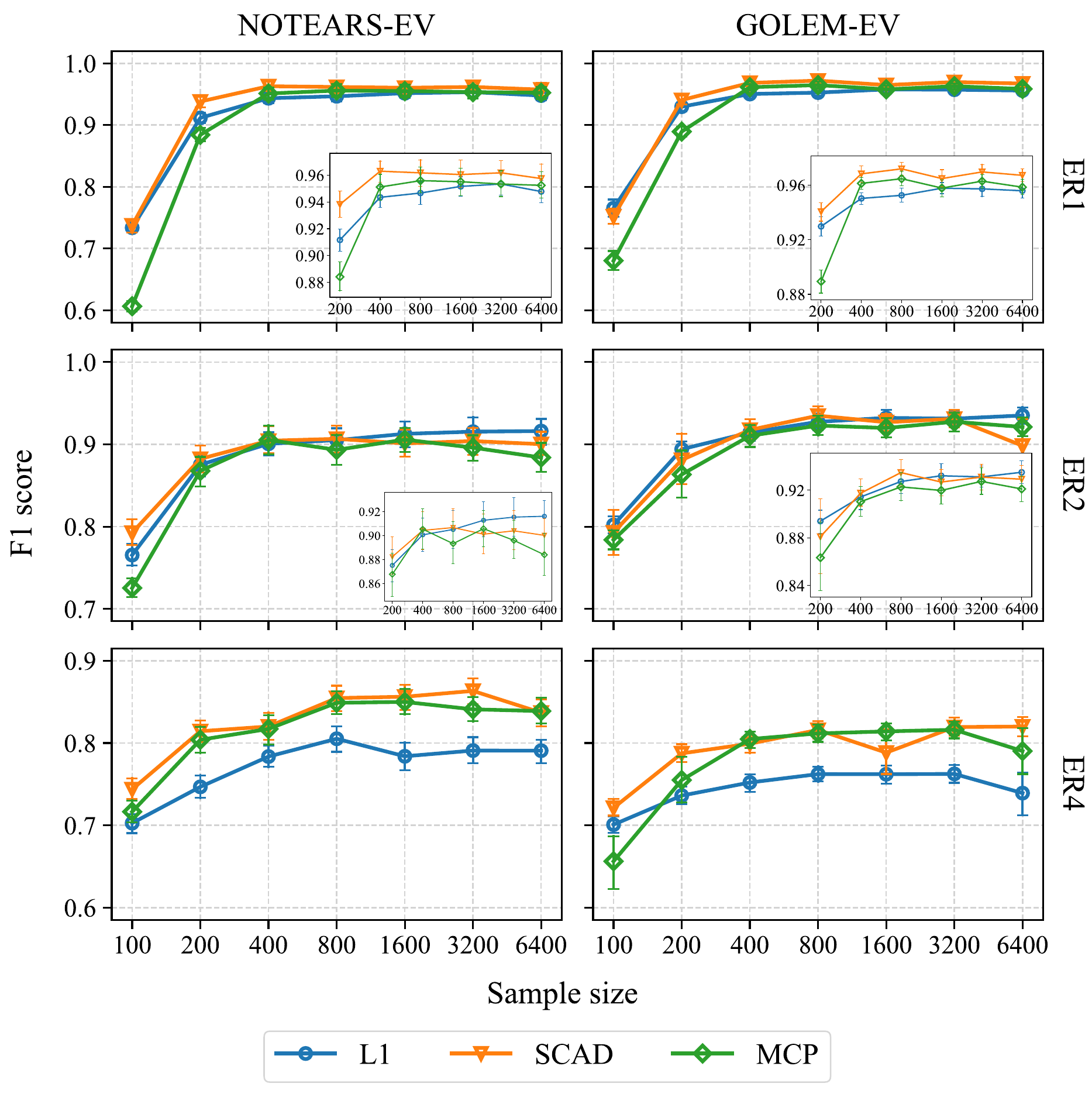}
    \label{fig:sparsity_50nodes_f1}
}\\
\caption{Empirical results of different sparsity penalties under different sample sizes. Error bars represent the standard errors computed over $30$ random repetitions.}
\label{fig:sparsity_result}
\end{figure}

%% file: ms.bbl
\begin{thebibliography}{75}
\providecommand{\natexlab}[1]{#1}
\providecommand{\url}[1]{\texttt{#1}}
\expandafter\ifx\csname urlstyle\endcsname\relax
  \providecommand{\doi}[1]{doi: #1}\else
  \providecommand{\doi}{doi: \begingroup \urlstyle{rm}\Url}\fi

\bibitem[Aragam et~al.(2019)Aragam, Amini, and Zhou]{Aragam2019globally}
B.~Aragam, A.~Amini, and Q.~Zhou.
\newblock Globally optimal score-based learning of directed acyclic graphs in
  high-dimensions.
\newblock In \emph{Advances in Neural Information Processing Systems},
  volume~32, 2019.

\bibitem[Bang-Jensen and Gutin(2009)]{Jorgen2009digraphs}
J.~Bang-Jensen and G.~Gutin.
\newblock \emph{Digraphs. Theory, Algorithms and Applications}.
\newblock Springer Monographs in Mathematics, 2009.

\bibitem[Bello et~al.(2022)Bello, Aragam, and Ravikumar]{Bello2022dagma}
K.~Bello, B.~Aragam, and P.~Ravikumar.
\newblock {DAGMA}: Learning {DAGs} via {M-matrices} and a log-determinant
  acyclicity characterization.
\newblock In \emph{Advances in Neural Information Processing Systems}, 2022.

\bibitem[Bellot and van~der Schaar(2021)]{Bellot2021deconfounded}
A.~Bellot and M.~van~der Schaar.
\newblock Deconfounded score method: Scoring {DAGs} with dense unobserved
  confounding.
\newblock \emph{arXiv preprint arXiv:2103.15106}, 2021.

\bibitem[Bertsekas(1999)]{Bertsekas1999nonlinear}
D.~P. Bertsekas.
\newblock \emph{Nonlinear Programming}.
\newblock Athena Scientific, 2nd edition, 1999.

\bibitem[Bhattacharya et~al.(2021)Bhattacharya, Nagarajan, Malinsky, and
  Shpitser]{Bhattacharya2020differentiable}
R.~Bhattacharya, T.~Nagarajan, D.~Malinsky, and I.~Shpitser.
\newblock Differentiable causal discovery under unmeasured confounding.
\newblock In \emph{International Conference on Artificial Intelligence and
  Statistics}, 2021.

\bibitem[Breheny and Huang(2011)]{Breheny2011coordinate}
P.~Breheny and J.~Huang.
\newblock Coordinate descent algorithms for nonconvex penalized regression,
  with applications to biological feature selection.
\newblock \emph{The Annals of Applied Statistics}, 5\penalty0 (1):\penalty0
  232--253, 2011.

\bibitem[Brouillard et~al.(2020)Brouillard, Lachapelle, Lacoste,
  Lacoste-Julien, and Drouin]{brouillard2020differentiable}
P.~Brouillard, S.~Lachapelle, A.~Lacoste, S.~Lacoste-Julien, and A.~Drouin.
\newblock Differentiable causal discovery from interventional data.
\newblock In \emph{Advances in Neural Information Processing Systems}, 2020.

\bibitem[Charpentier et~al.(2022)Charpentier, Kibler, and
  G{\"u}nnemann]{Charpentier2022differentiable}
B.~Charpentier, S.~Kibler, and S.~G{\"u}nnemann.
\newblock Differentiable {DAG} sampling.
\newblock In \emph{International Conference on Learning Representations}, 2022.

\bibitem[Chen et~al.(2021)Chen, Sun, Ellington, Xing, and
  Song]{Chen2021multitask}
X.~Chen, H.~Sun, C.~Ellington, E.~Xing, and L.~Song.
\newblock Multi-task learning of order-consistent causal graphs.
\newblock In \emph{Advances in Neural Information Processing Systems}, 2021.

\bibitem[Chickering(1996)]{Chickering1996learning}
D.~M. Chickering.
\newblock Learning {Bayesian} networks is {NP}-complete.
\newblock In \emph{Learning from Data: Artificial Intelligence and Statistics
  V}. Springer, 1996.

\bibitem[Chickering(2002)]{Chickering2002optimal}
D.~M. Chickering.
\newblock Optimal structure identification with greedy search.
\newblock \emph{Journal of Machine Learning Research}, 3\penalty0
  (Nov):\penalty0 507--554, 2002.

\bibitem[Chickering et~al.(2004)Chickering, Heckerman, and
  Meek]{Chickering2004large}
D.~M. Chickering, D.~Heckerman, and C.~Meek.
\newblock Large-sample learning of {Bayesian} networks is {NP}-hard.
\newblock \emph{Journal of Machine Learning Research}, 5, 2004.

\bibitem[Cussens(2011)]{Cussens2011bayesian}
J.~Cussens.
\newblock Bayesian network learning with cutting planes.
\newblock In \emph{Conference on Uncertainty in Artificial Intelligence}, 2011.

\bibitem[Deng et~al.(2023)Deng, Bello, Ravikumar, and Aragam]{Deng2023global}
C.~Deng, K.~Bello, P.~K. Ravikumar, and B.~Aragam.
\newblock Global optimality in bivariate gradient-based {DAG} learning.
\newblock In \emph{Advances in Neural Information Processing Systems}, 2023.

\bibitem[Dor and Tarsi(1992)]{Dor1992simple}
D.~Dor and M.~Tarsi.
\newblock A simple algorithm to construct a consistent extension of a partially
  oriented graph.
\newblock \emph{Technical Report R-185, Cognitive Systems Laboratory, UCLA},
  1992.

\bibitem[Erd\"os and R\'enyi(1959)]{Erdos1959random}
P.~Erd\"os and A.~R\'enyi.
\newblock On random graphs {I}.
\newblock \emph{Publicationes Mathematicae}, 6:\penalty0 290--297, 1959.

\bibitem[Fan and Li(2001)]{Fan2001variable}
J.~Fan and R.~Li.
\newblock Variable selection via nonconcave penalized likelihood and its oracle
  properties.
\newblock \emph{Journal of the American statistical Association}, 96\penalty0
  (456):\penalty0 1348--1360, 2001.

\bibitem[Faria et~al.(2022)Faria, Martins, and
  Figueiredo]{Faria2022differentiable}
G.~R.~A. Faria, A.~Martins, and M.~A.~T. Figueiredo.
\newblock Differentiable causal discovery under latent interventions.
\newblock In \emph{Conference on Causal Learning and Reasoning}, 2022.

\bibitem[Gao et~al.(2022)Gao, Ng, Gong, Shen, Huang, Liu, Zhang, and
  Bondell]{Gao2022missdag}
E.~Gao, I.~Ng, M.~Gong, L.~Shen, W.~Huang, T.~Liu, K.~Zhang, and H.~Bondell.
\newblock {MissDAG}: Causal discovery in the presence of missing data with
  continuous additive noise models.
\newblock In \emph{Advances in Neural Information Processing Systems}, 2022.

\bibitem[Gao et~al.(2023)Gao, Chen, Shen, Liu, Gong, and
  Bondell]{Gao2023feddag}
E.~Gao, J.~Chen, L.~Shen, T.~Liu, M.~Gong, and H.~Bondell.
\newblock Fed{DAG}: Federated {DAG} structure learning.
\newblock \emph{Transactions on Machine Learning Research}, 2023.
\newblock ISSN 2835-8856.

\bibitem[Gong et~al.(2022)Gong, Smith, Wang, Barton, Woodhead, Pawlowski,
  Jennings, and Zhang]{Gong2022neurips}
W.~Gong, D.~Smith, Z.~Wang, C.~Barton, S.~Woodhead, N.~Pawlowski, J.~Jennings,
  and C.~Zhang.
\newblock {NeurIPS} competition instructions and guide: Causal insights for
  learning paths in education.
\newblock \emph{arXiv preprint arXiv:2208.12610}, 2022.

\bibitem[Hoyer et~al.(2009)Hoyer, Janzing, Mooij, Peters, and
  Sch\"{o}lkopf]{Hoyer2009nonlinear}
P.~Hoyer, D.~Janzing, J.~M. Mooij, J.~Peters, and B.~Sch\"{o}lkopf.
\newblock Nonlinear causal discovery with additive noise models.
\newblock In \emph{Advances in Neural Information Processing Systems}, 2009.

\bibitem[Ikeuchi et~al.(2023)Ikeuchi, Ide, Zeng, Maeda, and
  Shimizu]{Ikeuchi2023python}
T.~Ikeuchi, M.~Ide, Y.~Zeng, T.~N. Maeda, and S.~Shimizu.
\newblock Python package for causal discovery based on {LiNGAM}.
\newblock \emph{Journal of Machine Learning Research}, 24\penalty0
  (14):\penalty0 1--8, 2023.

\bibitem[Jiang et~al.(2011)Jiang, Lim, Yao, and Ye]{Jiang2011statistical}
X.~Jiang, L.-H. Lim, Y.~Yao, and Y.~Ye.
\newblock Statistical ranking and combinatorial {Hodge} theory.
\newblock \emph{Mathematical Programming}, 127, 11 2011.

\bibitem[Kaiser and Sipos(2022)]{Kaiser2022unsuitability}
M.~Kaiser and M.~Sipos.
\newblock Unsuitability of {NOTEARS} for causal graph discovery when dealing
  with dimensional quantities.
\newblock \emph{Neural Processing Letters}, 54:\penalty0 1--9, 06 2022.

\bibitem[Kalainathan et~al.(2022)Kalainathan, Goudet, Guyon, Lopez-Paz, and
  Sebag]{Kalainathan2022structural}
D.~Kalainathan, O.~Goudet, I.~Guyon, D.~Lopez-Paz, and M.~Sebag.
\newblock Structural agnostic modeling: Adversarial learning of causal graphs.
\newblock \emph{Journal of Machine Learning Research}, 23\penalty0
  (219):\penalty0 1--62, 2022.

\bibitem[Kingma and Ba(2014)]{Kingma2014adam}
D.~Kingma and J.~Ba.
\newblock Adam: A method for stochastic optimization.
\newblock In \emph{International Conference on Learning Representations}, 2014.

\bibitem[Koivisto and Sood(2004)]{Koivisto2004exact}
M.~Koivisto and K.~Sood.
\newblock Exact {Bayesian} structure discovery in {Bayesian} networks.
\newblock \emph{Journal of Machine Learning Research}, 5\penalty0
  (Dec):\penalty0 549--573, 2004.

\bibitem[Koller and Friedman(2009)]{Koller09probabilistic}
D.~Koller and N.~Friedman.
\newblock \emph{Probabilistic Graphical Models: Principles and Techniques}.
\newblock MIT Press, Cambridge, MA, 2009.

\bibitem[Kool et~al.(2019)Kool, Van~Hoof, and Welling]{Kool2019stochastic}
W.~Kool, H.~Van~Hoof, and M.~Welling.
\newblock Stochastic beams and where to find them: The {G}umbel-top-k trick for
  sampling sequences without replacement.
\newblock In \emph{International Conference on Machine Learning}, 2019.

\bibitem[Lachapelle et~al.(2020)Lachapelle, Brouillard, Deleu, and
  Lacoste-Julien]{Lachapelle2020grandag}
S.~Lachapelle, P.~Brouillard, T.~Deleu, and S.~Lacoste-Julien.
\newblock Gradient-based neural {DAG} learning.
\newblock In \emph{International Conference on Learning Representations}, 2020.

\bibitem[Loh and B{{\"u}}hlmann(2014)]{Loh2014high}
P.-L. Loh and P.~B{{\"u}}hlmann.
\newblock High-dimensional learning of linear causal networks via inverse
  covariance estimation.
\newblock \emph{Journal of Machine Learning Research}, 15\penalty0
  (88):\penalty0 3065--3105, 2014.

\bibitem[Loh and Wainwright(2017)]{Loh2017support}
P.-L. Loh and M.~J. Wainwright.
\newblock Support recovery without incoherence: A case for nonconvex
  regularization.
\newblock \emph{The Annals of Statistics}, 45\penalty0 (6):\penalty0
  2455--2482, 2017.

\bibitem[Mena et~al.(2018)Mena, Belanger, Linderman, and
  Snoek]{Mena2018learning}
G.~Mena, D.~Belanger, S.~Linderman, and J.~Snoek.
\newblock Learning latent permutations with {Gumbel-Sinkhorn} networks.
\newblock In \emph{International Conference on Learning Representations}, 2018.

\bibitem[Ng and Zhang(2022)]{Ng2022towards}
I.~Ng and K.~Zhang.
\newblock Towards federated {Bayesian} network structure learning with
  continuous optimization.
\newblock In \emph{International Conference on Artificial Intelligence and
  Statistics}, 2022.

\bibitem[Ng et~al.(2020)Ng, Ghassami, and Zhang]{Ng2020role}
I.~Ng, A.~Ghassami, and K.~Zhang.
\newblock On the role of sparsity and {DAG} constraints for learning linear
  {DAGs}.
\newblock In \emph{Advances in Neural Information Processing Systems}, 2020.

\bibitem[Ng et~al.(2022{\natexlab{a}})Ng, Lachapelle, Ke, Lacoste-Julien, and
  Zhang]{Ng2022convergence}
I.~Ng, S.~Lachapelle, N.~R. Ke, S.~Lacoste-Julien, and K.~Zhang.
\newblock On the convergence of continuous constrained optimization for
  structure learning.
\newblock In \emph{International Conference on Artificial Intelligence and
  Statistics}, 2022{\natexlab{a}}.

\bibitem[Ng et~al.(2022{\natexlab{b}})Ng, Zhu, Fang, Li, Chen, and
  Wang]{Ng2022masked}
I.~Ng, S.~Zhu, Z.~Fang, H.~Li, Z.~Chen, and J.~Wang.
\newblock Masked gradient-based causal structure learning.
\newblock In \emph{SIAM International Conference on Data Mining},
  2022{\natexlab{b}}.

\bibitem[Nocedal and Wright(2006)]{Nocedal2006numerical}
J.~Nocedal and S.~J. Wright.
\newblock \emph{Numerical optimization}.
\newblock Springer series in operations research and financial engineering.
  Springer, 2nd edition, 2006.

\bibitem[Pamfil et~al.(2020)Pamfil, Sriwattanaworachai, Desai, Pilgerstorfer,
  Beaumont, Georgatzis, and Aragam]{Pamfil2020dynotears}
R.~Pamfil, N.~Sriwattanaworachai, S.~Desai, P.~Pilgerstorfer, P.~Beaumont,
  K.~Georgatzis, and B.~Aragam.
\newblock {DYNOTEARS}: Structure learning from time-series data.
\newblock In \emph{International Conference on Artificial Intelligence and
  Statistics}, 2020.

\bibitem[Pearl(1988)]{Pearl1988probabilistic}
J.~Pearl.
\newblock \emph{Probabilistic Reasoning in Intelligent Systems: Networks of
  Plausible Inference}.
\newblock Morgan Kaufmann, 1988.

\bibitem[Peters and Bühlmann(2013)]{Peters2013identifiability}
J.~Peters and P.~Bühlmann.
\newblock Identifiability of {Gaussian} structural equation models with equal
  error variances.
\newblock \emph{Biometrika}, 101\penalty0 (1):\penalty0 219--228, 2013.

\bibitem[Peters et~al.(2017)Peters, Janzing, and
  Sch{\"o}lkopf]{Peters2017elements}
J.~Peters, D.~Janzing, and B.~Sch{\"o}lkopf.
\newblock \emph{Elements of Causal Inference - Foundations and Learning
  Algorithms}.
\newblock MIT Press, 2017.

\bibitem[Ramsey et~al.(2017)Ramsey, Glymour, Sanchez-Romero, and
  Glymour]{Ramsey2017million}
J.~Ramsey, M.~Glymour, R.~Sanchez-Romero, and C.~Glymour.
\newblock A million variables and more: the fast greedy equivalence search
  algorithm for learning high-dimensional graphical causal models, with an
  application to functional magnetic resonance images.
\newblock \emph{International Journal of Data Science and Analytics},
  3\penalty0 (2):\penalty0 121--129, 2017.

\bibitem[Ravikumar et~al.(2011)Ravikumar, Wainwright, Raskutti, and
  Yu]{Ravikumar2011high}
P.~Ravikumar, M.~J. Wainwright, G.~Raskutti, and B.~Yu.
\newblock High-dimensional covariance estimation by minimizing
  $\ell_1$-penalized log-determinant divergence.
\newblock \emph{Electronic Journal of Statistics}, 5:\penalty0 935--980, 2011.

\bibitem[Reisach et~al.(2021)Reisach, Seiler, and Weichwald]{Reisach2021beware}
A.~Reisach, C.~Seiler, and S.~Weichwald.
\newblock Beware of the simulated {DAG}! {Causal} discovery benchmarks may be
  easy to game.
\newblock In \emph{Advances in Neural Information Processing Systems}, 2021.

\bibitem[Schatzman and Taylor(2002)]{Schatzman2002numerical}
M.~Schatzman and J.~Taylor.
\newblock \emph{Numerical Analysis A Mathematical Introduction}.
\newblock Oxford University Press, 2002.

\bibitem[Scheines et~al.(1998)Scheines, Spirtes, Glymour, Meek, and
  Richardson]{Scheines1998tetrad}
R.~Scheines, P.~Spirtes, C.~Glymour, C.~Meek, and T.~Richardson.
\newblock The {TETRAD} project: Constraint based aids to causal model
  specification.
\newblock \emph{Multivariate Behavioral Research}, 33:\penalty0 65--117, 1998.

\bibitem[Schwarz(1978)]{Schwarz1978estimating}
G.~Schwarz.
\newblock Estimating the dimension of a model.
\newblock \emph{The Annals of Statistics}, 6\penalty0 (2):\penalty0 461--464,
  1978.

\bibitem[Shimizu et~al.(2006)Shimizu, Hoyer, Hyv{\"a}rinen, and
  Kerminen]{Shimizu2006lingam}
S.~Shimizu, P.~O. Hoyer, A.~Hyv{\"a}rinen, and A.~Kerminen.
\newblock A linear {non-Gaussian} acyclic model for causal discovery.
\newblock \emph{Journal of Machine Learning Research}, 7\penalty0
  (Oct):\penalty0 2003--2030, 2006.

\bibitem[Shimizu et~al.(2011)Shimizu, Inazumi, Sogawa, Hyv{\"a}rinen, Kawahara,
  Washio, Hoyer, and Bollen]{Shimizu2011directlingam}
S.~Shimizu, T.~Inazumi, Y.~Sogawa, A.~Hyv{\"a}rinen, Y.~Kawahara, T.~Washio,
  P.~O. Hoyer, and K.~Bollen.
\newblock {DirectLiNGAM}: A direct method for learning a linear {non-Gaussian}
  structural equation model.
\newblock \emph{Journal of Machine Learning Research}, 12\penalty0
  (Apr):\penalty0 1225--1248, 2011.

\bibitem[Singh and Moore(2005)]{Singh2005finding}
A.~P. Singh and A.~W. Moore.
\newblock Finding optimal {Bayesian} networks by dynamic programming.
\newblock Technical report, Carnegie Mellon University, 2005.

\bibitem[Spirtes and Glymour(1991)]{Spirtes1991pc}
P.~Spirtes and C.~Glymour.
\newblock An algorithm for fast recovery of sparse causal graphs.
\newblock \emph{Social Science Computer Review}, 9:\penalty0 62--72, 1991.

\bibitem[Spirtes et~al.(2001)Spirtes, Glymour, and
  Scheines]{Spirtes2001causation}
P.~Spirtes, C.~Glymour, and R.~Scheines.
\newblock \emph{Causation, Prediction, and Search}.
\newblock MIT press, 2nd edition, 2001.

\bibitem[Stewart(1969)]{Stewart1969continuity}
G.~W. Stewart.
\newblock On the continuity of the generalized inverse.
\newblock \emph{SIAM Journal on Applied Mathematics}, 17\penalty0 (1):\penalty0
  33--45, 1969.
\newblock ISSN 00361399.

\bibitem[Sun et~al.(2021)Sun, Liu, Poupart, and Schulte]{Sun2021ntsnotears}
X.~Sun, G.~Liu, P.~Poupart, and O.~Schulte.
\newblock {NTS-NOTEARS}: Learning nonparametric temporal {DAGs} with
  time-series data and prior knowledge.
\newblock \emph{arXiv preprint arXiv:2109.04286}, 2021.

\bibitem[Vowels et~al.(2022)Vowels, Camgoz, and Bowden]{Vowels2022dags}
M.~J. Vowels, N.~C. Camgoz, and R.~Bowden.
\newblock D’ya like {DAGs}? {A} survey on structure learning and causal
  discovery.
\newblock \emph{ACM Computing Surveys}, 55\penalty0 (4), nov 2022.
\newblock ISSN 0360-0300.

\bibitem[Wainwright(2009)]{Wainwright2009sharp}
M.~J. Wainwright.
\newblock Sharp thresholds for high-dimensional and noisy sparsity recovery
  using $\ell _{1}$ -constrained quadratic programming ({Lasso}).
\newblock \emph{IEEE Transactions on Information Theory}, 55\penalty0
  (5):\penalty0 2183--2202, 2009.

\bibitem[Wang et~al.(2020)Wang, Menkovski, Wang, Du, and
  Pechenizkiy]{Wang2020causal}
Y.~Wang, V.~Menkovski, H.~Wang, X.~Du, and M.~Pechenizkiy.
\newblock Causal discovery from incomplete data: A deep learning approach.
\newblock \emph{arXiv preprint arXiv:2001.05343}, 2020.

\bibitem[Wei et~al.(2020)Wei, Gao, and Yu]{Wei2020nofears}
D.~Wei, T.~Gao, and Y.~Yu.
\newblock {DAGs} with no fears: A closer look at continuous optimization for
  learning {Bayesian} networks.
\newblock In \emph{Advances in Neural Information Processing Systems}, 2020.

\bibitem[Wilkinson(1965)]{Wilkinson1965rounding}
J.~H. Wilkinson.
\newblock \emph{Rounding Errors in Algebraic Processes}.
\newblock The Harlan D. Mills Collection, 1965.

\bibitem[Xu et~al.(2022)Xu, Gao, Huang, Song, and Gong]{Xu2022sparse}
D.~Xu, E.~Gao, W.~Huang, A.~Song, and M.~Gong.
\newblock On the sparse {DAG} structure learning based on adaptive {Lasso}.
\newblock \emph{arXiv preprint arXiv:2209.02946}, 2022.

\bibitem[Yang et~al.(2021)Yang, Liu, Chen, Shen, Hao, and
  Wang]{Yang2021causalvae}
M.~Yang, F.~Liu, Z.~Chen, X.~Shen, J.~Hao, and J.~Wang.
\newblock {CausalVAE}: Disentangled representation learning via neural
  structural causal models.
\newblock In \emph{Proceedings of the IEEE Conference on Computer Vision and
  Pattern Recognition}, 2021.

\bibitem[Yu et~al.(2019)Yu, Chen, Gao, and Yu]{Yu19daggnn}
Y.~Yu, J.~Chen, T.~Gao, and M.~Yu.
\newblock {DAG-GNN}: {DAG} structure learning with graph neural networks.
\newblock In \emph{International Conference on Machine Learning}, 2019.

\bibitem[Yu et~al.(2021)Yu, Gao, Yin, and Ji]{Yu2021nocurl}
Y.~Yu, T.~Gao, N.~Yin, and Q.~Ji.
\newblock {DAGs} with no curl: An efficient {DAG} structure learning approach.
\newblock In \emph{International Conference on Machine Learning}, 2021.

\bibitem[Yuan and Malone(2013)]{Yuan2013learning}
C.~Yuan and B.~Malone.
\newblock Learning optimal {Bayesian} networks: A shortest path perspective.
\newblock \emph{Journal of Artificial Intelligence Research}, 48\penalty0
  (1):\penalty0 23--65, 2013.

\bibitem[Zeng et~al.(2021)Zeng, Shimizu, Cai, Xie, Yamamoto, and
  Hao]{Zeng2020causal}
Y.~Zeng, S.~Shimizu, R.~Cai, F.~Xie, M.~Yamamoto, and Z.~Hao.
\newblock Causal discovery with multi-domain {LiNGAM} for latent factors.
\newblock In \emph{International Joint Conference on Artificial Intelligence},
  2021.

\bibitem[Zhang(2010)]{Zhang2010nearly}
C.-H. Zhang.
\newblock Nearly unbiased variable selection under minimax concave penalty.
\newblock \emph{The Annals of Statistics}, 38\penalty0 (2):\penalty0 894--942,
  2010.

\bibitem[Zhang et~al.(2022)Zhang, Ng, Gong, Liu, Abbasnejad, Gong, Zhang, and
  Shi]{Zhang2022truncated}
Z.~Zhang, I.~Ng, D.~Gong, Y.~Liu, E.~M. Abbasnejad, M.~Gong, K.~Zhang, and
  J.~Q. Shi.
\newblock Truncated matrix power iteration for differentiable {DAG} learning.
\newblock In \emph{Advances in Neural Information Processing Systems
  (NeurIPS)}, 2022.

\bibitem[Zheng(2020)]{Zheng2020thesis}
X.~Zheng.
\newblock \emph{Learning {DAGs} with Continuous Optimization}.
\newblock PhD thesis, Carnegie Mellon University, 2020.

\bibitem[Zheng et~al.(2018)Zheng, Aragam, Ravikumar, and
  Xing]{Zheng2018notears}
X.~Zheng, B.~Aragam, P.~Ravikumar, and E.~P. Xing.
\newblock {DAGs} with {NO TEARS}: Continuous optimization for structure
  learning.
\newblock In \emph{Advances in Neural Information Processing Systems}, 2018.

\bibitem[Zheng et~al.(2020)Zheng, Dan, Aragam, Ravikumar, and
  Xing]{Zheng2020learning}
X.~Zheng, C.~Dan, B.~Aragam, P.~Ravikumar, and E.~P. Xing.
\newblock Learning sparse nonparametric {DAGs}.
\newblock In \emph{International Conference on Artificial Intelligence and
  Statistics}, 2020.

\bibitem[Zheng et~al.(2024)Zheng, Huang, Chen, Ramsey, Gong, Cai, Shimizu,
  Spirtes, and Zhang]{Zheng2024causallearn}
Y.~Zheng, B.~Huang, W.~Chen, J.~Ramsey, M.~Gong, R.~Cai, S.~Shimizu,
  P.~Spirtes, and K.~Zhang.
\newblock Causal-learn: Causal discovery in {Python}.
\newblock \emph{Journal of Machine Learning Research}, 25\penalty0
  (60):\penalty0 1--8, 2024.

\bibitem[Zou(2006)]{Zou2006adaptive}
H.~Zou.
\newblock The adaptive {Lasso} and its oracle properties.
\newblock \emph{Journal of the American Statistical Association}, 101\penalty0
  (476):\penalty0 1418--1429, 2006.

\end{thebibliography}
